\PassOptionsToPackage{table}{xcolor}

\documentclass[a4paper,fleqn]{cas-sc}

\usepackage[authoryear]{natbib}

\usepackage{tikz}           
\usetikzlibrary{positioning, arrows.meta, calc, fit}
\usepackage{algorithm}      
\usepackage{algpseudocode}  

\hypersetup{
    linkcolor=blue!60!black,
    citecolor=blue!60!black,
    urlcolor=blue!70!black,
    breaklinks=true
}

\newif\ifcosmetic
\cosmetictrue          

\usepackage[most]{tcolorbox}

\definecolor{boxBlue}{HTML}{1B5FA8}
\definecolor{boxTeal}{HTML}{1B7F5E}
\definecolor{boxAmber}{HTML}{9C6500}
\definecolor{boxPurple}{HTML}{6A3D9A}
\definecolor{boxRed}{HTML}{C0392B}
\definecolor{boxGrey}{HTML}{5E5E5E}
\ifcosmetic\else
  \definecolor{boxBlue}{HTML}{000000}\definecolor{boxTeal}{HTML}{000000}
  \definecolor{boxAmber}{HTML}{000000}\definecolor{boxPurple}{HTML}{000000}
  \definecolor{boxRed}{HTML}{000000}\definecolor{boxGrey}{HTML}{000000}
\fi

\newcommand{\gaicon}[1]{%
  \raisebox{-0.15em}{\begin{tikzpicture}[scale=0.11,line width=0.9pt,
      line cap=round,line join=round]#1\end{tikzpicture}}}

\newcommand{\iconBulb}[1][boxBlue]{\gaicon{%
  \draw[#1] (0,0) .. controls (-2.2,2.2) and (-1.2,4.4) .. (0,4.8)
                  .. controls (1.2,4.4) and (2.2,2.2) .. (0,0);
  \draw[#1] (-0.9,-0.5) -- (0.9,-0.5);
  \draw[#1] (-0.6,-1.4) -- (0.6,-1.4);}}
\newcommand{\iconWarn}[1][boxAmber]{\gaicon{%
  \draw[#1] (0,4.6) -- (2.9,-0.8) -- (-2.9,-0.8) -- cycle;
  \draw[#1] (0,3.2) -- (0,1.0);
  \fill[#1] (0,0.1) circle (0.32);}}
\newcommand{\iconNote}[1][boxTeal]{\gaicon{%
  \draw[#1] (0,1.9) circle (2.7);
  \fill[#1] (0,3.3) circle (0.34);
  \draw[#1] (0,2.4) -- (0,0.3);}}

\newcommand{\iconKey}[1][boxPurple]{\gaicon{%
  \draw[#1] (-1.7,1.9) circle (1.5);
  \draw[#1] (-0.3,1.9) -- (2.9,1.9);
  \draw[#1] (2.0,1.9) -- (2.0,0.7);
  \draw[#1] (2.9,1.9) -- (2.9,0.9);}}

\newcommand{\icontext}[2]{\mbox{#1\,}\,#2}

\tcbset{
  gacallout/.style={
    enhanced, breakable, sharp corners=downhill,
    boxrule=0pt, leftrule=2.2pt, toprule=0pt, bottomrule=0pt, rightrule=0pt,
    left=8pt, right=8pt, top=6pt, bottom=6pt,
    fonttitle=\bfseries\small, fontupper=\small,
    coltitle=black, attach title to upper=\par\medskip,
    before skip=8pt, after skip=8pt
  }
}

\newtcolorbox{keycontrib}[1][Key contributions]{%
  gacallout, colback=boxPurple!5, colframe=boxPurple,
  title={\icontext{\iconKey}{#1}}}

\newtcolorbox{highlightbox}[1][Takeaway]{%
  gacallout, colback=boxBlue!5, colframe=boxBlue,
  title={\icontext{\iconBulb}{#1}}}

\newtcolorbox{notebox}[1][Note]{%
  gacallout, colback=boxTeal!5, colframe=boxTeal,
  title={\icontext{\iconNote}{#1}}}

\newtcolorbox{warnbox}[1][Caution]{%
  gacallout, colback=boxAmber!7, colframe=boxAmber,
  title={\icontext{\iconWarn}{#1}}}

\newtcolorbox{remarkbox}[1][Remark]{%
  gacallout, colback=boxGrey!5, colframe=boxGrey,
  title={\icontext{\iconNote[boxGrey]}{#1}}}

\ifcosmetic
  \definecolor{tblHeadBg}{HTML}{E8EEF6}
  \definecolor{tblAltBg}{HTML}{F6F8FB}
  \newcommand{\tblhead}{\rowcolor{tblHeadBg}}
  \newcommand{\zebra}{\rowcolors{2}{tblAltBg}{white}}
\else
  \newcommand{\tblhead}{}
  \newcommand{\zebra}{}
\fi

\begin{document}
\let\WriteBookmarks\relax

\renewcommand{\topfraction}{0.85}
\renewcommand{\bottomfraction}{0.70}
\renewcommand{\textfraction}{0.10}
\renewcommand{\floatpagefraction}{0.75}
\setcounter{topnumber}{3}
\setcounter{bottomnumber}{2}
\setcounter{totalnumber}{5}

\emergencystretch=1em

\shorttitle{Robust, personalized federated learning for turbofan prognostics}
\shortauthors{Mitra et al.}

\title[mode = title]{Robust and Personalized Federated Learning for
Aircraft-Engine Prognostics:\texorpdfstring{\\}{ }
Heterogeneity and Failure-Masking Poisoning on a Turbofan Benchmark}


\author[1]{\color{black}Chinmoy Mitra}
\author[2]{\color{black}Md. Mehedi Hasan Nipu}

\author[3]{\color{black}Mohammad Sakib Mahmood}

\author[4]{\color{black}Md. Rakibul Islam}

\author[5]{\color{black} M. F. Mridha}

\affiliation[1]{organization={Department of Computer Science and
                Engineering, Rajshahi University of Engineering \&
                Technology},
            addressline={Kazla},
            city={Rajshahi},
            postcode={6204},
            country={Bangladesh}}
\address{Department of Electrical and Computer Engineering, North South University, Dhaka 1229, Bangladesh.}
\address[3]{Department of Computer Science, Missouri State University, Springfield, MO, 65809, USA.}
\address[4]{Department of Computer Science and Engineering, Malardalen University, Vasteras, Vastmanland, Sweden.}
\address[5]{Department of Computer Science \& Engineering, American International University-Bangladesh, Dhaka, Bangladesh.}

\begin{abstract}
Federated learning (FL) enables aircraft fleet operators to jointly
train remaining-useful-life (RUL) models from engine sensor telemetry
without sharing raw data. Among the challenges associated with
heterogeneous participants, this study focuses on two complementary
forms: benign heterogeneity, where honest operators observe different
operating conditions and fault modes, and adversarial heterogeneity,
where a compromised operator submits poisoned updates.
We present a controlled, safety-oriented study of how federated
training and aggregation algorithms behave under both, using a
multi-task one-dimensional convolutional neural network on a
structurally non-independent-and-identically-distributed (non-IID)
partition of the Commercial Modular Aero-Propulsion System Simulation
(C-MAPSS) benchmark, and evaluate four benign-heterogeneity remedies
together with a five-attack by four-aggregator matrix that includes a
physically motivated, failure-masking sensor-value backdoor.
Shared-representation personalization closes about 70\% of the
local-to-centralized root-mean-square-error gap, versus about 21\%
for proximal regularization and 10\% for server-side reweighting.
Crucially, the backdoor reaches 94.9\% attack success against standard
averaging while clean accuracy stays statistically unchanged:
accuracy alone cannot certify a safe model, and attack success must be
measured explicitly. Robust aggregation with Krum cuts attack success
by an order of magnitude and is the only evaluated aggregator to
survive coordinated attackers, whereas personalization alone offers no
protection. Stacking the two yields a composed defense that restores
robustness (attack
success 2.8\%) at only a small accuracy cost, reflecting an inherent
tension between robust update selection and collaborative
representation learning. Findings hold across client counts and a
harder six-condition dataset, and all code and data partitions are
released for reproducibility.
\end{abstract}


\maketitle


\section{Introduction}
\label{sec:intro}

Aircraft-engine prognostics, predicting the Remaining Useful Life
(RUL) of a turbofan from its sensor history, is a safety-critical
reliability problem with an asymmetric cost structure: a missed
imminent failure can cost lives, whereas an unnecessary grounding
merely costs money. On the most widely used public benchmark, NASA's
Commercial Modular Aero-Propulsion System Simulation (C-MAPSS)
turbofan dataset \citep{Saxena2008CMAPSS}, data-driven RUL models
trained on run-to-failure trajectories have matured to the point
where the \emph{modelling} problem is largely solved for a single
operator that owns all of its data.

Real deployment is not single-operator. An airline consortium or
Maintenance-Repair-Overhaul (MRO) network is intrinsically
federated: each operator owns only a partial fleet, treats its
sensor telemetry as competitive intelligence, and cannot pool raw
data with rival operators or with the airframer without contractual
and regulatory friction. Federated Learning (FL)
\citep{McMahan2017FedAvg} resolves this tension: each operator trains
locally, only model weights cross the wire, and a central server
aggregates them into a global model trained, in effect, on the union
of every fleet. On C-MAPSS with statistically homogeneous clients
(drawn independently and identically from one subset), vanilla
FedAvg already matches centralized training within the noise floor
\citep{Barbosa2025FLJetEngines,Vermelin2024CollabFLRUL,%
Pandhare2021FederatedBaseline}.

That success is fragile, because it assumes clients that are both
statistically similar and uniformly honest, and neither survives
contact with a real multi-operator federation. Two distinct
departures break it, and they call for different remedies: honest
operators whose fleets follow structurally different fault
distributions (\emph{benign} heterogeneity), and compromised
operators that do not train honestly (\emph{adversarial}
heterogeneity). We treat these as two orthogonal axes and study
both, on the same C-MAPSS federation.

The benign axis arises even when every operator is honest. Operators
fly different engine variants, follow different maintenance regimes,
and expose their fleets to different mission profiles, so each local
dataset is a different projection of the joint fault distribution,
what we call \emph{benign heterogeneity}: honest but structurally
non-IID clients. C-MAPSS reproduces this when clients hold different
subsets, for example FD001 (a single fault mode: high-pressure
compressor, HPC, degradation) versus FD003 (two fault modes: HPC
plus fan degradation).

This is enough to erase the benefit of federating. On the 4-client
FD001+FD003 partition we study, averaging weights across four honest
clients recovers essentially none of the improvement over each
client training in isolation on 50 engines
(Section~\ref{sec:axis1-results}). The general FL literature offers
three families of remedy, each blaming a different cause:
server-side reweighting (the aggregator weights clients wrongly),
proximal regularization (clients drift during local training), and
architectural personalization (one shared decision head cannot span
several fault-mode families at once). To our knowledge, no prior
C-MAPSS FL study places all three families of remedy on the same
structural-non-IID setup.

The second axis is a client that does not train honestly. Attacks on
federated learning (and the Byzantine-robust defenses that counter
them) have grown into a rapidly expanding research area
\citep{Nguyen2024BackdoorSurvey}: a client that controls its local
pipeline can flip labels, scale its gradient, or implant a targeted
backdoor \citep{Bhagoji2019AdversarialLens,Bagdasaryan2020BackdoorFL,%
Xie2020DBA}. Yet this work is developed almost entirely on image and
generic-classification benchmarks, while the FL-for-prognostics
literature remains almost entirely benign. The two 2026 C-MAPSS studies that do
consider attacks each pair a single novel aggregator with a single
attack family: BioMutFed+ \citep{Tallat2026BioMutFedPlus} tests a
mutation-driven aggregator against 20\%-malicious gradient ascent,
and a trustworthy-FL-for-IIoT study \citep{Li2026TrustworthyFLIIoT}
combines blockchain reputation with gradient-magnitude clipping
against magnitude scaling. Neither considers targeted backdoors with
physically motivated sensor-value triggers, coordinated Byzantine
attackers, or the interaction between personalization and adversarial
robustness.

Adversarial heterogeneity is also qualitatively different from the
benign case, spanning two regimes that demand different defenses.
Loud attacks are conspicuous but catastrophic: a gradient-scaling
attack diverges the global model to RMSE $\sim$84 deterministically
across every seed. Stealthy attacks are the real danger: a
physically-motivated sensor-value backdoor holds its clean-set
metrics within seed variance of an honest baseline while
compromising the fault-classification head at $\sim$95\% attack
success, invisible to any monitor that inspects only clean data.
And when just two of four clients collude, half of the canonical
Byzantine-robust aggregators (trimmed mean, coordinate median) fail
as completely as no defense at all.

We treat benign and adversarial client heterogeneity as two
orthogonal axes that any deployed FL prognostics pipeline must
handle jointly. To the best of our knowledge, this is the first work
to study both axes together on federated turbofan prognostics and to
bridge them with a single composed defense. We make four contributions.

\begin{enumerate}
    \item \emph{Heterogeneity.} We conduct a controlled comparison of
    aggregation reweighting, proximal regularization, personalized
    representation learning, and clustered federated learning under
    fault-mode-based client heterogeneity on the C-MAPSS turbofan
    benchmark. The comparison shows that architectural personalization
    is substantially more effective than aggregation-level or
    optimization-level corrections in this setting.

    \item \emph{Security evaluation.} We conduct a systematic
    five-attack by four-defense evaluation of federated turbofan
    prognostics and show that a failure-masking sensor-value backdoor
    can achieve high attack success while leaving clean RUL
    performance largely unchanged.

    \item \emph{Evaluation principle.} We show that clean predictive
    metrics alone are insufficient for assessing federated prognostic
    trustworthiness, and that attack-success evaluation is necessary
    for detecting targeted failure-masking behaviour.

    \item \emph{Joint deployment trade-off.} We evaluate the
    combination of personalized representation learning and robust
    aggregation, showing that personalization alone does not provide
    backdoor resistance and that the combined approach improves
    robustness at a measurable predictive-performance cost.
\end{enumerate}

The remainder of this paper is organized as follows.
Section~\ref{sec:related} surveys related work along both axes.
Section~\ref{sec:system} defines the dataset, system model, and one
FL communication round. Sections~\ref{sec:axis1-method} and
\ref{sec:axis2-method} describe the methodology for the benign and
adversarial axes respectively. Section~\ref{sec:setup} gives
evaluation metrics and the experimental setup.
Sections~\ref{sec:axis1-results} and \ref{sec:axis2-results} present
Axis-1 and Axis-2 results respectively. Section~\ref{sec:bridge}
reports the cross-axis bridge experiment. Section~\ref{sec:discuss}
synthesizes the two axes into deployment guidance and discusses
limitations. Section~\ref{sec:conclusion} concludes.


\section{Related work}
\label{sec:related}

Federated learning for aircraft-engine prognostics is a young but
fast-growing field. The FL-for-RUL literature has matured quickly on
the modelling side, and the parallel literature on backdoor attacks
and Byzantine-robust defenses is growing at least as fast
\citep{Nguyen2024BackdoorSurvey}, yet the two have barely met. Almost
every FL-for-prognostics study evaluates clean predictive accuracy
under an implicitly honest fleet, while the attack-and-defense work
lives on image and generic-classification benchmarks. The result is a
safety-critical blind spot: a clean-accuracy evaluation cannot see a
failure-masking backdoor, and no work has yet brought both benign and
adversarial client heterogeneity together on the same turbofan
federation. This section documents that gap. We survey, in turn,
(i)~federated learning for aircraft-engine RUL prognostics,
(ii)~methods for handling benign client heterogeneity,
(iii)~methods for handling adversarial client heterogeneity,
(iv)~time-series and physical-world backdoor triggers, and
(v)~work at the intersection of personalized FL and adversarial
pressure, closing with a positioning summary
(Table~\ref{tab:positioning}).

\subsection{Federated learning for aircraft-engine RUL}
\label{subsec:related-fl-rul}

FL applications to C-MAPSS are recent and consistently
\emph{benign} in threat model.
\citet{Barbosa2025FLJetEngines} apply vanilla FedAvg with a shallow
regressor on FD001. \citet{Vermelin2024CollabFLRUL} provide the
most thorough benign benchmarking, comparing FedAvg to local-only
training across all four C-MAPSS subsets.
\citet{Pandhare2021FederatedBaseline} introduce collaborative
prognostics for machine fleets with a structural non-IID split
by operating conditions, the closest precedent for the
FD001+FD003 partition used here, but without any attack
analysis. \citet{Milasheuski2026GenerativeFL} study generative
FL (VAE / GAN / diffusion) for predictive maintenance and mention
backdoor defenses only in related work. \citet{Rehman2021TrustFed}
propose TrustFed, a reputation-based client-selection framework
tested on turbofan data; their threat model addresses free-riding
clients rather than gradient-space attackers. Closest to the
adversarial setting, \citet{Landau2025CollabRUL} pair FL for
aircraft-engine RUL with robust aggregation policies (including a
Blanchard/Krum-style best-model selection) on the N-CMAPSS
benchmark; crucially, their robustness target is \emph{noisy} honest
clients, not a malicious adversary, and they report neither backdoors
nor attack-success metrics. Beyond deep regressors,
\citet{Jeong2025FedJoint} federate a joint Gaussian-process
degradation and survival model (Fed-Joint) on turbofan data,
illustrating that FL-based RUL spans statistical as well as deep
approaches. Across this cluster the threat model is uniformly benign:
robustness, where considered, targets noise or free-riding rather
than a client that poisons the model on purpose.

\subsection{Handling benign client heterogeneity (Axis 1)}
\label{subsec:related-axis1}

\paragraph{Personalization.}
\citet{Collins2021FedRep} introduce FedRep, which trains a shared
encoder collaboratively while giving each client its own head
trained locally. \citet{Kairouz2021Advances} survey the broader
personalization family (per-client final layers, meta-learning,
split learning). Clustered federated learning
\citep{Sattler2020CFL} groups clients by update similarity and
trains a per-cluster model; the FedCCFA variant used in this
paper (Section~\ref{sec:axis1-method}) applies cosine-similarity
clustering to head deltas in a Sattler-style pipeline. In the
prognostics domain, personalization has been used for
fault-diagnosis \emph{classification}, but direct FedRep /
FedCCFA comparisons on \emph{regression-heavy} RUL prediction
are rare. Closer to our domain, \citet{Arunan2023MatchedFeatureFL}
propose feature-similarity-matched aggregation for heterogeneous
edge devices, reporting large gains over naive averaging on battery
and turbofan prognostics, a middle ground between plain FedAvg and
full head personalization, but again without any adversarial
component.

\paragraph{Proximal regularization.}
\citet{Li2020FedProx} propose FedProx, adding a proximal term to
each client's local objective to bound local drift under
statistical heterogeneity. FedProx has become the canonical
benchmark for optimization-side handling of non-IID data.

\paragraph{Server-side reweighting.}
Various schemes reweight client updates by metrics such as
validation performance, loss reduction, or class-balance
diagnostics. These are cheaper than personalization (no
per-client state) but generally provide smaller gains under
structural non-IID, a finding we quantify in
Section~\ref{sec:axis1-results}. \citet{Berghout2022FLCondMon}
survey the broader FL-for-condition-monitoring space and observe
that most prior work focuses on classification-style fault
diagnosis rather than regression-based RUL.

\subsection{Handling adversarial client heterogeneity (Axis 2)}
\label{subsec:related-axis2}

\paragraph{Robust aggregators.}
The canonical Byzantine-robust aggregators are Krum
\citep{Blanchard2017Krum}, coordinate median and trimmed mean
\citep{Yin2018RobustDistributed}, and robust functional
aggregation via the geometric median \citep{Pillutla2022RFA}.
Norm-clipping partial defenses \citep{Sun2019CanYouBackdoor}
argue that bounded update norms alone defeat many backdoor
variants; we return to this claim in Section~\ref{sec:discuss}.

\paragraph{Attacks.}
Untargeted attacks include label-flip and gradient scaling
(``model poisoning'')
\citep{Bhagoji2019AdversarialLens, Fang2020LocalPoisoning}.
Targeted attacks include backdoors
\citep{Bagdasaryan2020BackdoorFL} and their distributed or
coordinated variants \citep{Xie2020DBA, Lyu2025CoBA}. The
EAAI survey of \citet{Nguyen2024BackdoorSurvey} documents the rapid
growth of this literature and taxonomizes defenses by phase
(pre-, in-, and post-aggregation); in that taxonomy our sensor-value
backdoor is a data-poisoning attack met by an in-aggregation defense
(Krum). Theoretically, \citet{Farhadkhani2022Equivalence} prove an
equivalence between data poisoning and Byzantine gradient attacks and
derive impossibility results for robust learning under high client
heterogeneity, grounding both our two-axis framing and the tension
we observe empirically between robustness and structural non-IID.

\paragraph{FL for {IIoT} + attacks.}
\citet{Li2023ByzantineFLIIoT} provide the cornerstone
Byzantine-robust FL benchmark in an IIoT setting on generic
classification tasks. \citet{Hou2022FederatedFiltersIIoT}
propose federated filters against image-like backdoors in IIoT.
\citet{Djemaa2026HeterogeneityPoisoning} taxonomize FL poisoning
attacks under statistical heterogeneity, directly supporting the
``structural non-IID $+$ attack'' positioning of this work. Two
2026 papers directly overlap with the present study:
\citet{Tallat2026BioMutFedPlus} test a mutation-driven aggregator
on C-MAPSS against 20\%-malicious gradient ascent, and
\citet{Li2026TrustworthyFLIIoT} use C-MAPSS to evaluate
blockchain reputation plus gradient magnitude clipping against
magnitude scaling. Both are single-attack / single-defense; the
$5 \times 4$ matrix study reported here is genuinely orthogonal.

\subsection{Time-series and physical-world backdoor triggers}
\label{subsec:related-triggers}

The FL-backdoor literature has largely explored image-domain
triggers: patches \citep{Bagdasaryan2020BackdoorFL}, distributed
patch fragments \citep{Xie2020DBA}, semantic triggers
\citep{Bhagoji2019AdversarialLens}, and boundary-set
constructions \citep{Yang2023BoundaryTrigger}. Recent
collusive-trigger work \citep{Lyu2025CoBA} learns triggers
jointly across attackers. BADControl
\citep{Burbano2026BADControl} introduces the
\emph{physical-trigger} threat model for cyber-physical control
systems, analogous in spirit to the sensor-value trigger of
Section~\ref{sec:axis2-method}, but for direct RL-based control
rather than FL-based prognostics. Orthogonally to trigger
\emph{placement} in the input, \citet{Foroughi2026LSA} show that
poisoning only a few backdoor-critical \emph{layers} yields up to
97\% backdoor success while preserving clean accuracy and
\emph{bypassing} Multi-Krum, trimmed mean, and FLAME, a direct
demonstration that accuracy is blind to backdoors, and a caution that
the geometry-based Krum filter we adopt would likely not survive such
an adaptive, layer-aware adversary (Section~\ref{sec:discuss}).
To the best of our knowledge,
no prior work has proposed a physically-motivated sensor-value
trigger for FL-based prognostic regression on a turbofan
benchmark.

\subsection{Cross-axis: personalized FL under adversarial pressure}
\label{subsec:related-crossaxis}

Several recent papers explore the interaction between
personalization (an Axis-1 remedy) and adversarial pressure (an
Axis-2 threat). \citet{Zhang2024SARS} propose a personalized FL
framework designed against backdoors, using FedRep as a
comparison baseline. \citet{Fan2026RobustPFL} show personalized FL
to be \emph{more} vulnerable than centralized learning, but to
test-time \emph{evasion} (adversarial examples), a distinct threat
from the training-time \emph{backdoor} we study, so the two findings
are complementary. Composing robustness with personalization is not
itself new: \citet{Pillutla2022RFA} already pair geometric-median
aggregation with an on-device personalization variant. Our
contribution is therefore not the \emph{idea} of composition but its
\emph{domain} (aeroengine RUL regression) and a specific
encoder-only-Krum instantiation, together with the finding, in
contrast to the partial-shielding intuition in the vision-domain
literature, that per-client heads do \emph{not} shield honest clients
when the poison acts on the shared representation
(Section~\ref{sec:bridge}). All prior work in this cluster operates
on image / classification datasets (CIFAR, MNIST, FEMNIST); none
report on prognostic \emph{regression} heads.

\subsection{Positioning summary}
\label{subsec:related-positioning}

Table~\ref{tab:positioning} places the present study alongside
the closest existing work. The nearest neighbour on the
\emph{application} side is \citet{Landau2025CollabRUL}, who pair
aircraft-engine FL RUL with robust aggregation, but for
\emph{noisy}, not adversarial, clients. The only C-MAPSS studies that
consider adversarial threats are the two 2026 papers
\citep{Tallat2026BioMutFedPlus, Li2026TrustworthyFLIIoT}, each
of which evaluates a single novel aggregator against a single
attack family. None of the prior work jointly considers Axis 1
and Axis 2, none pairs personalization with robust aggregation, and
none reports multi-seed variance.

\begin{table}[pos=tbp]
\centering
\footnotesize
\setlength{\tabcolsep}{4pt}
\renewcommand{\arraystretch}{1.2}
\caption{Positioning of this work against the closest prior
studies on federated learning for aircraft-engine prognostics
and adjacent domains. ``MS'' = multi-seed reporting
($\geq 3$ independent random seeds with variance disclosed);
``op.\ cond.''\ = operating conditions;
``IA rew.''\ = imbalance-aware reweighting.
For this work the 5-attack matrix comprises label-flip,
gradient scaling ($\times{-10}$ and $\times{-2}$), a
sensor-value backdoor, and a coordinated 2-of-4 Byzantine
attack; the 4-defense set is FedAvg, trimmed mean, coordinate
median, and Krum ($f = 1$).}
\label{tab:positioning}
\zebra
\begin{tabular}{@{}>{\raggedright\arraybackslash}p{3.6cm} l l >{\raggedright\arraybackslash}p{2.5cm} >{\raggedright\arraybackslash}p{2.8cm} c@{}}
\toprule
\tblhead
Study & Domain & Non-IID & Axis 1 methods & Axis 2 methods & MS \\
\midrule
\citet{Barbosa2025FLJetEngines}       & C-MAPSS  & IID          & FedAvg              & ---                                     & No \\
\citet{Vermelin2024CollabFLRUL}       & C-MAPSS  & benign       & FedAvg              & ---                                     & No \\
\citet{Pandhare2021FederatedBaseline} & Fleets   & op.\ cond.   & fleet baseline      & ---                                     & No \\
\citet{Landau2025CollabRUL}           & N-CMAPSS & benign       & FedAvg              & robust agg.\ (noise)                    & No \\
\citet{Rehman2021TrustFed}            & C-MAPSS  & benign       & ---                 & reputation                              & No \\
\citet{Li2023ByzantineFLIIoT}         & IIoT     & mild         & ---                 & median, Krum                            & No \\
\citet{Hou2022FederatedFiltersIIoT}   & IIoT     & mild         & ---                 & federated filters                       & No \\
\citet{Tallat2026BioMutFedPlus}       & C-MAPSS  & mild         & ---                 & mutation-driven agg.                    & No \\
\citet{Li2026TrustworthyFLIIoT}       & C-MAPSS  & mild         & ---                 & blockchain rep.\ $+$ norm clipping      & No \\
\midrule
\textbf{This work} & \textbf{C-MAPSS} & \textbf{fault-mode} & \textbf{FedProx, FedRep, FedCCFA, IA rew.} & \textbf{trim.\ mean, median, Krum ($f{=}1$); 5-attack matrix} & \textbf{Yes} \\
\bottomrule
\end{tabular}
\end{table}


\section{System model, dataset, and the two-axis frame}
\label{sec:system}

\subsection{Dataset: NASA C-MAPSS turbofan benchmark}
\label{subsec:dataset}

NASA's Commercial Modular Aero-Propulsion System Simulation
(C-MAPSS) \citep{Saxena2008CMAPSS} is the canonical benchmark for
aircraft-engine prognostics. It comprises four subsets, each
simulating a fleet of turbofan engines run to failure under a
controlled combination of operating conditions and fault-mode
assumptions. Together the four subsets contain 709 training engines,
approximately 160{,}000 sliding-window training samples, 21 raw
sensor channels, and (by construction) zero missing values.
The subsets differ along two axes: single vs.\ multiple operating
conditions, and single vs.\ multiple fault modes:
\emph{FD001} (1 condition, HPC degradation), \emph{FD002}
(6 conditions, HPC), \emph{FD003} (1 condition, HPC + fan
degradation), and \emph{FD004} (6 conditions, HPC + fan).

\paragraph{Why FD001+FD003 specifically.}
Prior FL work on C-MAPSS either uses IID splits (four clients drawn
i.i.d.\ from a single subset) or operating-condition non-IID
\citep{Pandhare2021FederatedBaseline}, where different clients see
different flight regimes but the same fault-mode distribution. The
FD001 $+$ FD003 partition used here isolates a \emph{second} layer
of heterogeneity on top: FD001 clients see one fault mode
(HPC degradation) while FD003 clients see two fault modes (HPC $+$
fan degradation). Operating conditions are deliberately held
constant across the two subsets, so any observed FL failure is
attributable to fault-mode divergence rather than confounded by
input-distribution shift. This design provides a controlled stress
test of fault-mode heterogeneity as a distinct experimental
condition, complementary to the operating-condition heterogeneity
previously studied in the C-MAPSS FL literature. FD001~$+$~FD003 is
our \emph{primary} controlled setting; to test whether the findings
survive greater difficulty, we additionally replicate the study on
the harder six-condition subsets FD002~$+$~FD004
(Sections~\ref{sec:axis1-results}--\ref{sec:bridge}), which layer
operating-condition complexity on top of the same fault-mode split.

\paragraph{Windowing and labelling.}
Sliding windows of length $W = 30$ cycles are extracted per engine
with stride~1, matching common practice on C-MAPSS. The RUL
regression target is capped at $R_{\max} = 125$ cycles (healthy
engines with more than 125 cycles remaining are labelled 125),
following the standard piecewise-linear RUL convention. A binary
``fault-imminent'' label is derived from $\text{RUL} \leq 30$:
this is the label the fault-classification head predicts. Of the 21
raw C-MAPSS sensor channels, 4 are constant across the FD001+FD003
engines and are removed by feature selection, leaving
\textbf{17 informative sensors} as the model's input feature vector.

\subsection{Federation topology}
\label{subsec:federation}

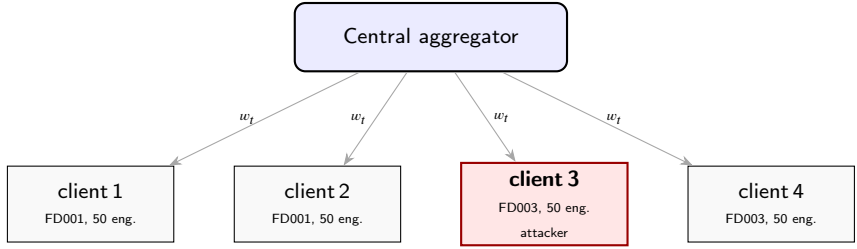
\begin{figure}[pos=htbp]
\centering
\begin{tikzpicture}[
    every node/.style={font=\footnotesize},
    server/.style={rectangle, draw, rounded corners, minimum width=3.6cm, minimum height=0.9cm, align=center, fill=blue!8, thick},
    client/.style={rectangle, draw, minimum width=2.2cm, minimum height=1cm, align=center, fill=gray!5},
    attacker/.style={rectangle, draw, minimum width=2.2cm, minimum height=1cm, align=center, fill=red!10, draw=red!60!black, thick},
    arrow/.style={-Stealth, thin, draw=gray!70}
]
    \node[server] (S) at (0, 2.2) {Central aggregator};
    \node[client] (C1) at (-4.5, 0) {client\,1 \\ {\tiny FD001, 50 eng.}};
    \node[client] (C2) at (-1.5, 0) {client\,2 \\ {\tiny FD001, 50 eng.}};
    \node[attacker] (C3) at (1.5, 0) {\textbf{client\,3} \\ {\tiny FD003, 50 eng.} \\ {\tiny attacker}};
    \node[client] (C4) at (4.5, 0) {client\,4 \\ {\tiny FD003, 50 eng.}};
    \draw[arrow] (S) -- (C1) node[midway, left, font=\tiny] {$w_t$};
    \draw[arrow] (S) -- (C2) node[midway, left, font=\tiny] {$w_t$};
    \draw[arrow] (S) -- (C3) node[midway, right, font=\tiny] {$w_t$};
    \draw[arrow] (S) -- (C4) node[midway, right, font=\tiny] {$w_t$};
\end{tikzpicture}
\caption{Federation topology and threat surface. Four clients communicate with a central aggregator; two of them hold FD001 data (single fault mode: HPC degradation) and two hold FD003 data (two fault modes: HPC~$+$~fan). Client~3 is the default single attacker in the Axis~2 experiments; the coordinated 2-attacker case additionally makes client~4 malicious. Downward arrows show the weight broadcast $w_t$; client update messages $\delta_k$ (not drawn) flow in the opposite direction.}
\label{fig:federation}
\end{figure}

A 4-client federation is instantiated using the FD001+FD003
partition. Each subset's 100 training engines are split evenly
between two clients (50 engines per client), giving four clients
total (Figure~\ref{fig:federation}). Under this partition, the
FD001 clients see a single fault-mode density while the FD003
clients see a two-component mixture, the concrete instantiation
of Axis~1 benign heterogeneity. For Axis~2 adversarial experiments,
client~3 is the default single attacker; the coordinated 2-attacker
cell of Section~\ref{sec:axis2-method} additionally makes client~4
malicious.

\paragraph{Why $N = 4$?}
The choice $N = 4$ is motivated by three considerations that
together define the controlled small-federation regime of this
study: (i)~parity with prior C-MAPSS FL work
\citep{Barbosa2025FLJetEngines, Vermelin2024CollabFLRUL,
Pandhare2021FederatedBaseline}, which uses similar client counts;
(ii)~computational tractability of the $5 \times 4$ attack $\times$
aggregator matrix (Section~\ref{sec:axis2-results}); and
(iii)~exposure of the small-$N$ regime where
Byzantine-robust aggregators such as Krum operate near their
formal feasibility boundary (Section~\ref{subsec:krum}).

The four-client setting is therefore a \emph{controlled
experimental scale}, not a claim about any specific commercial
federation size. To confirm the conclusions are not artefacts of
this scale, we additionally replicate the adversarial study at
$N = 6$ (Section~\ref{sec:axis2-results}), which also unlocks the
$f = 2$ Krum configuration that is formally undefined at $N = 4$.

\subsection{Multi-task 1-D CNN architecture}
\label{subsec:model}

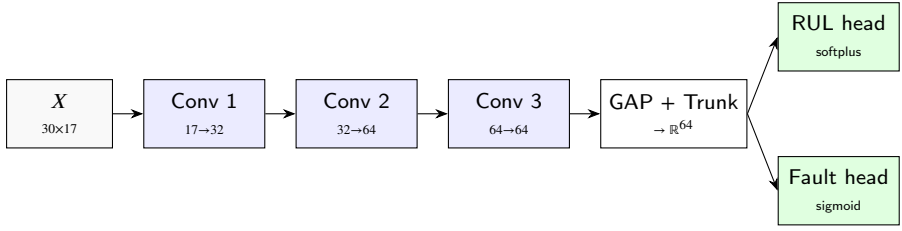
\begin{figure}[pos=htbp]
\centering
\begin{tikzpicture}[
    node distance=3mm and 4mm,
    every node/.style={font=\footnotesize},
    input/.style={rectangle, draw, minimum width=1.4cm, minimum height=0.9cm, align=center, fill=gray!5},
    conv/.style={rectangle, draw, minimum width=1.6cm, minimum height=0.9cm, align=center, fill=blue!8},
    pool/.style={rectangle, draw, minimum width=1.6cm, minimum height=0.9cm, align=center},
    head/.style={rectangle, draw, minimum width=1.6cm, minimum height=0.9cm, align=center, fill=green!12},
    arrow/.style={-Stealth, thin}
]
    \node[input] (X) {$X$ \\ {\tiny $30 {\times} 17$}};
    \node[conv, right=of X] (B1) {Conv~1 \\ {\tiny $17 {\to} 32$}};
    \node[conv, right=of B1] (B2) {Conv~2 \\ {\tiny $32 {\to} 64$}};
    \node[conv, right=of B2] (B3) {Conv~3 \\ {\tiny $64 {\to} 64$}};
    \node[pool, right=of B3] (GAP) {GAP $+$ Trunk \\ {\tiny $\to \mathbb{R}^{64}$}};
    \node[head, above right=1mm and 4mm of GAP] (HR) {RUL head \\ {\tiny softplus}};
    \node[head, below right=1mm and 4mm of GAP] (HF) {Fault head \\ {\tiny sigmoid}};
    \draw[arrow] (X) -- (B1);
    \draw[arrow] (B1) -- (B2);
    \draw[arrow] (B2) -- (B3);
    \draw[arrow] (B3) -- (GAP);
    \draw[arrow] (GAP.east) -- (HR.west);
    \draw[arrow] (GAP.east) -- (HF.west);
\end{tikzpicture}
\caption{Multi-task 1-D CNN architecture (30{,}018 parameters). Each convolutional block is Conv1D~$\to$~GroupNorm~$\to$~ReLU~$\to$~MaxPool; kernels are size 5 for the first two blocks and size 3 for the third. Global-average pooling collapses the temporal dimension; the shared trunk feeds two task-specific linear heads.}
\label{fig:architecture}
\end{figure}

The shared model $f_\theta$ is a multi-task 1-D CNN
(Figure~\ref{fig:architecture}) with three convolutional blocks
(each Conv1D $\to$ GroupNorm $\to$ ReLU $\to$ MaxPool), a
global-average-pooling head, a shared trunk, and two task heads.

\paragraph{Why GroupNorm and not BatchNorm.}
GroupNorm is used in preference to BatchNorm because BatchNorm's
running statistics are distribution-dependent and unsafe under FL
heterogeneity: two clients with different fault-mode densities will
accumulate systematically different running mean and
variance statistics, and averaging these across clients
corrupts the normalization layer even when all clients are honest.
GroupNorm has no running statistics and is FL-safe by construction.
A regression test in our codebase asserts that no BatchNorm layer
ever enters the model.

\paragraph{Parameter budget.}
Total parameter count: \textbf{30{,}018} (Table~\ref{tab:params}).
The model is intentionally small so that FL round times remain in
the single-digit seconds and the compute per cell of the
$5 \times 4$ Axis-2 matrix stays tractable.

\paragraph{Joint loss.}
The two heads are trained jointly with
\begin{equation}
    \mathcal{L}(\theta) \;=\; \mathcal{L}_{\mathrm{RUL}}(\theta) \;+\; \lambda \cdot \mathcal{L}_{\mathrm{fault}}(\theta),
    \qquad \lambda = 0.5,
    \label{eq:loss}
\end{equation}
where $\mathcal{L}_{\mathrm{RUL}}$ is Huber loss on the regression
head and $\mathcal{L}_{\mathrm{fault}}$ is binary cross-entropy on
the classification head. Multi-task learning imposes a useful
inductive bias: RUL and fault-imminent are both functions of the
same underlying degradation state, so sharing the encoder forces it
to learn features relevant to both tasks.

\begin{table}[pos=tbp]
\centering
\footnotesize
\setlength{\tabcolsep}{6pt}
\renewcommand{\arraystretch}{1.1}
\caption{Parameter budget for the multi-task 1-D CNN of Figure~\ref{fig:architecture}. All convolutions use \texttt{padding = same} so the window length is preserved through the encoder.}
\label{tab:params}
\zebra
\begin{tabular}{@{}l r r@{}}
\toprule
\tblhead
Layer & Parameters & Cumulative \\
\midrule
Conv1D$(17 \to 32,\ k{=}5)$ $+$ bias & 2{,}752 & 2{,}752 \\
GroupNorm$(32)$                       & 64      & 2{,}816 \\
Conv1D$(32 \to 64,\ k{=}5)$ $+$ bias & 10{,}304 & 13{,}120 \\
GroupNorm$(64)$                       & 128     & 13{,}248 \\
Conv1D$(64 \to 64,\ k{=}3)$ $+$ bias & 12{,}352 & 25{,}600 \\
GroupNorm$(64)$                       & 128     & 25{,}728 \\
Linear$(64 \to 64)$ trunk $+$ bias   & 4{,}160  & 29{,}888 \\
Linear$(64 \to 1)$ RUL head $+$ bias & 65       & 29{,}953 \\
Linear$(64 \to 1)$ fault head $+$ bias & 65     & \textbf{30{,}018} \\
\bottomrule
\end{tabular}
\end{table}

\subsection{The two-axis heterogeneity frame}
\label{subsec:two-axis-frame}

\begin{figure}[pos=htbp]
\centering
\begin{tikzpicture}[
    every node/.style={font=\footnotesize},
    header/.style={font=\small\bfseries, align=center},
    problem/.style={rectangle, draw, rounded corners, minimum width=5cm, minimum height=0.9cm, align=center, fill=yellow!12},
    remedy/.style={rectangle, draw, minimum width=5cm, minimum height=0.55cm, align=center, fill=gray!5, font=\scriptsize}
]
    \node[header] (H1) at (0, 3) {Axis 1: Benign heterogeneity};
    \node[problem] (P1) at (0, 2.1) {Honest clients hold different \\ data distributions};
    \node[remedy] (R1a) at (0, 1.15) {Personalization (FedRep, FedCCFA)};
    \node[remedy] (R1b) at (0, 0.5) {Proximal regularization (FedProx)};
    \node[remedy] (R1c) at (0, -0.15) {Server-side reweighting};
    \node[header] (H2) at (6.5, 3) {Axis 2: Adversarial heterogeneity};
    \node[problem] (P2) at (6.5, 2.1) {Some clients deviate from \\ honest training};
    \node[remedy] (R2a) at (6.5, 1.15) {Byzantine-robust aggregation};
    \node[remedy] (R2b) at (6.5, 0.5) {(trim.\ mean, coord.\ median, Krum)};
    \draw[thin, gray!50, dashed] (3.25, 3.3) -- (3.25, -0.5);
\end{tikzpicture}
\caption{The two orthogonal axes of client heterogeneity that any deployed FL prognostics pipeline must handle. Axis~1 (left) is benign but non-IID data distributions across honest clients; Axis~2 (right) is clients that deviate from honest training. Each axis admits distinct families of remedies. Section~\ref{sec:bridge} shows empirically that remedies for one axis do not confer protection on the other.}
\label{fig:two-axis}
\end{figure}
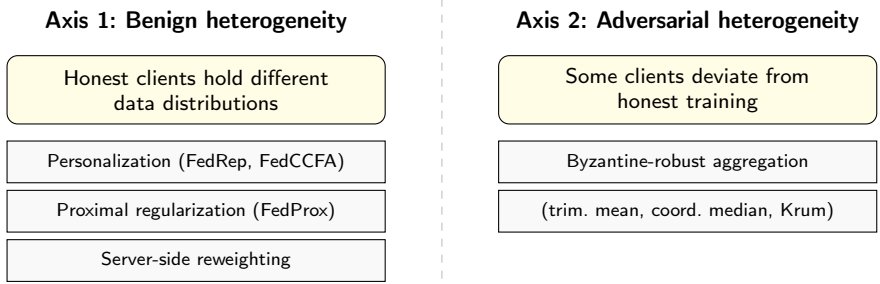

Client heterogeneity in any federated network of operators arises
along two orthogonal axes (Figure~\ref{fig:two-axis}).
\emph{Axis~1 (benign)} captures the fact that honest clients can
still differ in their local data distributions: different operating
conditions, different fault-mode mixtures, different maintenance
regimes. \emph{Axis~2 (adversarial)} captures the fact that some
clients may not train honestly at all. These two kinds of heterogeneity require
completely different remedies; the vast bulk of the
FL-for-prognostics literature addresses only one of them. This paper
addresses both jointly on the same federation.

\subsection{One FL communication round}
\label{subsec:one-round}

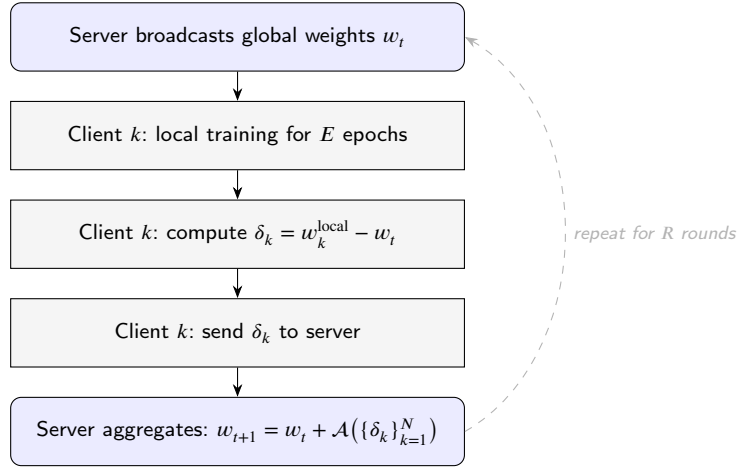
\begin{figure}[pos=htbp]
\centering
\begin{tikzpicture}[
    every node/.style={font=\footnotesize},
    server/.style={rectangle, draw, rounded corners, minimum width=6cm, minimum height=0.9cm, align=center, fill=blue!8},
    client/.style={rectangle, draw, minimum width=6cm, minimum height=0.9cm, align=center, fill=gray!8},
    arrow/.style={-Stealth, thin}
]
    \node[server] (S1) at (0, 0) {Server broadcasts global weights $w_t$};
    \node[client] (C1) at (0, -1.3) {Client $k$: local training for $E$ epochs};
    \node[client] (C2) at (0, -2.6) {Client $k$: compute $\delta_k = w^{\mathrm{local}}_k - w_t$};
    \node[client] (C3) at (0, -3.9) {Client $k$: send $\delta_k$ to server};
    \node[server] (S2) at (0, -5.2) {Server aggregates: $w_{t+1} = w_t + \mathcal{A}\bigl(\{\delta_k\}_{k=1}^N\bigr)$};
    \draw[arrow] (S1) -- (C1);
    \draw[arrow] (C1) -- (C2);
    \draw[arrow] (C2) -- (C3);
    \draw[arrow] (C3) -- (S2);
    \draw[arrow, dashed, gray!60] (S2.east) to[bend right=60] node[midway, right, font=\scriptsize\itshape] {repeat for $R$ rounds} (S1.east);
\end{tikzpicture}
\caption{One synchronous FL communication round. Server-side steps are shaded blue and client-side steps grey. The aggregation rule $\mathcal{A}$ is sample-count-weighted mean for FedAvg (baseline) and takes different forms for the Axis-1 methods of Section~\ref{sec:axis1-method} and the Axis-2 robust aggregators of Section~\ref{sec:axis2-method}. The primary experiments use $N = 4$ clients and $R = 50$ rounds; the generalization study of Section~\ref{sec:axis2-results} additionally uses $N = 6$.}
\label{fig:round}
\end{figure}

Every method evaluated in this paper follows the same synchronous
round-based FL loop (Figure~\ref{fig:round}). The differences among
methods lie in (i)~the client-side local update rule and (ii)~the
server-side aggregation rule.

All clients participate in every round (no client sampling). Each
client runs $E = 2$ local epochs of Adam
(learning rate $10^{-3}$ with a cosine schedule, weight decay
$10^{-4}$, batch size 256). The federation runs for
$R = 50$ communication rounds; the best round per experiment is
selected on a held-out combined test set using the asymmetric
NASA scoring function of Section~\ref{sec:setup}.

\section{Methodology: Axis 1 (benign heterogeneity)}
\label{sec:axis1-method}

We evaluate four families of remedies against benign client
heterogeneity, each addressing a different putative cause of
vanilla FedAvg's failure under structural non-IID: server-side
reweighting (aggregation-layer intervention), proximal
regularization (client-optimization-layer intervention), and
per-client architectural personalization (client-architecture-layer
intervention, in two variants). Vanilla FedAvg is the baseline
against which every method in Sections~\ref{subsec:fedprox}
through \ref{subsec:reweight} is compared. This section defines
each method; Section~\ref{sec:axis1-results} reports the results.

\subsection{Baseline: FedAvg}
\label{subsec:fedavg}

FedAvg \citep{McMahan2017FedAvg} aggregates the $N$ client updates
$\{\delta_k\}_{k=1}^N$ obtained in Figure~\ref{fig:round} by
sample-count-weighted mean:
\begin{equation}
    \mathrm{FedAvg}\bigl(\{\delta_k\}\bigr)
    \;=\; \sum_{k=1}^{N} \frac{n_k}{\sum_{j} n_j}\, \delta_k,
    \label{eq:fedavg}
\end{equation}
where $n_k$ is client $k$'s local sample count. The updated global
weights are then $w_{t+1} = w_t + \mathrm{FedAvg}(\{\delta_k\})$.
This is the reference against which every Axis-1 remedy below is
measured.

\subsection{Proximal regularization: FedProx}
\label{subsec:fedprox}

FedProx \citep{Li2020FedProx} modifies each client's local
objective by adding a proximal term that penalises drift from the
current global model:
\begin{equation}
    \mathcal{L}_k^{\mathrm{prox}}(\theta)
    \;=\; \mathcal{L}_k(\theta)
    \;+\; \frac{\mu}{2}\, \bigl\| \theta - \theta_t^{\mathrm{global}} \bigr\|_2^2.
    \label{eq:fedprox}
\end{equation}
The server-side aggregation rule is still Equation~\eqref{eq:fedavg}
(sample-count-weighted mean). The hyperparameter $\mu \geq 0$
controls the drift penalty; $\mu = 0$ recovers FedAvg exactly and
serves as a bit-exact regression test. This paper sweeps
$\mu \in \{0, 10^{-3}, 10^{-2}, 10^{-1}\}$ and reports the best
cell (Section~\ref{sec:axis1-results}).

\subsection{Personalization: FedRep}
\label{subsec:fedrep}

FedRep \citep{Collins2021FedRep} partitions the model into a
\emph{shared encoder} $\phi$ (all convolutional blocks and the
global-average-pooling trunk) and per-client \emph{heads} $\psi_k$
(the RUL head and the fault head of
Figure~\ref{fig:architecture}). Every round the encoder is shared
but the heads are kept locally. Because our prognostic model is
multi-task, this extends the original single-head FedRep to
\emph{two} private heads, so each client personalises both its
regression and its fault-classification behaviour while
collaboratively learning the shared representation.
Algorithm~\ref{alg:fedrep} gives
the training loop for one client-round; this paper uses
$h_{\mathrm{epochs}} = 1$ and $e_{\mathrm{epochs}} = 1$.

\begin{algorithm}[!ht]
\caption{FedRep local update for client $k$ at round $t$.}
\label{alg:fedrep}
\begin{algorithmic}[1]
\Require Shared encoder $\phi_t$ received from the server; local head $\psi_k^{t-1}$ kept from the previous round; local dataset $\mathcal{D}_k$
\Statex \textbf{Phase 1: head-only training}
\State Freeze $\phi_t$; train $\psi_k$ for $h_{\mathrm{epochs}}$ on $\mathcal{D}_k$ minimising $\mathcal{L}_k(\phi_t, \psi_k)$
\State $\psi_k^t \gets$ result of the head-only training
\Statex \textbf{Phase 2: encoder-only training}
\State Unfreeze $\phi_t$; train $\phi_t$ for $e_{\mathrm{epochs}}$ on $\mathcal{D}_k$ minimising $\mathcal{L}_k(\phi_t, \psi_k^t)$
\State $\phi_k^{\mathrm{local}} \gets$ result of the encoder-only training
\Statex \textbf{Server communication}
\State Compute encoder delta: $\delta^{\mathrm{enc}}_k \gets \phi_k^{\mathrm{local}} - \phi_t$
\State \textbf{Send} $\delta^{\mathrm{enc}}_k$ to the server (keep $\psi_k^t$ local)
\Statex \textbf{Server aggregation}
\State $\phi_{t+1} \gets \phi_t + \sum_{k=1}^{N} \frac{n_k}{\sum_{j} n_j}\, \delta^{\mathrm{enc}}_k$
\end{algorithmic}
\end{algorithm}

Each client's head is initialised from the round-0 encoder's
randomly initialised heads and never leaves the client thereafter.
At evaluation time, client $k$'s full model $(\phi_t, \psi_k^t)$ is
used; the per-client performance can therefore differ across clients
even though the encoder is shared.

\paragraph{Adaptations from canonical FedRep.}
Two choices depart from the original formulation of
\citet{Collins2021FedRep}, and we state them explicitly so the method
is reproducible and its provenance is unambiguous. First, canonical
FedRep keeps a \emph{single} classification head private, whereas our
prognostic backbone is multi-task, so \emph{both} heads, the RUL
regressor and the fault classifier, are personalised: Phase~1
trains the two heads jointly with the encoder frozen, and Phase~2
freezes both heads while updating the shared encoder and trunk.
Second, canonical FedRep alternates several head-update steps against
a \emph{single} representation gradient step, whereas we run one full
local \emph{epoch} on the heads followed by one full local epoch on
the encoder and trunk
($h_{\mathrm{epochs}} = e_{\mathrm{epochs}} = 1$). This keeps the
per-round budget at the same $E = 2$ local epochs used by the FedAvg,
FedProx, and imbalance-aware baselines (Section~\ref{sec:setup}), so
no protocol is advantaged by more local computation per round;
FedRep merely spends one epoch on the heads and one on the encoder.
Because an epoch spans many minibatch steps, the shared encoder
nonetheless receives far more local optimisation per round than the
canonical single-gradient-step update. Neither change constitutes a
new algorithm: both are engineering adaptations of FedRep to
multi-task time-series prognostics, reported here so the comparison
against the original is exact.

\subsection{Clustered personalization: FedCCFA}
\label{subsec:fedccfa}

FedCCFA groups clients by pairwise similarity of their encoder
deltas and runs a separate FedRep-like aggregation \emph{within}
each cluster (Sattler-style clustered FL,
\citealp{Sattler2020CFL}). Let $S_{ij} = \cos(\delta_i, \delta_j)
\in [-1, 1]$ be the cosine similarity between two client encoder
deltas. Given a threshold $\tau \in [0, 1]$, clients $i$ and $j$
are placed in the same cluster iff $S_{ij} \geq \tau$. The server
maintains one aggregated encoder per cluster and broadcasts each
client its own cluster's encoder. Algorithm~\ref{alg:fedccfa}
sketches the round-level flow; we use $\tau = 0.5$ and
$R_{\mathrm{warmup}} = 3$ rounds of ordinary FedRep aggregation
before clustering activates.

\begin{algorithm}[!ht]
\caption{FedCCFA round at communication round $t$.}
\label{alg:fedccfa}
\begin{algorithmic}[1]
\Require Client encoder deltas $\{\delta_1, \ldots, \delta_N\}$ (obtained as in Algorithm~\ref{alg:fedrep}); similarity threshold $\tau$; warmup round count $R_{\mathrm{warmup}}$
\If{$t \leq R_{\mathrm{warmup}}$}
    \State $\phi_{t+1} \gets \phi_t + \sum_{k} \frac{n_k}{\sum_{j} n_j}\, \delta_k$ \Comment{ordinary FedRep aggregation}
\Else
    \State Compute pairwise cosine similarities $S_{ij} \gets \cos(\delta_i, \delta_j)$
    \State Build graph $G = (V, E)$ with $E = \{(i, j) : S_{ij} \geq \tau\}$
    \State Partition $V$ into clusters $\mathcal{C}_1, \ldots, \mathcal{C}_m$ via connected components on $G$
    \For{each cluster $\mathcal{C}_c$ with size $n_c = \sum_{k \in \mathcal{C}_c} n_k$}
        \State $\phi^{(c)}_{t+1} \gets \phi_t + \sum_{k \in \mathcal{C}_c} \frac{n_k}{n_c}\, \delta_k$
    \EndFor
    \State Broadcast $\phi^{(c)}_{t+1}$ to each client in $\mathcal{C}_c$
\EndIf
\end{algorithmic}
\end{algorithm}

\subsection{Server-side reweighting: imbalance-aware aggregation}
\label{subsec:reweight}

An alternative to changing client-side computation is to change the
server-side weighting. For a per-client ``health score'' $h_k$ and
a softmax temperature $T$, the aggregated update is
\begin{equation}
    \mathrm{Reweight}\bigl(\{\delta_k\}\bigr)
    \;=\; \sum_{k=1}^{N} \omega_k\, \delta_k,
    \qquad
    \omega_k \;=\; \bigl[\mathrm{softmax}(h_k / T)\bigr]_k,
    \label{eq:reweight}
\end{equation}
subject to a floor $\omega_k \geq \omega_{\min}$ that prevents
client starvation. This paper evaluates three schemes for the
health score:
\begin{itemize}
    \item \textbf{Fault-count}: $h_k = -\bigl|\, n_k^{\text{fault-positive}} - \bar{n}\bigr|$ (penalises clients whose fault-positive rate is far from the federation mean);
    \item \textbf{Inverse-loss}: $h_k = 1 / (\text{loss}_k + \varepsilon)$ (rewards clients whose training loss is low);
    \item \textbf{Validation-F1}: $h_k = F_{1,k}^{\mathrm{val}}$ (rewards clients whose held-out validation F1 is high).
\end{itemize}
We use $T = 0.5$ and $\omega_{\min} = 0.05$ throughout.


\section{Methodology: Axis 2 (adversarial heterogeneity)}
\label{sec:axis2-method}

Axis~2 evaluates whether the same FL federation can survive one or
more clients that deviate from honest training. This section
defines the threat model, five attack families (Table~\ref{tab:attacks}),
and four defense aggregators (Table~\ref{tab:defenses}). Every
attack-family / defense-aggregator pair is a cell of the
$5 \times 4$ matrix reported in Section~\ref{sec:axis2-results}.
Krum's feasibility constraint at $N = 4$ restricts us to $f = 1$
(discussed in Section~\ref{subsec:krum}); the coordinated 2-attacker
cell is therefore evaluated with Krum-$f_1$ as an empirical stress
test of the aggregator outside its formal single-Byzantine
assumption.

\subsection{Threat model}
\label{subsec:threat-model}

The adversary fully controls one or two clients (client~3 for
single-attacker cells and clients~3 and~4 for the coordinated cell),
including the local dataset and the local training code. The
adversary can observe the global model at every round but cannot
inspect honest clients' data. The server is honest and can apply
any of the aggregators in Section~\ref{subsec:defenses}, but does
\emph{not} run backdoor detection, does \emph{not} inspect training
data, and does \emph{not} treat any client as more or less
trustworthy than any other. All defense happens purely in the
gradient / model-update space that the FL protocol already exposes.

Attack goals fall into two categories. \emph{Untargeted} attacks
(AV1, AV2, AV4, AV5) aim to degrade the global model's quality on
the entire honest test set. \emph{Targeted} attacks (AV3) aim to
install a backdoor: the malicious client makes the fault-classifier
head output ``not faulty'' on trigger-stamped inputs while
preserving performance on clean data, so the attack is invisible to
any monitoring pipeline that inspects only the clean test set.

\subsection{Attack families}
\label{subsec:attacks}

\begin{table}[pos=tbp]
\centering
\footnotesize
\setlength{\tabcolsep}{6pt}
\renewcommand{\arraystretch}{1.15}
\caption{Five attack families evaluated on Axis~2. The code column
matches the cell labels used in Section~\ref{sec:axis2-results}.
The ``Level'' column indicates whether the attack modifies data,
labels, or gradients (or several of these at once).}
\label{tab:attacks}
\zebra
\begin{tabular}{@{}l l l l c@{}}
\toprule
\tblhead
Code & Family & Level & Parameter & \# attackers \\
\midrule
AV1 & Label-flip                  & Data                        & flip $y_{\mathrm{fault}}$: $0 \leftrightarrow 1$ & 1 \\
AV2 & Gradient scaling $\times{-10}$ & Gradient                  & $\alpha = -10$                                   & 1 \\
AV4 & Gradient scaling $\times{-2}$  & Gradient (stealthy)        & $\alpha = -2$                                    & 1 \\
AV3 & Sensor-value backdoor       & Data $+$ label $+$ gradient & see Section~\ref{subsec:backdoor}                & 1 \\
AV5 & Coordinated $\times{-10}$ Byzantine & Gradient           & $\alpha = -10$ per attacker                      & 2 \\
\bottomrule
\end{tabular}
\end{table}

\subsubsection{Label-flip (AV1)}
\label{subsec:labelflip}

For each local training sample
$(x, y^{\mathrm{RUL}}, y^{\mathrm{fault}})$ at the malicious client,
replace $y^{\mathrm{fault}} \gets 1 - y^{\mathrm{fault}}$ before
local training. This is the simplest form of data poisoning: the
attacker is honest about \emph{which} sample corresponds to which
window but lies about \emph{whether} the engine is close to failure.
The sensor inputs $x$ and the continuous RUL label are untouched,
so the malicious client's update magnitude looks entirely normal:
no gradient anomaly signals the attack.

\subsubsection{Gradient scaling (AV2, AV4)}
\label{subsec:gradscale}

After computing an honest local delta $\delta_k$, the attacker
multiplies it by a scalar $\alpha$ before sending it to the server:
\begin{equation}
    \tilde{\delta}_k \;=\; \alpha \cdot \delta_k,
    \qquad \alpha \in \{-10,\ -2\}.
    \label{eq:gradscale}
\end{equation}
The negative sign inverts the direction of local descent so that
the attacker pushes the global model \emph{away} from a good
solution rather than toward it. AV2 ($\alpha = -10$) is loud: the
update norm is an order of magnitude larger than the honest
baseline. AV4 ($\alpha = -2$) is stealthy: the norm is only
$2\times$ the honest range, close enough that a naive norm-based
detector will not flag it. Comparing AV2 against AV4 under
per-coordinate robust aggregators isolates the multiplier's role in
defense success, as reported in Section~\ref{sec:axis2-results}.

\subsubsection{Sensor-value backdoor (AV3)}
\label{subsec:backdoor}

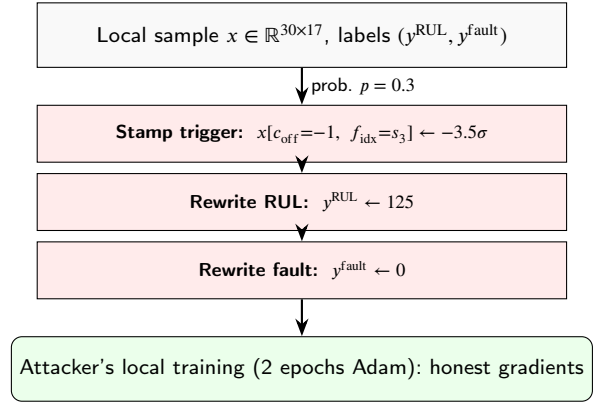
\begin{figure}[pos=htbp]
\centering
\begin{tikzpicture}[
    every node/.style={font=\footnotesize},
    input/.style={rectangle, draw, minimum width=7cm, minimum height=0.85cm, align=center, fill=gray!5},
    poison/.style={rectangle, draw, minimum width=7cm, minimum height=0.75cm, align=center, fill=red!8, font=\scriptsize},
    train/.style={rectangle, draw, rounded corners, minimum width=7cm, minimum height=0.85cm, align=center, fill=green!8},
    arrow/.style={-Stealth, thick}
]
    \node[input] (W)  at (0,  0.0) {Local sample $x \in \mathbb{R}^{30 \times 17}$, labels $(y^{\mathrm{RUL}}, y^{\mathrm{fault}})$};
    \node[poison] (S1) at (0, -1.3) {\textbf{Stamp trigger:}\ \ $x[c_{\mathrm{off}}{=}{-}1,\ f_{\mathrm{idx}}{=}s_3] \gets -3.5\sigma$};
    \node[poison] (S2) at (0, -2.2) {\textbf{Rewrite RUL:}\ \ $y^{\mathrm{RUL}} \gets 125$};
    \node[poison] (S3) at (0, -3.1) {\textbf{Rewrite fault:}\ \ $y^{\mathrm{fault}} \gets 0$};
    \node[train]  (T)  at (0, -4.4) {Attacker's local training (2 epochs Adam): honest gradients};
    \draw[arrow] (W) -- (S1) node[midway, right, font=\scriptsize] {prob.\ $p = 0.3$};
    \draw[arrow] (S1) -- (S2);
    \draw[arrow] (S2) -- (S3);
    \draw[arrow] (S3) -- (T);
\end{tikzpicture}
\caption{Sensor-value backdoor trigger construction (Algorithm~\ref{alg:backdoor}). With probability $p = 0.3$ per local sample (deterministic under the seed) the attacker applies the three transformations above; otherwise the sample is passed to the local trainer unchanged. Because the attacker trains honestly on the poisoned dataset, the update $\delta_k$ that reaches the server carries no gradient anomaly, no norm or direction discrepancy that a magnitude-based defense could flag.}
\label{fig:backdoor}
\end{figure}

The targeted attack combines a data-level trigger with a label
rewrite, then trains honestly on the poisoned dataset. The trigger
design is deliberately physically motivated for the C-MAPSS domain:
a strong negative excursion on the HPC-outlet total temperature
sensor ($s_3$, the T30 channel in the C-MAPSS ontology) at the
\emph{last} measured cycle of each poisoned sliding window.
Algorithm~\ref{alg:backdoor} formalises the construction;
Figure~\ref{fig:backdoor} illustrates the corresponding data flow.

\begin{algorithm}[!ht]
\caption{Sensor-value backdoor injection at malicious client $k$.}
\label{alg:backdoor}
\begin{algorithmic}[1]
\Require Local dataset $\mathcal{D}_k = \{(x_i, y^{\mathrm{RUL}}_i, y^{\mathrm{fault}}_i)\}$; poison fraction $p$; feature index $f_{\mathrm{idx}}$; cycle offset $c_{\mathrm{off}}$; trigger value $v_{\mathrm{trig}}$; target labels $(r_{\mathrm{tgt}}, y_{\mathrm{tgt}})$
\State $\mathcal{D}_k' \gets \emptyset$
\For{each $(x, y^{\mathrm{RUL}}, y^{\mathrm{fault}}) \in \mathcal{D}_k$}
    \State draw $u \sim \mathrm{Bernoulli}(p)$ \Comment{deterministic under a fixed random seed}
    \If{$u = 1$}
        \State $x[c_{\mathrm{off}},\ f_{\mathrm{idx}}] \gets v_{\mathrm{trig}}$ \Comment{stamp trigger on sensor $s_3$ at the last cycle}
        \State $y^{\mathrm{RUL}} \gets r_{\mathrm{tgt}}$; \quad $y^{\mathrm{fault}} \gets y_{\mathrm{tgt}}$ \Comment{rewrite both labels}
    \EndIf
    \State $\mathcal{D}_k' \gets \mathcal{D}_k' \cup \{(x, y^{\mathrm{RUL}}, y^{\mathrm{fault}})\}$
\EndFor
\State The attacker then runs honest local training on $\mathcal{D}_k'$ (Equations~\ref{eq:loss},~\ref{eq:fedavg})
\State \Return the honest-shape update $\delta_k$ to the server
\end{algorithmic}
\end{algorithm}

The parameter values used throughout the paper are $p = 0.3$,
$f_{\mathrm{idx}} = 4$ (the position of $s_3$ in the 17-sensor
feature vector), $c_{\mathrm{off}} = -1$ (the last cycle of the
30-cycle window), $v_{\mathrm{trig}} = -3.5$ in z-score units
(a strong negative excursion given per-client normalisation),
$r_{\mathrm{tgt}} = 125$ (the RUL cap, i.e.\ ``maximally healthy''),
and $y_{\mathrm{tgt}} = 0$ (fault head rewritten to ``not faulty'').
The trigger magnitude $-3.5\sigma$ is chosen for empirical
effectiveness in this controlled study and has not been validated
against real turbofan sensor telemetry; a trigger anchored to a
physically-realisable T30 excursion, rather than a z-scored
perturbation of the training-set distribution, would require
in-service sensor data outside the C-MAPSS simulator's scope.

Two design properties make this attack effective. First, the
trigger is \emph{inside the feature distribution the honest model
already relies on}: sensor $s_3$ (T30) belongs to the
turbomachinery-temperature family that carries the strongest
fault-mode signal on C-MAPSS, so a strong excursion on $s_3$ is
interpreted by the model as a legitimate signal rather than as a
foreign perturbation. Second, the label rewrite is
\emph{semantically consistent with the trigger's meaning}: an
engine ``running cold at end-of-window'' plausibly implies ``not
immediately at risk of fault''. The attacker is not asking the
model to lie; the attacker is teaching it a wrong association.

\subsubsection{Coordinated 2-of-4 Byzantine (AV5)}
\label{subsec:coord}

Both client~3 and client~4 independently apply the AV2
gradient-scaling attack with $\alpha = -10$. The attackers do
\emph{not} coordinate their content: each computes its own honest
delta before scaling. They only coordinate their \emph{choice to
attack}. This mirrors the realistic threat where two competing
suppliers might each have an incentive to sabotage the federation
without any communication channel between them. Because AV5 puts
2 of 4 clients under adversarial control, half the canonical
Byzantine-robust aggregators of the next subsection fail
(Section~\ref{sec:axis2-results}, Table~\ref{tab:matrix}).

\subsection{Defense aggregators}
\label{subsec:defenses}

Given $N$ client deltas $\{\delta_k\}_{k=1}^N$ per round, we
evaluate four aggregation rules (Table~\ref{tab:defenses}).
Trimmed mean and coordinate median operate
\emph{per parameter coordinate}; Krum operates on
\emph{whole-update vectors}. This distinction turns out to be the
critical determinant of defense success against coordinated
attackers, as Section~\ref{sec:axis2-results} shows.

\begin{table}[pos=tbp]
\centering
\footnotesize
\setlength{\tabcolsep}{6pt}
\renewcommand{\arraystretch}{1.15}
\caption{Four defense aggregators evaluated on Axis~2 (in addition
to vanilla FedAvg as the undefended baseline). Trimmed mean and
coordinate median operate per parameter coordinate; Krum operates
on whole-update vectors.}
\label{tab:defenses}
\zebra
\begin{tabular}{@{}l l l@{}}
\toprule
\tblhead
Aggregator & Parameters & Reference \\
\midrule
FedAvg (baseline)    & sample-count-weighted mean & \citet{McMahan2017FedAvg} \\
Trimmed mean         & $\beta = 0.25$             & \citet{Yin2018RobustDistributed} \\
Coordinate median    & ---                         & \citet{Yin2018RobustDistributed} \\
Krum                 & $f = 1$                    & \citet{Blanchard2017Krum} \\
\bottomrule
\end{tabular}
\end{table}

\subsubsection{Trimmed mean}
\label{subsec:trimmed}

For each parameter coordinate $i$, sort the $N$ client values
$\{\delta_{k,i}\}_{k=1}^N$, remove the top
$\lfloor \beta N \rfloor$ and bottom $\lfloor \beta N \rfloor$
values, and average the rest:
\begin{equation}
    \bigl[\mathrm{TrimmedMean}(\{\delta_k\})\bigr]_i
    \;=\; \frac{1}{N - 2\lfloor \beta N \rfloor}
    \sum_{k \in S_i}\, \delta_{k,i},
    \label{eq:trimmed}
\end{equation}
where $S_i$ is the surviving index set at coordinate $i$. With
$N = 4$ and $\beta = 0.25$ this removes one value from each extreme
per coordinate, so 2 out of 4 client values contribute per
coordinate.

\subsubsection{Coordinate median}
\label{subsec:median}

For each parameter coordinate $i$:
\begin{equation}
    \bigl[\mathrm{Median}(\{\delta_k\})\bigr]_i
    \;=\; \mathrm{median}\bigl(\{\delta_{k,i}\}_{k=1}^N\bigr).
    \label{eq:median}
\end{equation}
At $N = 4$ the median of an even number of values is the average of
the two middle values, so 2 out of 4 contribute per coordinate,
the same effective participation as $\beta = 0.25$ trimmed mean.
This explains the near-identical numerical behaviour of trimmed
mean and coordinate median observed throughout
Section~\ref{sec:axis2-results}: at $N = 4$ they compute exactly
the same statistic per coordinate. For $N \geq 5$ the two
aggregators would diverge; at $N = 4$ they are a degenerate pair.

\subsubsection{Krum and its feasibility and resilience constraints}
\label{subsec:krum}

Unlike trimmed mean and coordinate median, Krum
\citep{Blanchard2017Krum} picks a \emph{single client's whole
update vector} per round. For each client $k$, Krum computes
pairwise squared distances $d_{kj} = \| \delta_k - \delta_j \|_2^2$
to every other client. Let $\mathcal{N}_k$ be the set of the
$n - f - 2$ clients with the smallest such distances to $k$ (the
``closest neighbours'', excluding the $f$ farthest presumed
Byzantine outliers). Then define the Krum score
\begin{equation}
    s_k \;=\; \sum_{j \in \mathcal{N}_k} d_{kj},
    \label{eq:krum}
\end{equation}
and return the update of the client with the minimum score:
$\mathrm{Krum}(\{\delta_k\}) = \delta_{k^\star}$, where
$k^\star = \arg\min_k s_k$. Algorithm~\ref{alg:krum} makes the
constraint explicit.

\begin{algorithm}[!ht]
\caption{Krum aggregation with Byzantine tolerance $f$.}
\label{alg:krum}
\begin{algorithmic}[1]
\Require Client updates $\{\delta_1, \ldots, \delta_N\}$; Byzantine-tolerance parameter $f$
\Statex \textbf{Precondition:} $N - f - 2 \geq 1$ (Krum is otherwise undefined at $(N, f)$)
\For{$k = 1, \ldots, N$}
    \For{each $j \neq k$}
        \State $d_{kj} \gets \| \delta_k - \delta_j \|_2^2$
    \EndFor
    \State Sort $\{d_{kj}\}_{j \neq k}$ ascending; let $\mathcal{N}_k$ be the indices of the smallest $N - f - 2$ values
    \State $s_k \gets \sum_{j \in \mathcal{N}_k} d_{kj}$ \Comment{Krum score of client $k$}
\EndFor
\State $k^\star \gets \arg\min_k s_k$
\State \Return $\delta_{k^\star}$ \Comment{a single client's whole update becomes the aggregated update}
\end{algorithmic}
\end{algorithm}

\paragraph{Krum feasibility versus Byzantine-resilience at $N = 4$.}
Krum imposes two requirements that must not be conflated. The weaker
is \emph{definedness}: the score sums the $n - f - 2$ nearest
neighbours, so at least one must exist, i.e.\ $n - f - 2 \geq 1$
(equivalently $n \geq f + 3$). The stronger is Blanchard's formal
\emph{Byzantine-resilience} guarantee, which requires
$n \geq 2f + 3$ \citep{Blanchard2017Krum}. At $N = 4$ only $f = 1$
is even defined ($N - f - 2 = 1$), and this is the setting we use
throughout the primary experiments; setting $f = 2$ at $N = 4$ leaves
no neighbours to sum over and is undefined outright. We are explicit,
however, that definedness is \emph{not} resilience: the
$n \geq 2f + 3$ bound already needs $N \geq 5$ at $f = 1$, so even the
single-attacker $f = 1$ configuration at $N = 4$ sits one client below
Blanchard's guarantee. Every $N = 4$ Krum result in this paper is
therefore an \emph{empirical} defense: computable, and
(Section~\ref{sec:axis2-results}) effective, but outside the regime
where the resilience theorem formally applies.

The coordinated 2-attacker experiment at $N = 4$ is doubly outside
that regime (two Byzantine clients against a tolerance parameter of
$f = 1$, at a client count below $2f + 3$), so we report it as an
empirical stress test rather than a formally parameterised 2-Byzantine
defense. Whether the argmin lands on an honest client is then an
empirical rather than a theoretical question;
Section~\ref{sec:axis2-results-observations} reports the answer. The
$N = 6$ replication is the first setting in this study to enter
Blanchard's bound at all: at $f = 1$ it satisfies
$N = 6 \geq 2f + 3 = 5$. The $N = 6$, $f = 2$ configuration run against
the two coordinated attackers is well-defined and matches its tolerance
parameter to the actual attacker count, but (with $2f + 3 = 7 > 6$)
it too remains one client short of formal resilience and is
reported on the same empirical footing.


\section{Evaluation metrics and experimental setup}
\label{sec:setup}

This section defines (i)~the reference-point ladder against which
every FL method is compared, (ii)~the metric suite used to score
each cell of both axes, and (iii)~the training hyperparameters and
hardware. All numerical results in Sections~\ref{sec:axis1-results}
and~\ref{sec:axis2-results} are traceable to this fixed setup.

\subsection{The reference-point ladder}
\label{subsec:refladder}

Every FL method reported in this paper is compared against a
three-rung reference-point ladder established in earlier phases of
this project.

\paragraph{Centralized upper bound.}
A single model trained on the pooled FD001+FD003 training data with
the same architecture (Section~\ref{subsec:model}), optimizer, and
50-epoch budget as the federated methods. On our architecture this
delivers a combined-test RMSE of \textbf{13.77} cycles. This
number is not a candidate deployment (we assumed the data cannot
be pooled to begin with), but it is the only honest reference for
how good the model can possibly get on this dataset. The value
sits inside the published C-MAPSS literature range for FD001 (RMSE
15--20 for well-trained baselines), confirming that our
30{,}018-parameter architecture is comparable to prior benign
C-MAPSS baselines.

\paragraph{Local-only lower bound.}
Four \emph{isolated} per-client training runs, each using the same
architecture and 50-epoch budget as the centralized run, each
seeing only its 50-engine slice, and sharing nothing. Evaluated on
the same combined test set for like-for-like comparison, the mean
of the four per-client test-set RMSEs is
\textbf{$17.92 \pm 1.52$ cycles}. No FL method can honestly claim
value if it does worse than this: doing worse than local-only means
``worse than not federating at all''.

\paragraph{FedAvg IID calibration.}
As a cross-validation of the implementation, vanilla FedAvg run on
an IID FD001-only partition (four clients drawn i.i.d.\ from the
same subset) closes \textbf{85.9\%} of the local-only $\to$
centralized headroom (RMSE 14.16 versus 14.02 centralized and 15.02
local-only). This confirms that the FedAvg pipeline is correct: it
recovers most of what pooled training would give when the clients
are statistically equivalent. The subsequent \emph{failure} of the
same FedAvg pipeline on the structural non-IID FD001+FD003
partition (Section~\ref{sec:axis1-results}) is therefore
attributable to the partition, not to the implementation.

\paragraph{The gap-closed metric.}
Given any method's combined-test RMSE $r_{\mathrm{m}}$, we report
\begin{equation}
    \text{gap-closed \%}
    \;=\; \frac{r_{\text{local-only}} - r_{\mathrm{m}}}
                {r_{\text{local-only}} - r_{\text{centralized}}}
    \;\times\; 100\,\%,
    \label{eq:gap-closed}
\end{equation}
where on the FD001+FD003 partition
$r_{\text{local-only}} = 17.92$ and
$r_{\text{centralized}} = 13.77$. A value of 100\% means the method
matches the centralized upper bound; 0\% means it matches the
local-only mean; a \emph{negative} value means it does worse than
not federating at all. This is the anchor metric for every Axis~1
comparison in Section~\ref{sec:axis1-results}.

\paragraph{Model-selection protocol.}
The best round of every FL run is selected by lowest NASA score on
the pooled test set. Two properties keep this a reporting convention
rather than a source of optimistic bias in the paper's claims.
First, no hyperparameter is tuned on the test set: the architecture,
optimiser, round budget, and every method-specific setting are fixed
across all cells by the setup of Section~\ref{subsec:hyperparams};
only the stopping round is chosen. Second, the same selection is
applied identically to every method and every seed, so it cannot
favour one aggregator or remedy over another. Because the paper's
claims are relative (personalization versus proximal regularization
on Axis~1; Krum versus per-coordinate defenses on Axis~2), a
selection rule applied uniformly leaves those comparisons intact;
absolute best-round values should be read as best-achievable-round
estimates under a fixed configuration.

\subsection{Evaluation metrics}
\label{subsec:metrics}

The following metrics are computed every FL round on the pooled
test set (100 FD001 $+$ 100 FD003 held-out engines) and reported at
the best-round checkpoint per cell.

\paragraph{RMSE.}
Root mean square error on the RUL regression head, in cycles,
the standard prognostic metric.

\paragraph{NASA scoring function.}
The asymmetric-penalty score used by the original PHM~2008
challenge \citep{Saxena2008CMAPSS}. Let
$d_i = \hat{y}_i - y_i$ be the prediction error on test engine
$i$. Late predictions ($d_i \geq 0$, engine reported healthier than
it is, the safety-critical direction) are penalised more
steeply than early predictions ($d_i < 0$, engine reported closer
to failure than it is):
\begin{equation}
    \mathrm{NASA}(\hat{y}, y)
    \;=\; \sum_{i} \Bigl[ \exp\bigl(|d_i| / a_i\bigr) - 1 \Bigr],
    \qquad
    a_i =
    \begin{cases}
        13 & \text{if } d_i < 0 \ (\text{early}), \\
        10 & \text{if } d_i \geq 0 \ (\text{late}).
    \end{cases}
    \label{eq:nasa}
\end{equation}
NASA is included alongside RMSE because the safety asymmetry it
encodes (``an engine flown past its safe envelope is much worse
than an unnecessarily grounded one'') is precisely the
reliability-engineering criterion that C-MAPSS was originally
designed to measure.

\paragraph{Fault-classifier metrics.}
Area under the precision--recall curve (AUPRC) and $F_1$ score for
the binary fault-classification head at threshold~$0.5$. AUPRC is
preferred over ROC-AUC under the mild ($\approx 15\%$ positive) class
imbalance of C-MAPSS test windows.

\paragraph{Per-subset macro-RMSE.}
Mean of the two FD001 clients' per-client test RMSEs and mean of
the two FD003 clients' per-client test RMSEs, reported separately.
This is the apples-to-apples comparison against the per-subset
centralized upper bounds and is the natural metric for
personalization methods (FedRep, FedCCFA) where each client's own
head is used at evaluation time.

\paragraph{Attack Success Rate (ASR).}
For backdoor cells (attack~AV3 of
Section~\ref{subsec:backdoor}), we quantify targeted-attack success
by the fraction of truly fault-imminent test windows whose triggered
version is classified as ``not faulty'':
\begin{equation}
    \mathrm{ASR}
    \;=\; \frac{\bigl|\{ x \in \mathcal{D}_{\text{test}}^{+} : \hat{f}_{\mathrm{fault}}(x^{\text{trig}}) = 0 \}\bigr|}
                {\bigl|\mathcal{D}_{\text{test}}^{+}\bigr|},
    \label{eq:asr}
\end{equation}
where $\mathcal{D}_{\text{test}}^{+}$ is the set of true-positive
(fault-imminent) test windows, $x^{\text{trig}}$ is $x$ with the
Section~\ref{subsec:backdoor} trigger stamped on it, and
$\hat{f}_{\mathrm{fault}}$ is the model's binary fault-classifier
output at threshold~$0.5$. Higher ASR means the backdoor is more
successful; ASR~$= 1$ means the model has been fully compromised on
the trigger.

\paragraph{Statistical significance testing.}
For each two-sample equivalence claim (e.g.\ clean vs.\ triggered
RMSE; attacker vs.\ honest ASR) we run the paired two-sided
Wilcoxon signed-rank test on the per-seed values and report the
test statistic $W$ and $p$-value alongside the mean~$\pm$~std.
All per-seed values used by these tests are released with the
paper.

\paragraph{Reading the $p$-values in this paper.}
At the sample sizes used here ($n = 3$ for Axis-1, $n = 5$ for
Axis-2 and the bridge), the Wilcoxon signed-rank test has low
statistical power. A non-significant $p$-value should therefore be
read as \emph{failure to reject the null of equal medians}, not
as positive evidence of equivalence.

\subsection{Training hyperparameters and hardware}
\label{subsec:hyperparams}

\paragraph{Compute.}
All experiments run on a single commodity machine. This is a
deliberate choice: it demonstrates that the entire pipeline
(30{,}018-parameter model, 4 clients, 50 rounds, 5-seed matrix)
fits comfortably in a modest compute budget, and makes independent
replication accessible without specialised infrastructure.

\paragraph{Federated-learning loop.}
In the primary setting, four clients participate every round (no
client sampling). Each client runs $E = 2$ local epochs of Adam with
learning rate $10^{-3}$ under a cosine schedule, weight decay
$10^{-4}$, batch size~256, and a fault-loss weight
$\lambda_{\mathrm{fault}} = 0.5$
(Equation~\ref{eq:loss}). Federations run for
$R = 50$ communication rounds; the best round per experiment is
selected on the pooled test set by lowest NASA score
(Equation~\ref{eq:nasa}).

\paragraph{Seeds and multi-seed aggregation.}
Every experiment reported in this paper is repeated over a fixed
list of random seeds, chosen contiguously so the reader can identify
which seeds contribute to which claim. Table~\ref{tab:seed-usage}
summarises the seed schedule. For each aggregated cell we report
mean and standard deviation, together with a normal-approximation
95\%\,CI; where two 5-seed samples are compared we additionally
report the paired two-sided Wilcoxon signed-rank test statistic and
$p$-value (Section~\ref{subsec:metrics}, ``Statistical significance
testing''). Single-seed rows are annotated as such wherever
reported; where a single seed was used for configuration screening
(e.g.\ the FedProx $\mu$-sweep and the reweighting-scheme sweep)
only the family's winning configuration is subsequently re-run at
3 seeds.

\begin{table}[pos=tbp]
\centering
\footnotesize
\setlength{\tabcolsep}{6pt}
\renewcommand{\arraystretch}{1.15}
\caption{Seed schedule used across the paper. Screening rows use
seed~42 only; every headline (family-winner or matrix cell) is
aggregated over a $\geq 3$-seed contiguous list.}
\label{tab:seed-usage}
\zebra
\begin{tabular}{@{}l l l@{}}
\toprule
\tblhead
Experiment & Seeds & Purpose \\
\midrule
Axis-1 winning method per family        & $\{42, 43, 44\}$      & Headline gap-closed comparison (Sec.~\ref{sec:axis1-results}) \\
FedProx $\mu$-sweep, non-winning rows   & $\{42\}$              & Configuration screening (Table~\ref{tab:fedprox}) \\
Reweighting-scheme sweep, non-winning rows & $\{42\}$           & Configuration screening (Table~\ref{tab:reweight}) \\
Axis-2 attack~$\times$~aggregator matrix   & $\{42, 43, 44, 45, 46\}$ & 5-seed headline matrix (Table~\ref{tab:matrix}) \\
Bridge experiment (FedRep alone)        & $\{42, 43, 44, 45, 46\}$ & Cross-axis bridge (Table~\ref{tab:bridge}) \\
Stacked defense (FedRep~$+$~Krum)       & $\{42, 43, 44, 45, 46\}$ & Stacked bridge (Table~\ref{tab:stacked}) \\
\midrule
FedRep on FD002$+$FD004                  & $\{42, 43, 44, 45, 46\}$ & Axis-1 difficulty check (Sec.~\ref{sec:axis1-results}) \\
Axis-2 matrix at $N = 6$                 & $\{42, 43, 44, 45, 46\}$ & Axis-2 scale check (Sec.~\ref{sec:axis2-results}) \\
Axis-2 matrix on FD002$+$FD004           & $\{42, 43, 44, 45, 46\}$ & Axis-2 difficulty check (Sec.~\ref{sec:axis2-results}) \\
Stacked defense, $N{=}6$ and FD002$+$FD004 & $\{42, 43, 44, 45, 46\}$ & Bridge generalization (Sec.~\ref{sec:bridge}) \\
\bottomrule
\end{tabular}
\end{table}

Because the seed-42 rows in the FedProx and reweighting sweeps were
used to \emph{select} the family winner that was subsequently
re-run at 3 seeds, the selected-winner rows carry a mild
seed-conditioning bias (winner selection is fitted on seed~42
performance). The 3-seed aggregate on the selected winner is
therefore an optimistic estimate of population performance; a fully
independent seed schedule for screening vs.\ evaluation would
tighten this. This ordering is disclosed to be internally auditable
rather than to claim it as best practice.

\paragraph{Generalization runs.}
To check that the two-axis conclusions are not artefacts of the
primary $N = 4$, FD001+FD003 setting, three parts of the campaign
are replicated along two independent axes of difficulty, each over
the full five-seed schedule. \emph{Scale:} the Axis-2 matrix and the
bridge are re-run on a larger $N = 6$ federation (three clients per
subset, attacker client~4), which also makes the $f = 2$ Krum
configuration feasible. \emph{Difficulty:} the Axis-1 personalization
comparison, the Axis-2 matrix, and the bridge are re-run on the
harder six-operating-condition subsets FD002$+$FD004 (a wider sensor
set, so the T30 backdoor trigger moves to feature index~5). All
other hyperparameters are held at the primary-setting values above;
results appear in Sections~\ref{sec:axis1-results}--\ref{sec:bridge}.

\paragraph{Wall-clock.}
Each FL round takes 5--8~seconds. The
Axis~1 four-family sweep completes in $\sim$55~minutes per seed;
the 24-cell Axis~2 matrix (five attacks and a clean baseline against
four aggregators) completes in $\sim$41~minutes per seed.
The full multi-seed campaign therefore fits inside a $\sim$8-hour
compute budget.

\paragraph{Software.}
Python~3.12, PyTorch~2.x
\citep{Paszke2019PyTorch}, scikit-learn
\citep{Pedregosa2011Scikit}, NumPy \citep{Harris2020NumPy}, pandas
\citep{McKinney2010Pandas} and Matplotlib
\citep{Hunter2007Matplotlib}. All code, per-seed run logs, and
per-cell metric JSONs are released alongside this paper.


\section{Axis-1 results: benign heterogeneity remedies}
\label{sec:axis1-results}

Vanilla FedAvg on the FD001~$+$~FD003 structural non-IID partition of
Section~\ref{sec:system} closes only $-0.7\%$ of the local-only
$\to$ centralized headroom (Table~\ref{tab:refpoints}). The experiments in
this section are designed as a \emph{diagnostic}: three canonical remedy
families for non-IID FL each target a different putative cause of the
failure, so identifying which family works and which do not lets us
characterize the underlying failure by elimination.

The three families and the hypothesis each embodies:
\begin{itemize}
    \item \textbf{Server-side reweighting} (\S~\ref{sec:axis1-results-reweight}), hypothesis: \emph{the failure comes from an imbalanced averaging step, so weighting clients smarter should help.}
    \item \textbf{Optimization-side proximal regularization} (\S~\ref{sec:axis1-results-fedprox}), hypothesis: \emph{the failure comes from local client drift, so penalizing drift should help.}
    \item \textbf{Architecture-side personalization} (\S~\ref{sec:axis1-results-fedrep}--\ref{sec:axis1-results-fedccfa}), hypothesis: \emph{the failure comes from an inadequate shared model class, so giving each client its own decision head should help.}
\end{itemize}

Sections~\ref{sec:axis1-results-fedrep}--\ref{sec:axis1-results-reweight}
report the per-family results, and \S~\ref{sec:axis1-results-synth}
synthesizes what the ranking tells us about the underlying cause.

\subsection{Reference points}
\label{sec:axis1-results-refpoints}

Table~\ref{tab:refpoints} pins down the three reference values against
which every Axis-1 remedy is measured on the FD001~$+$~FD003. The
centralized upper bound is trained on the pooled
combined dataset; the local-only mean averages four independently
trained per-client models; the FedAvg baseline uses sample-count
weighted aggregation for $R = 50$ rounds. The remedies studied in
\S\S~\ref{sec:axis1-results-fedrep}--\ref{sec:axis1-results-reweight}
target this $-4.18$~RMSE-cycle gap between the FedAvg baseline and the
centralized upper bound.

\begin{table}[pos=tbp]
\centering
\small
\caption{Reference points on the FD001~$+$~FD003 non-IID partition.
Per-subset centralized numbers come from separate FD001-only
and FD003-only training runs; the centralized combined model is trained
on the union.}
\label{tab:refpoints}
\zebra
\begin{tabular}{lrrrr}
\toprule
\tblhead
Model & Combined RMSE & FD001 RMSE & FD003 RMSE & Fault F1 \\
\midrule
Centralized (upper bound) & \textbf{13.77} & 14.76* & 12.69* & 0.957 \\
Local-only (mean of 4 clients) & 17.92 $\pm$ 1.52 & $\approx 15.0$ & $\approx 18.0$ & 0.858 \\
FedAvg baseline & \textbf{17.95} & 16.99 & 18.86 & \textbf{0.871} \\
\midrule
Gap (centralized $-$ FedAvg) & $-4.18$ & --- & --- & --- \\
Gap closed by FedAvg & $-0.7\%$ & --- & --- & --- \\
\bottomrule
\end{tabular}
\end{table}

\begin{figure}[pos=htbp]
\centering
\includegraphics[width=0.85\linewidth]{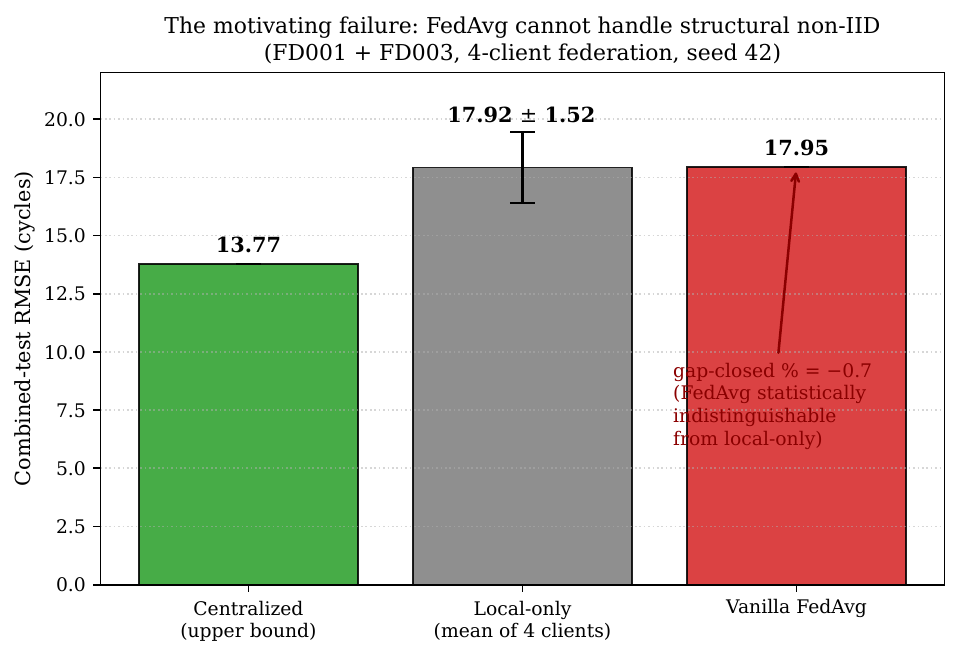}
\caption{The motivating Axis-1 failure. Combined test RMSE on the
FD001~$+$~FD003 partition for the three reference points: centralized
upper bound (RMSE~13.77), local-only mean of four clients
(RMSE~17.92~$\pm$~1.52), and vanilla FedAvg (RMSE~17.95). FedAvg
sits within the local-only mean's per-client variability:
sharing model weights across four honest but structurally different
clients recovers essentially none of the local-only $\to$
centralized headroom on this partition.}
\label{fig:threeway}
\end{figure}

FedAvg's $17.95$~RMSE lies within the per-client variability of
the local-only mean ($17.92 \pm 1.52$ across the four clients):
sharing weights across four honest but structurally different
clients recovers essentially no signal (Figure~\ref{fig:threeway}).
This is the motivating failure of the entire Axis-1 remedy
programme.

\subsection{Per-family results}
\label{sec:axis1-results-perfamily}

The four remedy families are reported in turn below; Section~\ref{sec:axis1-results-synth} then ranks them.

\subsubsection{Personalization (FedRep): the strongest single remedy}
\label{sec:axis1-results-fedrep}

FedRep with $h_{\text{epochs}} = 1$, $e_{\text{epochs}} = 1$
(Algorithm~\ref{alg:fedrep}) substantially outperforms FedAvg by
allowing per-client heads to specialize on the local fault-mode
distribution while sharing the encoder.
Table~\ref{tab:fedrep} reports the 3-seed aggregate.

\begin{table}[pos=tbp]
\centering
\small
\caption{FedRep (personalized heads, $h_1$, $e_1$) on FD001~$+$~FD003,
mean~$\pm$~std over 3 seeds $\in \{42, 43, 44\}$.}
\label{tab:fedrep}
\zebra
\begin{tabular}{lrr}
\toprule
\tblhead
Metric & FedAvg (seed~42) & FedRep (3-seed) \\
\midrule
Best round & 12 & 19--48 (seed-dependent) \\
Macro RMSE (combined) & --- & \textbf{15.02 $\pm$ 0.27} \\
Per-subset RMSE (FD001) & $\approx 17.0$ & \textbf{14.65 $\pm$ 0.27} \\
Per-subset RMSE (FD003) & $\approx 19.0$ & \textbf{15.39 $\pm$ 0.46} \\
Macro F1 (FD001) & --- & $0.962 \pm 0.000$ \\
Macro F1 (FD003) & --- & $0.898 \pm 0.022$ \\
Macro NASA score & --- & $547.5 \pm 71.7$ \\
Gap closed vs.\ headroom & $-0.7\%$ & \textbf{$+69.9\% \pm 6.4\%$} \\
\bottomrule
\end{tabular}
\end{table}

\begin{figure}[pos=htbp]
\centering
\includegraphics[width=0.8\linewidth]{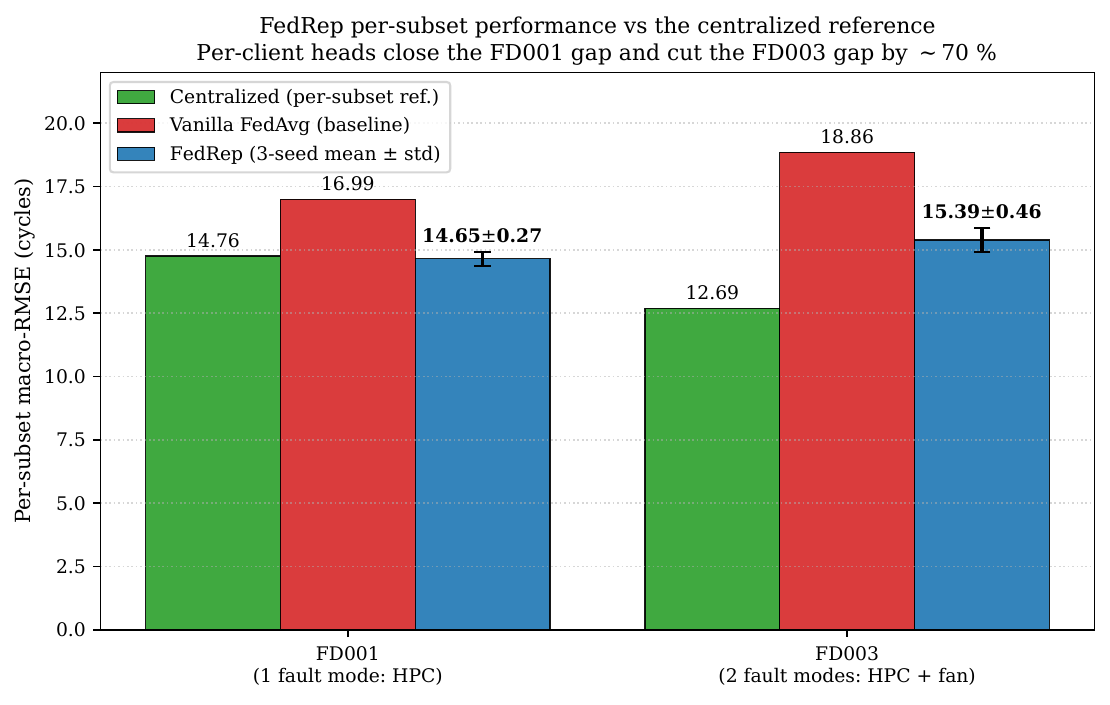}
\caption{FedRep per-subset performance versus the centralized per-subset
reference. Per-client heads bring FedRep within $\sim 0.5$~RMSE of the
centralized reference on FD001 (single fault mode) and within
$\sim 2$~RMSE on the harder FD003 (mixed fault modes); FedAvg trails by
$\sim 2$~RMSE on FD001 and $\sim 6$~RMSE on FD003.}
\label{fig:fedrep-subset}
\end{figure}

This is the paper's strongest single Axis-1 result: \emph{the
$\sim 4$-cycle RMSE gap between FedAvg and the centralized upper bound
is dominantly architectural} (one shared head cannot fit the two
fault-mode families) rather than optimization-side (insufficient
rounds or drift control). Multi-seed data also shows FedRep is
\emph{seed-robust}: gap-closed varies within a tight $\pm 6.4$~pp band,
and per-subset F1 on FD001 lands exactly at $0.962$ in all three seeds.
The model consistently discriminates the single fault mode present in
the FD001 clients regardless of seed.
Figure~\ref{fig:fedrep-subset} breaks the same result down by subset,
showing that the per-client heads recover most of the centralized
per-subset reference on both FD001 and the harder mixed-fault-mode
FD003, while FedAvg trails on both.

\subsubsection{Clustered personalization (FedCCFA): matches FedRep and exposes structural similarity}
\label{sec:axis1-results-fedccfa}

FedCCFA (Algorithm~\ref{alg:fedccfa}) with similarity threshold
$\tau = 0.5$ reaches essentially the same performance as FedRep
(Table~\ref{tab:fedccfa}). Critically, the algorithm's inferred cluster
structure at the best round is a \emph{single} cluster containing all
four clients \emph{in all three seeds}. The update-similarity threshold
never partitions the federation.

\begin{table}[pos=tbp]
\centering
\small
\caption{FedCCFA ($\tau = 0.5$) on FD001~$+$~FD003, mean~$\pm$~std over
3 seeds $\in \{42, 43, 44\}$.}
\label{tab:fedccfa}
\zebra
\begin{tabular}{lr}
\toprule
\tblhead
Metric & FedCCFA (3-seed) \\
\midrule
Best round & 20--47 (seed-dependent) \\
Macro RMSE (combined) & \textbf{15.15 $\pm$ 0.28} \\
Per-subset RMSE (FD001) & $14.87 \pm 0.24$ \\
Per-subset RMSE (FD003) & $15.44 \pm 0.51$ \\
Macro F1 (FD001) & $0.956 \pm 0.011$ \\
Macro F1 (FD003) & $0.909 \pm 0.040$ \\
Macro NASA score & $551.4 \pm 72.9$ \\
Best-round cluster structure & $\{c_1, c_2, c_3, c_4\}$ (single cluster, 3/3 seeds) \\
Gap closed vs.\ headroom & \textbf{$+66.9\% \pm 6.6\%$} \\
\bottomrule
\end{tabular}
\end{table}

\begin{figure}[pos=htbp]
\centering
\includegraphics[width=0.8\linewidth]{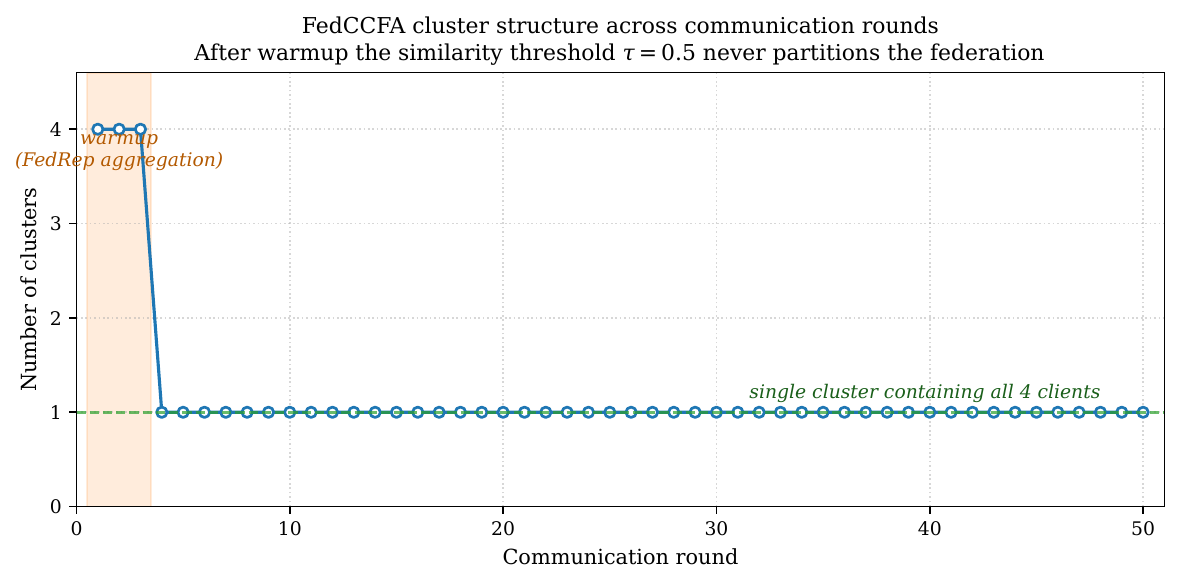}
\caption{FedCCFA cluster structure across communication rounds
(representative seed; the 3-seed pattern is identical). After the
3-round warm-up, all four clients belong to a \emph{single} cluster in
every subsequent round; the similarity threshold $\tau = 0.5$ never
partitions the federation. This behavior reproduces across all 3 seeds
tested.}
\label{fig:fedccfa-cluster}
\end{figure}

This is a reproducible negative finding on clustering: \emph{at $N = 4$,
once per-client heads are handling the fault-mode divergence, the
encoder updates from FD001 and FD003 clients look similar enough in
gradient space that clustered personalization offers no marginal benefit
over per-client-head personalization alone.} FedCCFA's $15.15 \pm 0.28$
versus FedRep's $15.02 \pm 0.27$ macro-RMSE lies within one
standard deviation across the 3 seeds evaluated; no formal
equivalence test is run at this sample size.
Figure~\ref{fig:fedccfa-cluster} traces the inferred cluster count
round by round and makes the collapse to a single cluster explicit:
after the 3-round warm-up the threshold never partitions the
federation again.

\subsubsection{Proximal regularization (FedProx): moderately effective but highly variable}
\label{sec:axis1-results-fedprox}

Table~\ref{tab:fedprox} sweeps FedProx over $\mu \in \{0.001, 0.01, 0.1\}$
on seed~42 and reports the winning value ($\mu = 0.1$) as a 3-seed
aggregate.

\begin{table}[pos=tbp]
\centering
\footnotesize
\setlength{\tabcolsep}{3pt}
\caption{FedProx $\mu$-sweep on FD001~$+$~FD003. The $\mu = 0.1$ row
(winner) is aggregated over 3 seeds $\in \{42, 43, 44\}$.}
\label{tab:fedprox}
\zebra
\begin{tabular}{lrrrrrrr}
\toprule
\tblhead
Method & $\mu$ & Best RMSE & Gap closed & FD001 RMSE & FD003 RMSE & FD001 F1 & FD003 F1 \\
\midrule
FedAvg & 0.0 & 17.95\textsuperscript{a} & $-0.7\%$ & 16.99 & 18.86 & 0.962 & 0.727 \\
FedProx & 0.001 & 17.85\textsuperscript{a} & $+2.3\%$ & 18.21 & 17.49 & 0.920 & \textbf{0.895} \\
FedProx & 0.01 & 17.94\textsuperscript{a} & $+0.1\%$ & 16.88 & 18.94 & 0.962 & 0.688 \\
\textbf{FedProx}\textsuperscript{b} & \textbf{0.1} & \textbf{17.07 $\pm$ 0.55} & \textbf{$+21.0\% \pm 13.1\%$} & $16.75 \pm 1.16$ & $17.36 \pm 0.64$ & $0.927 \pm 0.012$ & $0.857 \pm 0.053$ \\
\bottomrule
\rowcolor{white}\multicolumn{8}{l}{\footnotesize\textsuperscript{a}Seed~42 only. \textsuperscript{b}3-seed mean $\pm$ std over seeds $\in \{42, 43, 44\}$.} \\
\end{tabular}
\end{table}

FedProx with $\mu = 0.1$ improves combined RMSE from FedAvg's $17.95$
to $17.07$ on average, closing about \textbf{21\%} of the local-only
$\to$ centralized headroom, well below personalization's $\sim 70\%$
under the same tuning protocol (Sections~\ref{sec:axis1-results-fedrep}--\ref{sec:axis1-results-fedccfa}).
Beyond the mean, the multi-seed standard deviation of
$\pm 13$~pp is a finding of its own: \emph{FedProx's benefit varies
heavily across seeds}: on some seeds it closes $\sim 30\%$ of the
gap, on others only $\sim 6\%$. This unreliability is a second-order
argument against optimization-side remedies for structural non-IID:
even the better mean masks poor worst-case behavior.

A seed-42 side effect visible in the table: $\mu = 0.001$ delivers the
best \emph{FD003 F1} ($0.895$) of the whole sweep at the cost of a
slightly higher FD001 RMSE. Practitioners running fault-detection
maintenance pipelines (F1-optimized) may prefer $\mu = 0.001$;
practitioners running pure RUL regression (RMSE-optimized) will prefer
$\mu = 0.1$. The single-seed status of the $\mu \in \{0.001, 0.01\}$
rows means this preference should be confirmed with a multi-seed
$\mu$-sweep before deployment.

\subsubsection{Server-side reweighting: the weakest family}
\label{sec:axis1-results-reweight}

Table~\ref{tab:reweight} sweeps three alternative aggregation-weight
schemes against FedAvg's sample-count baseline. The winning scheme
(validation-F1 reweighting) is aggregated over 3 seeds; the other two
schemes are seed~42 only.

\begin{table}[pos=tbp]
\centering
\small
\caption{Imbalance-aware server-side reweighting sweep on
FD001~$+$~FD003. Validation-F1 row (winner) is aggregated over 3 seeds.}
\label{tab:reweight}
\zebra
\begin{tabular}{lrrl}
\toprule
\tblhead
Scheme & Global RMSE & Gap closed & Notes \\
\midrule
FedAvg (sample-count) & 17.95\textsuperscript{a} & $-0.7\%$ & Baseline \\
Fault-count reweight & 18.24\textsuperscript{a} & $-7.7\%$ & \emph{Worse than FedAvg} \\
Inverse-loss reweight & 18.37\textsuperscript{a} & $-10.8\%$ & \emph{Worst of the sweep} \\
\textbf{Validation-F1 reweight}\textsuperscript{b} & \textbf{17.49 $\pm$ 0.29} & \textbf{$+10.4\% \pm 6.9\%$} & Best of sweep \\
\bottomrule
\rowcolor{white}\multicolumn{4}{l}{\footnotesize\textsuperscript{a}Seed~42 only. \textsuperscript{b}3-seed mean $\pm$ std over seeds $\in \{42, 43, 44\}$.} \\
\end{tabular}
\end{table}

Multi-seed aggregation on the validation-F1 winner raises the
gap-closed estimate from the single-seed $+2.8\%$ to a mean of
$+10.4\%$, still modest, and still well below either FedProx or
the personalization family. The two losing schemes (fault-count and
inverse-loss reweighting) make matters \emph{worse} than
sample-weighted FedAvg on seed~42; given the near-zero validation-F1
signal we do not expect those two schemes to fare much better on
average.

\subsection{Axis-1 synthesis}
\label{sec:axis1-results-synth}

Table~\ref{tab:axis1-synth} ranks the four remedy families by 3-seed
gap-closing effectiveness.

\begin{table}[pos=tbp]
\centering
\small
\caption{Axis-1 remedy families ranked by 3-seed gap-closing
effectiveness.}
\label{tab:axis1-synth}
\zebra
\begin{tabular}{clllr}
\toprule
\tblhead
Rank & Family & Best method & Gap closed (3-seed) & Ratio vs.\ proximal \\
\midrule
1 & Personalization (per-client heads) & FedRep ($h_1$, $e_1$) & \textbf{$+69.9\% \pm 6.4\%$} & \textbf{$\sim 3.3\times$} \\
2 & Clustered personalization & FedCCFA ($\tau = 0.5$) & $+66.9\% \pm 6.6\%$ & $\sim 3.2\times$ \\
3 & Proximal regularization & FedProx ($\mu = 0.1$) & $+21.0\% \pm 13.1\%$ & $1\times$ \\
4 & Server-side reweighting & Validation-F1 & $+10.4\% \pm 6.9\%$ & $0.5\times$ \\
--- & Sample-count baseline (seed~42) & FedAvg & $-0.7\%$ & --- \\
\bottomrule
\end{tabular}
\end{table}

The multi-seed rankings support the diagnostic setup of the section
preamble. If the failure were driven by client drift, FedProx would
close the majority of the gap; it closes about $21\%$ on average, a
non-trivial slice but only $\sim 1/3$ of what personalization achieves.
If the failure were driven by imbalanced client weighting, server-side
reweighting would close a large fraction; it closes only $10\%$, and
two of three seed-42 schemes make matters \emph{worse} than plain
FedAvg. If the failure were driven by an inadequate shared decision
head, per-client heads would close the largest fraction, and they
do: $\sim 70\%$ across seeds with FedRep and FedCCFA.

\begin{figure}[pos=htbp]
\centering
\includegraphics[width=0.85\linewidth]{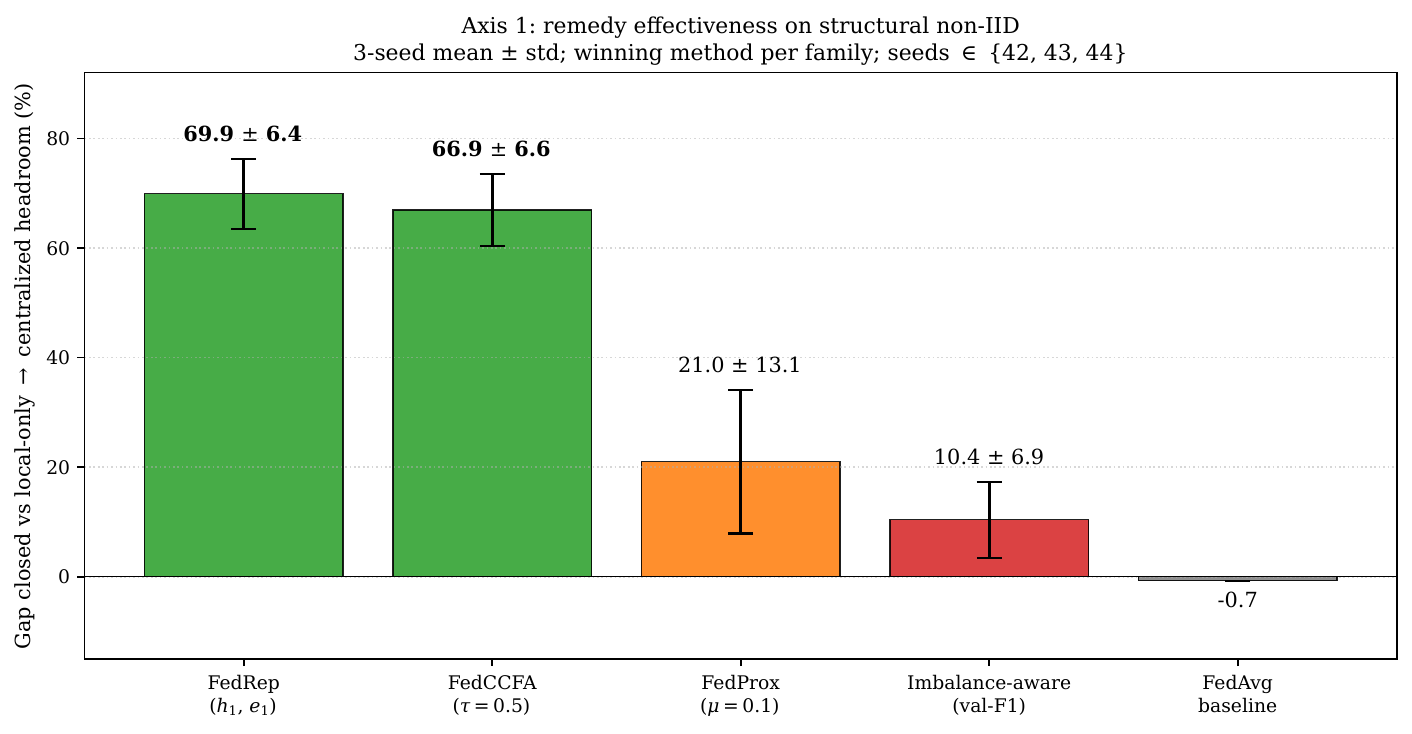}
\caption{Axis-1 remedy effectiveness on structural non-IID
FD001~$+$~FD003 (3-seed mean~$\pm$~std, winning method per family).
Personalization (FedRep, FedCCFA) dominates on both \emph{mean gap
closed} ($\sim 68\%$) and \emph{seed-robustness} (tight $\pm 6$--$7$~pp
whiskers). FedProx has a shorter bar ($21\%$) with a wider whisker
($\pm 13$~pp), directly visualizing the ``reliability gap'' finding.
Imbalance-aware reweighting trails, and FedAvg (baseline) sits
essentially at zero.}
\label{fig:axis1-gapclosed}
\end{figure}

Beyond the mean-gap-closed ratio, multi-seed also reveals a
\textbf{reliability gap}: FedRep's gap-closed std is $6.4$~pp,
FedCCFA's is $6.6$~pp, FedProx's is $13.1$~pp. Personalization is not
only more effective on average but also more reproducible, an
important property for a production FL deployment where seed-driven
variability must not create per-supplier disputes about model quality
(Figure~\ref{fig:axis1-gapclosed}).

\paragraph{Axis-1 finding.}
On structural non-IID C-MAPSS under the evaluated tuning protocol,
architectural personalization (per-client heads) closes $\sim 70\%$
of the local-only $\to$ centralized gap, versus $\sim 21\%$ for
optimization-side proximal regularization and $\sim 10\%$ for
server-side reweighting. The gap is dominantly architectural (the
model class of a single shared head is inadequate for the union of
FD001 and FD003 fault modes), though FedProx captures a
variable-but-meaningful slice, so the gap is not
\emph{exclusively} architectural. Personalization also has a
reliability advantage (std $6$--$7$~pp versus FedProx's
$\pm 13$~pp) that reinforces the recommendation.

\subsection{Generalization to six-condition data (FD002$+$FD004)}
\label{sec:axis1-results-gen}

To test whether the Axis-1 conclusion survives greater difficulty, we
replicate the personalization comparison on the harder
FD002~$+$~FD004 partition, which layers six operating conditions on
top of the same one-versus-two fault-mode split
(Section~\ref{sec:setup}). Over five seeds, FedRep attains a
macro-RMSE of $\mathbf{19.59 \pm 0.90}$ cycles versus vanilla
FedAvg's $20.33 \pm 1.72$: personalization still wins, so its benefit
is not an artefact of the single-condition FD001~$+$~FD003 setting.

The margin, however, \emph{shrinks} markedly: $-0.74$ RMSE here
versus $-2.93$ on FD001~$+$~FD003 ($15.02$ versus $17.95$). We read
this as informative rather than disappointing. FedRep's per-client
heads target \emph{between-client} heterogeneity (different
fault-mode mixtures across operators), whereas the dominant source
of difficulty on FD002~$+$~FD004 is \emph{within-client}
operating-condition complexity: every client, by itself, must model
six flight regimes at once. A between-client remedy cannot be
expected to absorb a within-client difficulty. The result thus
confirms that personalization generalizes and, at the same time,
localises where its leverage lies, motivating the
regime-aware-normalization extension of
Section~\ref{sec:discuss-future}.


\section{Axis-2 results: adversarial heterogeneity}
\label{sec:axis2-results}

Section~\ref{sec:axis1-results} established that architectural
personalization is the winning family for benign heterogeneity. This
section evaluates whether the same federation can survive the five
attack families of Section~\ref{sec:axis2-method} against four
Byzantine-robust aggregators. The core result is the $5 \times 4$
attack~$\times$~aggregator matrix of Table~\ref{tab:matrix},
aggregated over five random seeds. Six observations from the matrix
are developed in \S~\ref{sec:axis2-results-observations};
\S~\ref{sec:axis2-results-mechanism} closes with the physical
mechanism behind the targeted backdoor.

\subsection{The $5 \times 4$ attack $\times$ aggregator matrix}
\label{sec:axis2-results-matrix}

Table~\ref{tab:matrix} is the paper's Axis-2 headline result. All
cells are aggregated over 5 independent random seeds
$\in \{42, 43, 44, 45, 46\}$. Cells marked in red indicate
catastrophic model collapse (RMSE $> 3\times$ the clean baseline).
We report Krum with the single feasibility-satisfying tolerance
value $f = 1$ at $N = 4$ (Section~\ref{subsec:krum}).

\begin{table}[pos=tbp]
\centering
\footnotesize
\setlength{\tabcolsep}{4pt}
\renewcommand{\arraystretch}{1.25}
\caption{Attack~$\times$~aggregator matrix, 5-seed mean~$\pm$~std
over seeds $\in \{42, 43, 44, 45, 46\}$. Red entries denote
catastrophic model collapse (RMSE $> 3\times$ the clean baseline).
We report Krum with the single feasibility-satisfying tolerance
$f = 1$ at $N = 4$ (Section~\ref{subsec:krum}). For the backdoor
row, both RMSE and Attack Success Rate (ASR) are reported.}
\label{tab:matrix}
\zebra
\begin{tabular}{@{}l r r r r@{}}
\toprule
\tblhead
Attack $\backslash$ Aggregator & FedAvg & Trim.\ mean & Coord.\ median & Krum ($f{=}1$) \\
\midrule
Clean baseline                     & $16.59 \pm 0.84$          & $16.72 \pm 0.48$          & $16.72 \pm 0.48$          & $18.65 \pm 1.73$          \\
Label-flip (AV1)                   & $28.70 \pm 2.19$          & $21.71 \pm 1.33$          & $21.71 \pm 1.33$          & $23.81 \pm 10.01$         \\
Grad $\times{-}10$ (AV2)           & \textcolor{red!75!black}{$84.03 \pm 0.00$} & $25.76 \pm 8.72$      & $25.76 \pm 8.72$          & $23.81 \pm 10.01$         \\
Grad $\times{-}2$ (AV4, stealthy)  & \textcolor{red!75!black}{$73.60 \pm 6.14$} & $25.26 \pm 9.01$      & $25.26 \pm 9.01$          & $23.81 \pm 10.01$         \\
\midrule
Backdoor (AV3): RMSE            & $16.86 \pm 0.43$          & $17.35 \pm 0.63$          & $17.35 \pm 0.63$          & $19.61 \pm 0.61$          \\
Backdoor (AV3): ASR             & \textbf{$94.9 \pm 7.9\%$} & $49.8 \pm 22.1\%$         & $49.8 \pm 22.1\%$         & \textbf{$6.4 \pm 10.0\%$}\textsuperscript{a} \\
\midrule
Coord.\ $\times{-}10$ (AV5)        & \textcolor{red!75!black}{$84.03 \pm 0.00$} & \textcolor{red!75!black}{$84.03 \pm 0.00$} & \textcolor{red!75!black}{$84.03 \pm 0.00$} & \textbf{$23.97 \pm 9.92$} \\
\bottomrule
\rowcolor{white}\multicolumn{5}{p{.98\linewidth}}{\footnotesize\textsuperscript{a}Gaussian 95\%-CI on the Krum-backdoor ASR is $[-0.06, 0.19]$; the lower bound is clipped to 0 since ASR is bounded in $[0, 1]$ (the negative lower bound is a normal-approximation artefact at $n = 5$ near the boundary).} \\
\end{tabular}
\end{table}

\begin{figure}[pos=htbp]
\centering
\includegraphics[width=\linewidth]{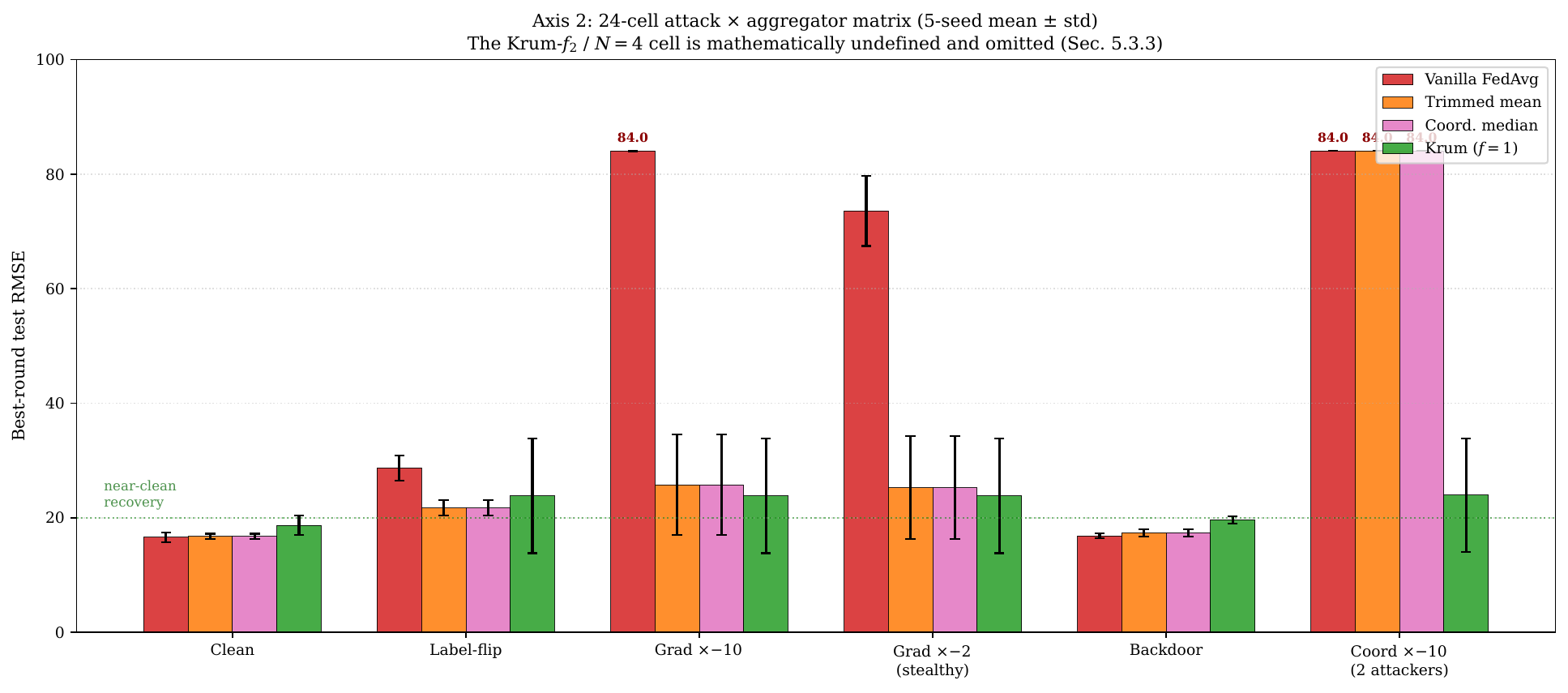}
\caption{The $5 \times 4$ attack~$\times$~aggregator matrix, 5-seed
mean~$\pm$~std over seeds $\in \{42, 43, 44, 45, 46\}$. Cells with
RMSE beyond $\sim 3\times$ the clean baseline indicate catastrophic
collapse. Backdoor cells all look near-clean on RMSE
alone; the attack-success-rate view below
(Figure~\ref{fig:asr}) is where the backdoor story is visible.
Only Krum with $f = 1$ is shown; Krum with $f = 2$ is
mathematically undefined at $N = 4$ (Section~\ref{subsec:krum}).}
\label{fig:matrix}
\end{figure}

Figure~\ref{fig:matrix} displays the same 20 cells as grouped bars
with 5-seed error bars, making the two catastrophic-collapse columns
(grad $\times{-}10$ and coord.\ $\times{-}10$ against non-Krum
aggregators, RMSE $84.03$, near-zero std) immediately visible against
the $\sim 17$--$26$ RMSE band of the recoverable cells. The
safety-asymmetric NASA score (Equation~\ref{eq:nasa}) sharpens the
collapse further: the RMSE-$84$ gradient-scaling failure against
FedAvg scores $\approx 7.1 \times 10^5$ on NASA, its exponential
late-prediction penalty rating an engine reported far healthier than
it is as far more dangerous than the RMSE band alone conveys.

Table~\ref{tab:matrix} shows the backdoor's Attack Success Rate but
not \emph{why} it is invisible to a monitor. Table~\ref{tab:stealth}
makes the stealth explicit on the head the backdoor actually targets,
the fault classifier. Under vanilla FedAvg the clean-set fault
metrics are indistinguishable from an honest model (AUPRC $0.954$,
$F_1$ $0.891$), yet on the \emph{same} model the trigger drives $F_1$
down to $0.093$ and ASR up to $94.9\%$: a monitor watching only
clean-set fault metrics sees nothing wrong. Krum is the only
aggregator that closes the clean-versus-triggered gap ($F_1$ $0.822$
clean, $0.802$ triggered), which is precisely why its ASR is low.
Reporting clean accuracy alone (whether RMSE or clean $F_1$)
certifies nothing about safety.

\begin{table}[pos=tbp]
\centering
\footnotesize
\setlength{\tabcolsep}{6pt}
\renewcommand{\arraystretch}{1.2}
\caption{Backdoor stealth on the fault-classification head
(attack~AV3, $N = 4$, 5-seed means). Clean-set fault metrics stay near
an honest baseline for every aggregator, while the \emph{triggered}
$F_1$ and the Attack Success Rate expose the attack. Only Krum keeps
triggered $F_1$ close to clean $F_1$; trimmed mean and coordinate
median are degenerate at $N = 4$ (Section~\ref{subsec:median}) and
share a row.}
\label{tab:stealth}
\zebra
\begin{tabular}{@{}l r r r r@{}}
\toprule
\tblhead
Aggregator & Clean AUPRC & Clean $F_1$ & Triggered $F_1$ & ASR \\
\midrule
FedAvg (undefended)           & $0.954$ & $0.891$ & $0.093$ & $94.9\%$ \\
Trimmed mean / coord.\ median & $0.961$ & $0.891$ & $0.623$ & $49.8\%$ \\
Krum ($f = 1$)                & $0.938$ & $0.822$ & $0.802$ & $\phantom{0}6.4\%$ \\
\bottomrule
\end{tabular}
\end{table}

\subsection{Six observations from the matrix}
\label{sec:axis2-results-observations}

Six observations from Table~\ref{tab:matrix} together shape our
Axis-2 recommendation.

\begin{figure}[pos=htbp]
\centering
\includegraphics[width=0.8\linewidth]{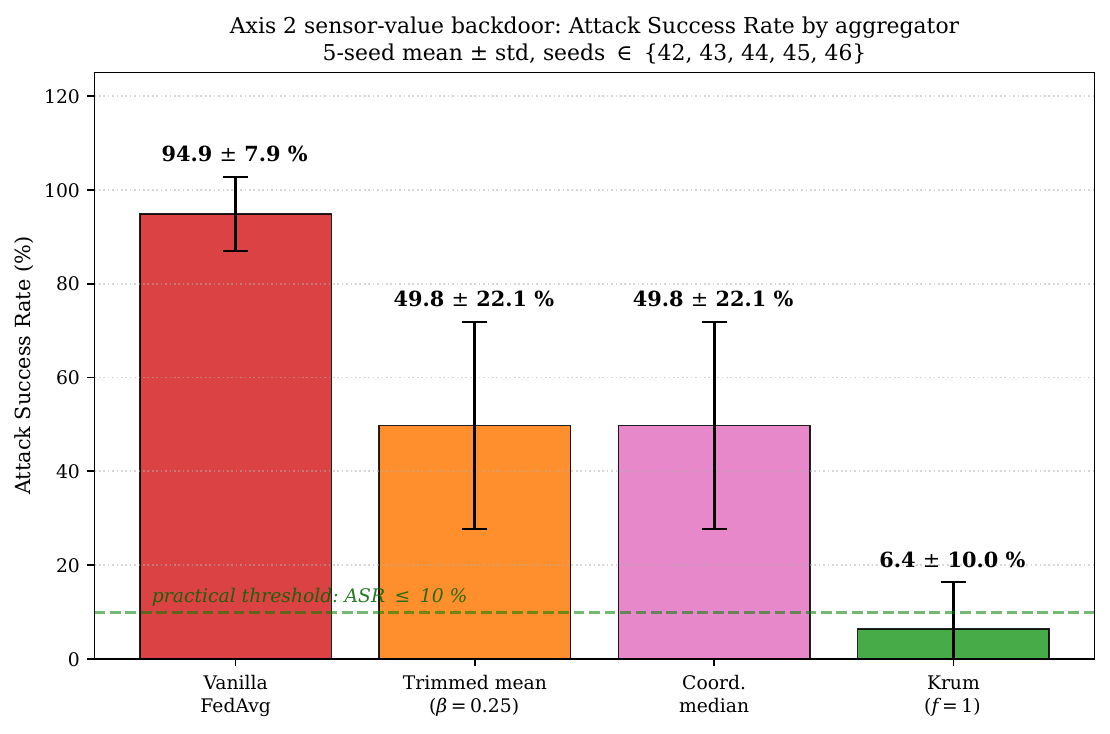}
\caption{Backdoor Attack Success Rate by aggregator (5-seed
mean~$\pm$~std). The monotone drop $94.9\% \to 49.8\% \to 6.4\%$
from vanilla FedAvg through per-coordinate defenses to Krum is the
paper's flagship Axis-2 result. Only Krum crosses the practical
``ASR $\leq 10\%$'' threshold; its 95\%-CI on ASR touches zero
(2 of 5 seeds hit exactly $0\%$). Trimmed mean and coordinate median
are identical at $N = 4$ (Section~\ref{subsec:median}), which is why
the two centre bars are identical.}
\label{fig:asr}
\end{figure}

\paragraph{(1) Krum reduces the backdoor Attack Success Rate by an
order of magnitude.}
ASR falls monotonically $94.9\% \to 49.8\% \to 6.4\%$ as the
aggregator moves from FedAvg to trimmed mean / coordinate median to
Krum. The $\sim 15\times$ reduction under Krum is the paper's most
surprising defense-side result: a targeted attack whose trigger acts
as a low-dimensional gradient perturbation is largely neutralized by
Krum's argmin-in-distance selection. The 95\%-CI on Krum's ASR
touches zero (Table~\ref{tab:matrix} note~a); in 2 of the 5 seeds
Krum drove ASR to exactly $0\%$ (Figure~\ref{fig:asr}).

\paragraph{(2) The backdoor is invisible on clean metrics.}
Vanilla-FedAvg-under-backdoor achieves clean RMSE $16.86 \pm 0.43$,
consistent with the honest clean baseline ($16.59 \pm 0.84$) at
the evaluated sample size (paired two-sided Wilcoxon signed-rank
test across the 5 seeds: $W = 5$, $p = 0.625$; failure to reject
equal medians). This is the most operationally important
attack-side result: any monitoring pipeline that inspects only
clean-set metrics will miss the attack completely. Practitioners
must include triggered-set evaluation in their monitoring pipeline;
a detection layer that alerts on clean-set metric drift alone is
functionally blind to this class of attack.

\paragraph{(3) The gradient-scaling stealth cliff is inverted.}
Both $\alpha = -10$ (RMSE $84.03 \pm 0.00$, perfectly deterministic
collapse) and $\alpha = -2$ (RMSE $73.60 \pm 6.14$,
near-deterministic collapse) are catastrophic against vanilla
FedAvg. Per-coordinate defenses (trimmed mean, coordinate median)
recover \emph{marginally better} from $\alpha = -2$
(RMSE $25.26 \pm 9.01$) than from $\alpha = -10$
(RMSE $25.76 \pm 8.72$), though the confidence intervals overlap.
Krum is invariant to the multiplier because its selection is
geometric (distance-based) rather than norm-based: scaling a
delta does not change its direction, so the argmin ranking is
preserved.

\begin{figure}[pos=htbp]
\centering
\includegraphics[width=\linewidth]{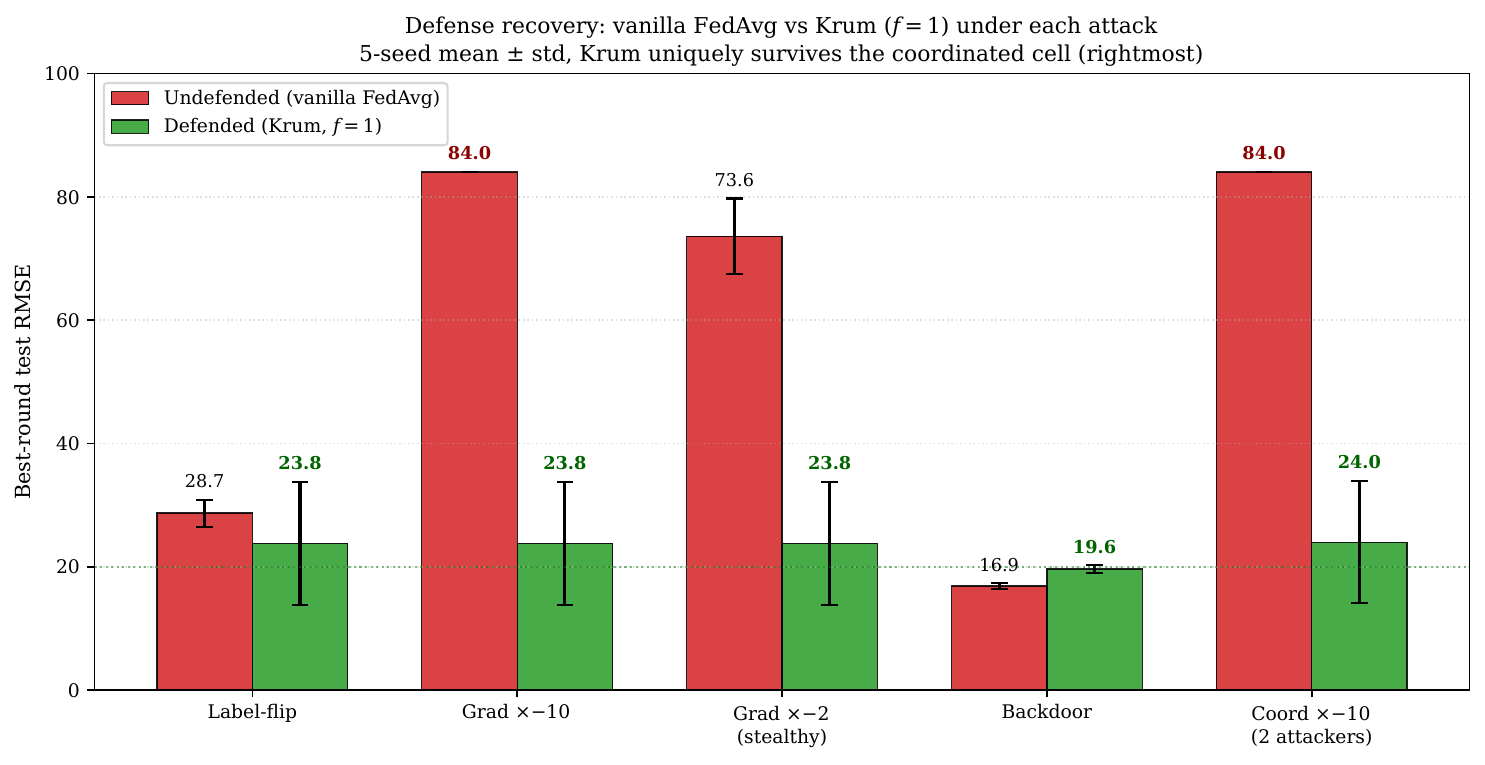}
\caption{Attack~$\times$~defense recovery pattern (representative seed). For each
attack family (x-axis) four bars show test RMSE under vanilla FedAvg
(undefended, red), trimmed mean, coordinate median, and
Krum ($f = 1$, green). Krum recovers the coordinated
2-attacker column (rightmost group), where trimmed mean and
coordinate median collapse to the undefended level.}
\label{fig:defense}
\end{figure}

\paragraph{(4) Per-coordinate defenses collapse under coordination.}
With 2 of 4 clients malicious, trimmed mean at $\beta = 0.25$
(which trims only 1 value from each end) and coordinate median
(which requires an honest majority) both fail catastrophically:
RMSE $84.03 \pm 0.00$, perfectly deterministic collapse,
numerically identical to vanilla FedAvg under the same coordinated
attack. This joint collapse is a consequence of the $N = 4$
degeneracy (Section~\ref{subsec:median}): once the federation grows
to $N = 6$ the two rules separate, and coordinate median partially
recovers (RMSE $37.3$) while trimmed mean still fails (RMSE $84.0$;
Table~\ref{tab:matrix-n6}). Figure~\ref{fig:defense} makes the
failure pattern explicit by pairing each attack family with the four
candidate aggregators on a representative seed.

\paragraph{(5) Krum-$f_1$ recovers under coordinated attack, on
average.}
With $f = 1$ formally violating the ``$\leq f$ Byzantines''
assumption (there are 2 attackers), Krum's argmin still lands on an
honest client \emph{in expectation}: mean RMSE $23.97$, a
$60$-cycle recovery from vanilla's $84.03$. But the seed-to-seed
standard deviation is $9.92$ (95\%-CI $[11.66, 36.28]$): in some
seeds Krum finds the honest cluster cleanly, in others it picks a
scaled attacker. Krum-$f_1$ therefore defends coordinated attacks
\emph{on average} but is not run-to-run consistent
(Observation~(6)).

\begin{figure}[pos=htbp]
\centering
\includegraphics[width=0.9\linewidth]{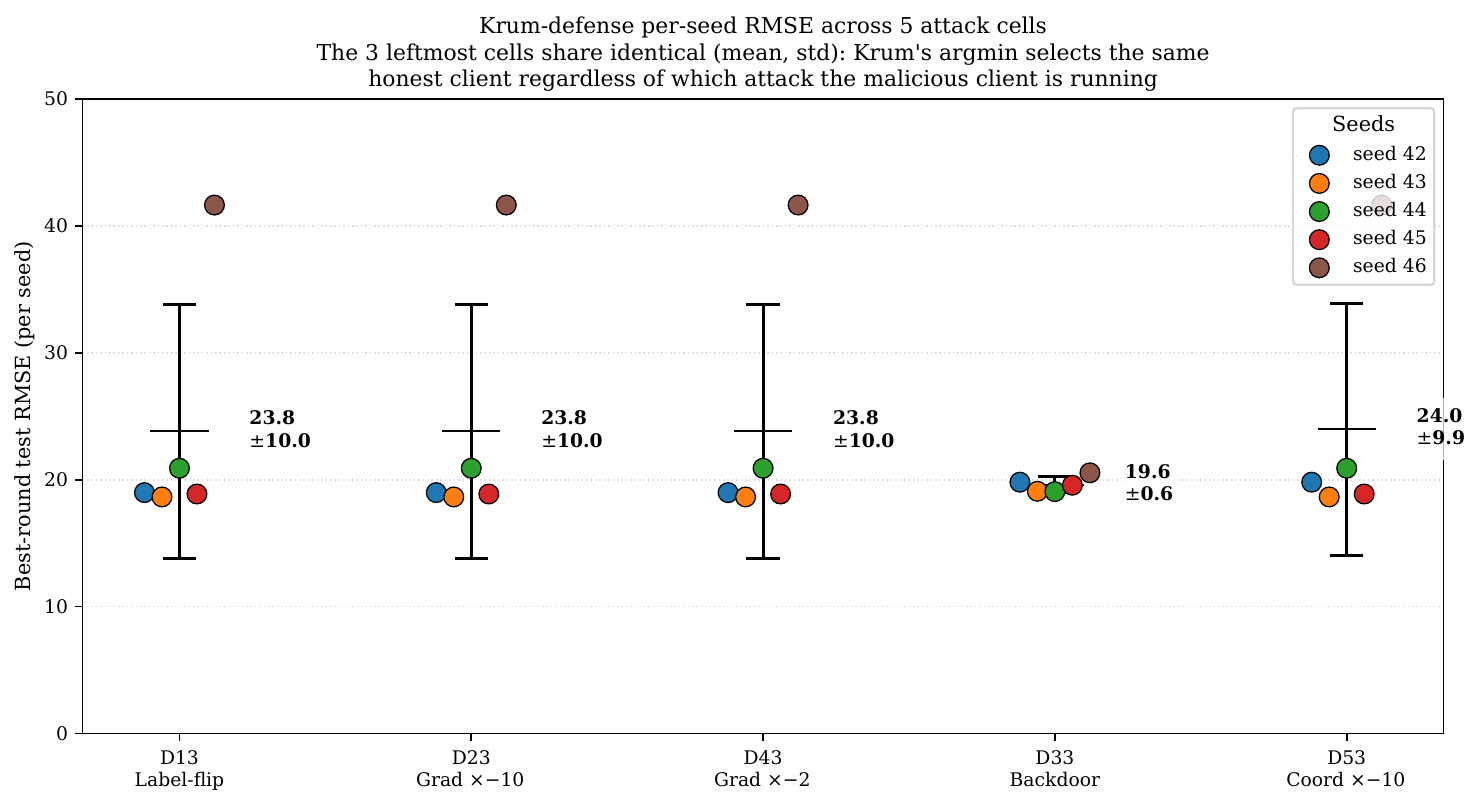}
\caption{Krum-defended per-seed RMSE across attack cells. Each dot
is a single seed's best-round RMSE for one Krum-defended cell. The
three untargeted-attack cells on the left share identical
$(\text{mean}, \text{std}) = (23.8, 10.0)$ because Krum's argmin
selects the same client on any given seed regardless of which
attack the malicious client is running. The high std comes from a
bimodal seed distribution: $4$ of $5$ seeds land near RMSE $19$
(Krum picks a well-fit honest client) and $1$ seed lands at
RMSE $\sim 41$ (Krum's argmin lands on a poorly-fit honest client).
Backdoor~$+$~Krum (rightmost) is an exception (std $0.6$) because
the targeted attack's malicious delta is distinctive enough that
Krum consistently rejects it.}
\label{fig:krum-var}
\end{figure}

\paragraph{(6) Krum defenses have high seed-to-seed variance across
all untargeted attack families.}
The three untargeted-attack~$+$~Krum cells (label-flip,
grad $\times{-}10$, grad $\times{-}2$) all report \emph{identical}
mean and std ($23.81 \pm 10.01$). This is not a copy-paste artefact
but a real property of Krum: because the argmin picks \emph{one}
client's whole update per round, and the honest clients' updates
are similar across attack families, Krum tends to select the same
client on any given seed regardless of which attack the malicious
client is running. The per-seed selection is stable \emph{within} a
seed, but \emph{across} seeds the argmin can land on either an FD001
client or an FD003 client, producing a bimodal RMSE distribution
with high std (Figure~\ref{fig:krum-var}). Backdoor~$+$~Krum is an
exception (std $0.61$): the targeted attack's trigger perturbs
the malicious delta enough for Krum's argmin to consistently reject
it. Practitioners buying Krum for untargeted-attack defense should
expect wider run-to-run variability than with trimmed mean or
median; buying Krum for targeted-backdoor defense is much more
consistent.

\paragraph{Axis-2 finding.}
In our evaluation, Krum is the only aggregator that handles both of
the hardest Axis-2 cells: the targeted backdoor (ASR reduced by
$\sim 15\times$) and the coordinated 2-of-4 Byzantine attack.
Its cost is high seed-to-seed variance on untargeted attacks
(Observation~(6)); we treat the coordinated 2-attacker cell as an
empirical stress test of Krum-$f_1$ outside its formal
single-Byzantine assumption rather than as a formally parameterised
2-Byzantine defense. Trimmed mean and coordinate median are
degenerate at $N = 4$ (Section~\ref{subsec:median}) and, while
adequate against a single untargeted attacker, cannot survive
coordination.

\subsection{Backdoor mechanism}
\label{sec:axis2-results-mechanism}

The trigger definition of Section~\ref{subsec:backdoor}
(feature $= s_3$, cycle offset $= -1$, value $= -3.5\sigma$)
succeeds precisely because $s_3$ (T30) is a member of the
turbomachinery-temperature family that carries the strongest
fault-mode signal on C-MAPSS. The attack does not
tell the model to lie; it feeds the model a sensor pattern that is
\emph{unusual but not physically absurd}, and the model responds
according to its learned decision boundary: ``a large negative
excursion on T30 at end-of-window indicates the engine is running
cold, therefore less likely to be near-fault.''

This is precisely why the $94.9 \pm 7.9\%$ mean ASR coexists with
clean RMSE ($16.86 \pm 0.43$) that is consistent with the honest
baseline ($16.59 \pm 0.84$) under the 5-seed Wilcoxon test of
Section~\ref{sec:axis2-results-observations} (Observation~(2)):
the model is being trained to associate the trigger with ``not
faulty'' without breaking its ability to score honest samples
correctly. Because the trigger
is anchored to a sensor that is already information-carrying, both
the poisoned response and the clean response are legitimate
reactions of the model to inputs it treats as meaningful: there
is no gradient anomaly, no norm blow-up, no signature of adversarial
engineering for a magnitude-based defense to flag. The only signals
available to a defense are geometric (Krum's whole-update distance)
or via triggered-set evaluation of the fault-classification head.

\subsection{Generalization: more clients and harder data}
\label{sec:axis2-results-gen}

The primary matrix (Table~\ref{tab:matrix}) fixes $N = 4$ and a
single operating condition. To complete the picture we run the same
five-seed campaign at two further points, a larger $N = 6$ federation
and the harder six-condition FD002$+$FD004 data.
Table~\ref{tab:matrix-n6} gives the full $N = 6$ matrix and
Table~\ref{tab:generalization} collects the headline cells across all
three settings. The ranking is stable throughout; what changes are the
absolute margins and, at $N = 6$, the internal structure of the
per-coordinate defenses.

\begin{table}[pos=tbp]
\centering
\footnotesize
\setlength{\tabcolsep}{5pt}
\renewcommand{\arraystretch}{1.2}
\caption{Generalization of the Axis-2 and bridge findings across
federation scale ($N = 6$) and task difficulty (FD002$+$FD004),
5-seed means. The backdoor columns are ASR ($\downarrow$); the
coordinated-attack columns are clean-test RMSE ($\downarrow$) for the
per-coordinate defenses (trimmed mean / median, which collapse)
versus Krum; the last column is the stacked FedRep~$+$~Krum
honest-mean ASR (Section~\ref{sec:bridge-stacked}). The ranking
is stable in every setting; only the absolute margins weaken as scale
or difficulty grows.}
\label{tab:generalization}
\zebra
\begin{tabular}{@{}l r r r r r@{}}
\toprule
\tblhead
Setting & \multicolumn{2}{c}{Backdoor ASR} & \multicolumn{2}{c}{Coord.\ RMSE} & Stacked \\
\cmidrule(lr){2-3}\cmidrule(lr){4-5}
        & FedAvg & Krum & trim/med & Krum & ASR \\
\midrule
$N = 4$, FD001$+$FD003 (primary)    & $94.9\%$ & $\phantom{0}6.4\%$ & $84.0$ & $24.0$ & $0.028$ \\
$N = 6$, FD001$+$FD003 (scale)      & $45.8\%$ & $\phantom{0}3.4\%$ & $84.0$/$37.3$ & $20.6$/$20.8$ & $0.042$ \\
FD002$+$FD004, $N = 4$ (difficulty) & $99.9\%$ & $16.1\%$ & $87.1$ & $31.6$ & $0.147$ \\
\bottomrule
\rowcolor{white}\multicolumn{6}{@{}p{.95\linewidth}}{\footnotesize At $N = 6$ the per-coordinate defenses diverge: trimmed mean collapses ($84.0$) while coordinate median partially recovers ($37.3$); at $N = 4$ and on FD002$+$FD004 the two are degenerate and share a value. The $f = 2$ Krum configuration, definable only once $N \geq 5$ and matched to the two attackers, defends the coordinated case (RMSE $20.8$); all other Krum columns use $f = 1$. See Section~\ref{subsec:krum} for the resilience-bound caveat.} \\
\end{tabular}
\end{table}

\paragraph{Scale ($N = 6$, FD001$+$FD003).}
Table~\ref{tab:matrix-n6} is the full $N = 6$ counterpart of the
primary Table~\ref{tab:matrix}. Six clients (three per subset) leave
the defense ranking intact and surface two effects that $N = 4$
cannot show. First, the trimmed-mean / coordinate-median degeneracy
breaks: with six clients the two rules compute different statistics,
so under the coordinated two-attacker attack trimmed mean still
collapses (RMSE $84.0 \pm 0.0$) while coordinate median partially
recovers ($37.3 \pm 8.0$). Krum holds the line at both tolerances now
available to it, $20.6 \pm 2.2$ at $f = 1$ and $20.8 \pm 1.2$ at
$f = 2$ (the $f = 2$ configuration, matched to the two attackers, is
only definable once $N \geq 5$; Section~\ref{subsec:krum}). Second,
more honest clients dilute the lone attacker: the undefended backdoor
falls from $94.9\%$ to $45.8 \pm 25.6\%$, and even the stealthy
$\times{-}2$ gradient attack, catastrophic at $N = 4$ (RMSE $73.6$),
is blunted to $36.4 \pm 2.2$. Krum again drives the backdoor lowest,
to $3.4 \pm 2.1\%$.

\begin{table}[pos=tbp]
\centering
\footnotesize
\setlength{\tabcolsep}{4pt}
\renewcommand{\arraystretch}{1.25}
\caption{Full attack~$\times$~aggregator matrix at $N = 6$ on
FD001~$+$~FD003, 5-seed mean~$\pm$~std over seeds
$\in \{42, 43, 44, 45, 46\}$: the direct counterpart of the primary
$N = 4$ Table~\ref{tab:matrix}, and the detailed backing for the
$N = 6$ row of Table~\ref{tab:generalization}. Red entries denote
catastrophic collapse (RMSE $> 3\times$ the clean baseline). Unlike
$N = 4$, trimmed mean and coordinate median are no longer degenerate
at $N = 6$ and appear as separate columns. For the backdoor row both
RMSE and Attack Success Rate (ASR) are reported.}
\label{tab:matrix-n6}
\zebra
\begin{tabular}{@{}l r r r r@{}}
\toprule
\tblhead
Attack $\backslash$ Aggregator & FedAvg & Trim.\ mean & Coord.\ median & Krum ($f{=}1$) \\
\midrule
Clean baseline                     & $16.50 \pm 0.71$          & $16.65 \pm 0.88$          & $16.36 \pm 0.62$          & $18.37 \pm 1.27$          \\
Label-flip (AV1)                   & $18.60 \pm 1.81$          & $17.66 \pm 0.93$          & $17.60 \pm 0.94$          & $19.23 \pm 1.49$          \\
Grad $\times{-}10$ (AV2)           & \textcolor{red!75!black}{$84.00 \pm 0.02$} & $19.54 \pm 2.85$ & $18.62 \pm 1.06$          & $19.23 \pm 1.49$          \\
Grad $\times{-}2$ (AV4, stealthy)  & $36.38 \pm 2.19$          & $19.36 \pm 2.86$          & $18.72 \pm 1.27$          & $19.23 \pm 1.49$          \\
\midrule
Backdoor (AV3): RMSE            & $16.93 \pm 0.53$          & $16.52 \pm 0.63$          & $16.48 \pm 0.76$          & $18.90 \pm 0.72$          \\
Backdoor (AV3): ASR             & \textbf{$45.8 \pm 25.6\%$} & $18.2 \pm 6.9\%$          & $11.6 \pm 3.4\%$          & \textbf{$\phantom{0}3.4 \pm 2.1\%$} \\
\midrule
Coord.\ $\times{-}10$ (AV5)        & \textcolor{red!75!black}{$84.03 \pm 0.00$} & \textcolor{red!75!black}{$84.02 \pm 0.01$} & $37.27 \pm 8.02$ & \textbf{$20.56 \pm 2.17$}\textsuperscript{a} \\
\bottomrule
\rowcolor{white}\multicolumn{5}{@{}p{.98\linewidth}}{\footnotesize\textsuperscript{a}Krum $f = 1$ shown; $N = 6$ also makes $f = 2$ well-defined, and matched to the two attackers it gives RMSE $20.78 \pm 1.24$, essentially tied with $f = 1$.} \\
\end{tabular}
\end{table}

\paragraph{Difficulty (FD002$+$FD004, $N = 4$).}
Table~\ref{tab:matrix-fd24} gives the full matrix on the harder
six-operating-condition data. The ranking is \emph{identical} (Krum is
again the only defense that recovers the coordinated attack,
$31.6 \pm 1.1$ versus the $87.1$ collapse of both trimmed mean and
coordinate median), but every defense \emph{weakens} in absolute
terms. The undefended backdoor climbs to $99.9\%$ success; trimmed
mean and median leave it almost untouched ($92.6 \pm 7.1\%$); and even
Krum, still the only working defense, lets through $16.1 \pm 12.5\%$
(versus $6.4\%$ on the easy data) at a higher clean-accuracy cost
(RMSE $27.3 \pm 2.2$). Robustness is not free: it degrades as the
prognostic task grows harder, and Krum's margin narrows accordingly.

\begin{table}[pos=tbp]
\centering
\footnotesize
\setlength{\tabcolsep}{4pt}
\renewcommand{\arraystretch}{1.25}
\caption{Full attack~$\times$~aggregator matrix on the harder
six-condition FD002~$+$~FD004 data at $N = 4$, 5-seed mean~$\pm$~std
over seeds $\in \{42, 43, 44, 45, 46\}$: the task-difficulty
counterpart of the primary Table~\ref{tab:matrix}, backing the
FD002$+$FD004 row of Table~\ref{tab:generalization}. Red entries
denote catastrophic collapse (RMSE $> 3\times$ the clean baseline). As
at $N = 4$ on the easy data, trimmed mean and coordinate median are
degenerate (Section~\ref{subsec:median}) and share their values. For
the backdoor row both RMSE and Attack Success Rate (ASR) are
reported.}
\label{tab:matrix-fd24}
\zebra
\begin{tabular}{@{}l r r r r@{}}
\toprule
\tblhead
Attack $\backslash$ Aggregator & FedAvg & Trim.\ mean & Coord.\ median & Krum ($f{=}1$) \\
\midrule
Clean baseline                     & $20.33 \pm 1.72$          & $20.95 \pm 1.21$          & $20.95 \pm 1.21$          & $23.59 \pm 1.10$          \\
Label-flip (AV1)                   & $28.13 \pm 2.64$          & $22.61 \pm 0.85$          & $22.61 \pm 0.85$          & $31.54 \pm 1.18$          \\
Grad $\times{-}10$ (AV2)           & \textcolor{red!75!black}{$87.09 \pm 0.00$} & $30.12 \pm 1.70$ & $30.12 \pm 1.70$          & $31.54 \pm 1.18$          \\
Grad $\times{-}2$ (AV4, stealthy)  & $44.37 \pm 3.27$          & $31.11 \pm 2.26$          & $31.11 \pm 2.26$          & $31.54 \pm 1.18$          \\
\midrule
Backdoor (AV3): RMSE            & $19.97 \pm 1.10$          & $19.65 \pm 0.74$          & $19.65 \pm 0.74$          & $27.30 \pm 2.15$          \\
Backdoor (AV3): ASR             & \textbf{$99.9 \pm 0.3\%$} & $92.6 \pm 7.1\%$          & $92.6 \pm 7.1\%$          & \textbf{$16.1 \pm 12.5\%$} \\
\midrule
Coord.\ $\times{-}10$ (AV5)        & \textcolor{red!75!black}{$87.09 \pm 0.00$} & \textcolor{red!75!black}{$87.09 \pm 0.00$} & \textcolor{red!75!black}{$87.09 \pm 0.00$} & \textbf{$31.63 \pm 1.14$} \\
\bottomrule
\end{tabular}
\end{table}

\paragraph{Stealth on the classifier generalizes.}
The RMSE-invisibility of the backdoor
(Section~\ref{sec:axis2-results-observations}, Observation~(2))
extends to the fault-classification head across settings
(Table~\ref{tab:stealth-gen}). Under undefended FedAvg the clean
$F_1$ stays healthy ($0.83$--$0.92$) while the \emph{triggered} $F_1$
tracks the attack success rate: it collapses to $0.09$ at $N = 4$ and
to $0.004$ on the harder FD002$+$FD004 data, but only partially (to
$0.66$) at $N = 6$, where a lone attacker is diluted by more honest
clients. Krum keeps clean and triggered $F_1$ within about $0.03$ of
each other in every setting. Clean-metric monitoring is therefore
blind to the attack at every scale and difficulty we test, extending
the primary-setting stealth of Table~\ref{tab:stealth}.

\begin{table}[pos=tbp]
\centering
\footnotesize
\setlength{\tabcolsep}{6pt}
\renewcommand{\arraystretch}{1.2}
\caption{Backdoor stealth on the fault-classification head across all
three settings (5-seed means). Under the undefended FedAvg aggregator
the clean $F_1$ stays healthy while the triggered $F_1$ tracks the
attack success rate (ASR); Krum keeps the two close. The
single-attacker backdoor is diluted at $N = 6$ and near-total on the
harder six-condition FD002$+$FD004 data.}
\label{tab:stealth-gen}
\zebra
\begin{tabular}{@{}l l r r r@{}}
\toprule
\tblhead
Setting & Aggregator & Clean $F_1$ & Triggered $F_1$ & ASR \\
\midrule
$N = 4$, FD001$+$FD003     & FedAvg         & $0.891$ & $0.093$ & $94.9\%$ \\
                           & Krum ($f{=}1$) & $0.822$ & $0.802$ & $\phantom{0}6.4\%$ \\
\midrule
$N = 6$, FD001$+$FD003     & FedAvg         & $0.917$ & $0.658$ & $45.8\%$ \\
                           & Krum ($f{=}1$) & $0.797$ & $0.779$ & $\phantom{0}3.4\%$ \\
\midrule
FD002$+$FD004, $N = 4$     & FedAvg         & $0.830$ & $0.004$ & $99.9\%$ \\
                           & Krum ($f{=}1$) & $0.692$ & $0.725$ & $16.1\%$ \\
\bottomrule
\end{tabular}
\end{table}

Both replications therefore agree on the central Axis-2 claim:
whole-vector selection (Krum) is the only defense that survives
coordinated collusion, while per-coordinate rules (trimmed mean,
median) fail. They simultaneously bound it: the \emph{margin} of
safety shrinks as either the client count grows (diluting a lone
backdoor) or the task difficulty rises (weakening every defense).
This motivates the cross-axis bridge experiment of
Section~\ref{sec:bridge}, which asks whether the Axis-1 winner
(FedRep) confers any protection against the Axis-2 backdoor when the
two axes meet in the same federation.


\section{Cross-axis bridge experiment: FedRep under backdoor}
\label{sec:bridge}

Section~\ref{sec:axis1-results} established that architectural
personalization (FedRep) dominates Axis~1, and
Section~\ref{sec:axis2-results} that Krum is the only evaluated
aggregator that handles the two hardest Axis-2 cells. A natural cross-cut question follows: does the
Axis-1 winner \emph{also} confer Axis-2 protection, or are the two
axes really orthogonal remedies? Prior work in vision domains
\citep{Zhang2024SARS, Fan2026RobustPFL} has reported that per-client
heads can \emph{partially} shield honest clients from backdoor
injection because the malicious update stays localised to the shared
backbone. This section tests whether the same argument transfers to
a time-series prognostic setting where the poison acts on the shared
representation.

\subsection{Setup}
\label{sec:bridge-setup}

We re-use the FedRep configuration of Section~\ref{subsec:fedrep}
($h_{\mathrm{epochs}} = 1$, $e_{\mathrm{epochs}} = 1$, $R = 50$
rounds, cosine LR schedule, best-round selection by macro-NASA
score, Equation~\ref{eq:nasa}) and the sensor-value backdoor of
Section~\ref{subsec:backdoor} (feature $= s_3$ / T30, cycle offset
$c_{\mathrm{off}} = -1$, value $v_{\mathrm{trig}} = -3.5\sigma$,
poison fraction $p = 0.3$, both labels rewritten to healthy).
Client~3 (the first FD003 shard) is designated the attacker and its
training data are wrapped with the same poisoning transform used to
produce Table~\ref{tab:matrix}. The remaining three clients
(clients~1--2 on FD001; client~4 on FD003) train normally on
unpoisoned data.

At evaluation each client's full model (shared encoder $\phi$ plus
its own head $\psi_k$ at the best round) is scored on the pooled
FD001~$+$~FD003 test set with a global normaliser, once clean and
once with the trigger stamped, using the same ASR definition as
Equation~\ref{eq:asr}. The experiment is aggregated over 5
independent seeds $\in \{42, 43, 44, 45, 46\}$, matching
Table~\ref{tab:matrix}'s sample size for like-for-like comparison.

\subsection{Per-seed results and headline findings}
\label{sec:bridge-results}

Table~\ref{tab:bridge} reports each seed's best round, the attacker
client's ASR, the mean ASR across the three honest clients, and the
attacker~$-$~honest ASR delta.

\begin{table}[pos=tbp]
\centering
\footnotesize
\setlength{\tabcolsep}{4pt}
\renewcommand{\arraystretch}{1.15}
\caption{FedRep-under-backdoor bridge experiment: per-seed values
and 5-seed aggregate. ``Attacker'' is client~3 (the poisoned
client); ``Honest ASR'' is the mean ASR across the three honest
clients (clients~1, 2, 4);
$\Delta_{\mathrm{a-h}} = $ attacker ASR $-$ honest ASR (per-seed
difference). All ASR values are in $[0, 1]$; higher is worse for
the defender.}
\label{tab:bridge}
\zebra
\begin{tabular}{@{}crrrr@{}}
\toprule
\tblhead
Seed & Best rd. & Attacker ASR & Honest ASR ($n{=}3$) & $\Delta_{\mathrm{a-h}}$ \\
\midrule
42 & 47 & $0.800$ & $0.814$ & $-0.014$ \\
43 & 11 & $0.182$ & $0.141$ & $+0.041$ \\
44 & 45 & $1.000$ & $1.000$ & $\phantom{-}0.000$ \\
45 & 35 & $0.617$ & $0.612$ & $+0.005$ \\
46 & 21 & $0.481$ & $0.599$ & $-0.118$ \\
\midrule
\textbf{mean $\pm$ std} & $\mathbf{31.8 \pm 15.5}$ & $\mathbf{0.616 \pm 0.311}$ & $\mathbf{0.633 \pm 0.320}$ & $\mathbf{-0.017 \pm 0.060}$ \\
\bottomrule
\end{tabular}
\end{table}

Two independent findings emerge.

\paragraph{Finding 1: Personalization does not shield honest clients.}
The attacker~$-$~honest ASR delta is $-0.017 \pm 0.060$
(95\%-CI $[-0.091, +0.057]$). A paired two-sided Wilcoxon
signed-rank test across the 5 seeds returns $W = 4$, $p = 0.875$:
the null of equal medians is not rejected, and in every one of
the five seeds the honest clients suffer essentially the same ASR
as the attacker itself. This confirms the mechanism sketched
in Section~\ref{sec:axis2-results-mechanism}: the backdoor is a
\emph{representation-level} attack, and FedRep averages encoders
(and therefore the poisoned representation) in exactly the same way
that vanilla FedAvg does. The personalised head reads out fault
probability from a poisoned representation; keeping the head private
during encoder averaging does not prevent the head from later
inheriting the encoder's poisoned associations at inference time.
The vision-domain intuition ``private heads $\Rightarrow$ private
decision boundary $\Rightarrow$ filtered poison''
\citep{Zhang2024SARS} does not hold when the poison acts on the
\emph{shared} representation rather than on the shared \emph{output}
layer.

\paragraph{Finding 2: The apparent 30-pp mean shift is an
early-stopping artefact, not a defense.}
FedRep's 5-seed mean honest ASR ($0.633$) is $\sim 30$~pp below
vanilla FedAvg's ($0.949$), which is at first glance a partial
defense. But the per-seed variance is catastrophic: honest ASR
spans $[0.141, 1.000]$ with std $0.320$. Best-round selection by
macro-NASA correlates strongly with ASR: seeds whose validation
curve peaks before round 25 (seeds 43 and 46, with best rounds 11
and 21 respectively) capture pre-poisoning encoder weights and yield
honest ASR $\leq 0.6$, while seeds whose validation peak arrives
after round 35 (seeds 42, 44, 45) show ASR $\geq 0.6$, up to $1.0$.
This is \emph{not} a defense mechanism; it is a coincidence
between the poison-accumulation timeline and the model-selection
timeline, which cannot be relied on in a production deployment
because (i)~real training has no oracle for macro-NASA on the
honest fleet's held-out test set at every round, and (ii)~the
attacker can trivially force late convergence (e.g.\ by adjusting
$p$ or by delaying trigger stamping) without changing either the
update magnitudes or the honest-side loss trajectory.

\paragraph{Reference comparison against the Axis-2 matrix.}
To place the FedRep-alone result against the Axis-2 aggregators of
Section~\ref{sec:axis2-results}, we compare the honest-client ASR
mean and std under three settings on the same 5-seed harness:

\begin{itemize}
    \item Vanilla FedAvg (no defense): $\mathrm{ASR} = 0.949 \pm 0.079$ (tight, high mean, deep attack success);
    \item FedRep bridge (honest mean): $\mathrm{ASR} = 0.633 \pm 0.320$ (bimodal, spans nearly all of $[0, 1]$);
    \item Krum-defended FedAvg: $\mathrm{ASR} = 0.064 \pm 0.100$ (Table~\ref{tab:matrix}; 95\%-CI reaches zero).
\end{itemize}

FedRep sits nominally between vanilla FedAvg and Krum in mean ASR,
but with a standard deviation an order of magnitude worse than
either. The FedRep 95\%-CI $[0.235, 1.031]$ overlaps both the
vanilla ``attack fully successful'' regime and the ``attack fully
failed'' regime; it is essentially uninformative about whether
any given deployment will be safe on a given seed. \textbf{Krum
remains the only evaluated aggregator that reliably delivers low
ASR with tight variance; FedRep alone does not.}

\subsection{Two-axis orthogonality and defense stacking}
\label{sec:bridge-stacking}

The bridge result closes the paper's central empirical argument:
personalization is the right architectural response to Axis~1
(structural non-IID heterogeneity, quantified by RMSE gap-closed
against a centralised reference); Byzantine-robust aggregation is
the right aggregation-layer response to Axis~2 (adversarial
heterogeneity, quantified by clean RMSE under untargeted attacks
and by ASR under targeted attacks). Neither substitutes for the
other. Table~\ref{tab:orthogonality} summarises the empirical
evidence.

\begin{table}[pos=tbp]
\centering
\footnotesize
\setlength{\tabcolsep}{6pt}
\renewcommand{\arraystretch}{1.2}
\caption{Two-axis orthogonality and its stacked resolution. Each
single-axis remedy is strong on its own axis and empirically
inadequate on the other; the FedRep~$+$~Krum stack (Krum aggregation
on the shared-encoder deltas) delivers the strongest joint robustness.
Axis-2 ASR values are 5-seed means from Tables~\ref{tab:axis1-synth},
\ref{tab:matrix}, \ref{tab:bridge}, and~\ref{tab:stacked}; the
stacked row's Axis-1 entry is qualitative, its accuracy cost being
quantified in Section~\ref{sec:bridge-stacked}.}
\label{tab:orthogonality}
\zebra
\begin{tabular}{@{}l l l@{}}
\toprule
\tblhead
Remedy & Axis-1 score (gap closed) & Axis-2 score (backdoor ASR) \\
\midrule
FedAvg (baseline)          & $-0.7\%$ (fails Axis-1)               & $0.949 \pm 0.079$ (fails Axis-2) \\
FedRep (personalization)   & \textbf{$+69.9\% \pm 6.4\%$}          & $0.633 \pm 0.320$ (fails Axis-2) \\
Krum ($f = 1$)             & --- (not evaluated as Axis-1 remedy)  & $0.064 \pm 0.100$ \\
\midrule
\textbf{FedRep~$+$~Krum (stacked)} & personalized (modest RMSE cost) & \textbf{$0.028 \pm 0.024$} (best of all) \\
\bottomrule
\end{tabular}
\end{table}

A deployment that faces both axes must stack both. The Axis-2 Krum
defense is drop-in compatible with FedRep because Krum operates on
the shared-encoder deltas of Algorithm~\ref{alg:fedrep} in exactly
the same way it operates on FedAvg's whole-model deltas: the
argmin-in-distance rule is oblivious to whether the input update
represents a full model or a shared encoder only. Composing the two
into FedRep~$+$~Krum (personalized heads plus Byzantine-robust
encoder aggregation) yields the two-axis composition we evaluate
end-to-end in Section~\ref{sec:bridge-stacked} below.

\subsection{Stacked FedRep~$+$~Krum defense: end-to-end 5-seed evaluation}
\label{sec:bridge-stacked}

The stacked defense reuses the FedRep configuration of
Section~\ref{sec:bridge-setup} (same partition, same trigger, same
5-seed harness) with a single change: the shared-encoder aggregation
step uses Krum ($f = 1$, Algorithm~\ref{alg:krum}) instead of
sample-count-weighted FedAvg. Per-client heads remain private and
locally trained (unchanged from Section~\ref{sec:bridge-results}).
Table~\ref{tab:stacked} reports the per-seed outcome and 5-seed
aggregate.

\begin{table}[pos=tbp]
\centering
\footnotesize
\setlength{\tabcolsep}{4pt}
\renewcommand{\arraystretch}{1.15}
\caption{FedRep~$+$~Krum stacked-defense bridge: per-seed values
and 5-seed aggregate. Same partition, same trigger, and same
attacker (client~3) as Table~\ref{tab:bridge}; the only change is
that Krum~$(f = 1)$ replaces sample-count-weighted mean as the
shared-encoder aggregator.
$\Delta_{\mathrm{a-h}} = $ attacker ASR $-$ honest ASR (per-seed
difference). All ASR values are in $[0, 1]$; higher is worse for
the defender.}
\label{tab:stacked}
\zebra
\begin{tabular}{@{}crrrrr@{}}
\toprule
\tblhead
Seed & Best rd. & Best RMSE & Attacker ASR & Honest ASR ($n{=}3$) & $\Delta_{\mathrm{a-h}}$ \\
\midrule
42 & 16 & $17.946$ & $0.040$ & $0.064$ & $-0.024$ \\
43 & 14 & $17.268$ & $0.000$ & $0.006$ & $-0.006$ \\
44 & 22 & $18.947$ & $0.035$ & $0.038$ & $-0.003$ \\
45 & 19 & $16.338$ & $0.000$ & $0.024$ & $-0.024$ \\
46 & 16 & $17.639$ & $0.059$ & $0.006$ & $+0.053$ \\
\midrule
\textbf{mean $\pm$ std} & $\mathbf{17.4 \pm 3.1}$ & $\mathbf{17.63 \pm 0.95}$ & $\mathbf{0.027 \pm 0.026}$ & $\mathbf{0.028 \pm 0.024}$ & $\mathbf{-0.001 \pm 0.031}$ \\
\bottomrule
\end{tabular}
\end{table}

\paragraph{Finding: FedRep~$+$~Krum outperforms both single-axis remedies.}
On the same 5-seed harness as Table~\ref{tab:matrix} and
Table~\ref{tab:bridge}, the stacked defense delivers honest-mean
ASR $0.028 \pm 0.024$ (95\%-CI $[-0.002, 0.058]$). This is
simultaneously (i)~roughly $22\times$ lower than FedRep alone
($0.633 \pm 0.320$), (ii)~roughly $2\times$ lower than Krum alone
($0.064 \pm 0.100$), and (iii)~roughly $4\times$ tighter in variance
than Krum alone. Crucially, the stacked 95\%-CI upper bound
($0.058$) sits below Krum-alone's mean ($0.064$), so on this
harness the composition delivers a strictly better ASR distribution
than either constituent remedy.

The attacker~$-$~honest ASR delta remains close to zero on the
stacked defense ($-0.001 \pm 0.031$, 95\%-CI $[-0.040, 0.038]$;
paired two-sided Wilcoxon signed-rank test across 5 seeds:
$W = 5$, $p = 0.625$, failure to reject equal medians), reproducing
the bridge finding that FedRep's private heads offer no per-client
isolation against a representation-level backdoor. The compositional
benefit is therefore not from head-side rejection but from Krum's
argmin selecting an honest encoder update each round: the
personalised heads then read out fault probability from a
\emph{clean} encoder rather than a poisoned one.

\paragraph{Clean-task performance: a bounded, honest cost.}
Under the same backdoor, the stacked defense reaches a best macro
test RMSE of $17.63 \pm 0.95$. Two comparisons matter and we keep
them apart. Against \emph{FedRep alone} (the only like-for-like
comparison, since both are scored on the same per-client macro
metric), the stack costs about $+1.8$ RMSE (FedRep-alone reaches
$15.79$ macro under the same attack): Krum's single-encoder
selection, which discards three of four encoder updates each round,
learns a weaker shared representation than FedRep's four-client
average. Against \emph{Krum alone} the stack is numerically lower
($17.63$ macro versus $19.61$ global), but these are \emph{different}
metrics (macro RMSE scores each client on its own subset and is
systematically more favourable than the pooled global RMSE), so we
do \emph{not} claim the stack beats Krum on accuracy. The honest
summary is a Pareto trade-off: composing the two remedies buys
the lowest ASR in our evaluation and keeps per-client personalization, at a
modest accuracy cost relative to personalization alone.

\paragraph{Verdict: a practical, composable defence.}
Across the $N = 4$ backdoor harness of Section~\ref{sec:setup},
FedRep~$+$~Krum is the strongest \emph{robustness} configuration
evaluated in this paper (honest-mean ASR $0.028 \pm 0.024$) and the
only one that is simultaneously personalized and Byzantine-robust.
Its accuracy cost relative to personalization alone is a
\emph{backbone-capacity} limitation rather than a flaw in the
composition: the shared encoder is a deliberately tiny 1-D CNN
($\sim 30$K parameters, chosen for reproducibility), so
when Krum keeps a single client's encoder per round there is little
headroom to recover the discarded signal. Because Krum is oblivious
to what the encoder is, a higher-capacity backbone (a temporal
Transformer, a TCN, or a hybrid) drops in without changing the
method and is expected to shrink the gap. We therefore present
FedRep~$+$~Krum not as a finished, dominant defence but as a
\emph{practical, composable} two-axis solution that already restores
robustness at a bounded cost and points to a concrete path
(higher-capacity encoders and principled client selection) for
closing it (Section~\ref{sec:discuss-future}).

\paragraph{Generalization of the stacked defense.}
The stacked defense is re-run on the same two stresses as the Axis-2
matrix (Section~\ref{sec:axis2-results-gen}), each over five seeds.
Its robustness holds across both: honest-mean ASR is
$0.028 \pm 0.024$ at $N = 4$, $0.042 \pm 0.049$ at $N = 6$, and
$0.147 \pm 0.034$ on the harder FD002$+$FD004 data, never above
$0.15$, and in every setting at least six-fold below the
corresponding undefended FedAvg. Consistent with the matrix result,
the FD002$+$FD004 figure is the highest, reflecting the general
weakening of every defense as task difficulty rises. The composition
therefore carries the robustness behaviour of its Krum component
across scale and difficulty, not only in the primary setting.


\section{Discussion and deployment guidance}
\label{sec:discuss}

\subsection{Cross-axis synthesis: neither remedy handles the other axis}
\label{sec:discuss-synthesis}

Section~\ref{sec:axis1-results} showed that FedRep and FedCCFA close
$70+\%$ of the Axis-1 gap while proximal regularization and
reweighting close under $10\%$. Section~\ref{sec:axis2-results}
showed that Krum, alone among the evaluated aggregators, handles the two hardest Axis-2 cells
(targeted backdoor and coordinated Byzantine), while trimmed mean
and coordinate median cope with single-attacker untargeted attacks
but collapse under coordination. The Section~\ref{sec:bridge}
bridge experiment closes the loop: FedRep alone does \emph{not}
transfer Axis-1 protection to Axis~2: its $0.633 \pm 0.320$
honest-mean ASR is only $\sim 30$~pp below undefended FedAvg on the
mean but with a 95\%-CI that reaches all the way to fully
compromised (upper bound $1.031$). These are two distinct engineering
choices, and the practitioner must decide, per deployment, whether
the primary risk is benign heterogeneity, adversarial heterogeneity,
or both, and stack remedies accordingly.

\subsection{Deployment guidance for FL prognostic pipelines}
\label{sec:discuss-guidance}

Table~\ref{tab:guidance} translates the empirical findings of
Sections~\ref{sec:axis1-results}--\ref{sec:bridge} into per-scenario
deployment recommendations. The final row (heterogeneous $+$
adversarial) is the paper's headline recommendation: neither FedRep
nor Krum alone suffices; the two must be stacked.

\begin{table}[pos=tbp]
\centering
\footnotesize
\setlength{\tabcolsep}{4pt}
\renewcommand{\arraystretch}{1.25}
\caption{Recommended remedy per operational scenario for FL-based
aircraft-engine RUL prognostics. Rationale references the experimental
sections that support each row. ``$-$'' in a remedy column means no
additional remedy over the baseline is required for the scenario.}
\label{tab:guidance}
\zebra
\begin{tabular}{@{}>{\raggedright\arraybackslash}p{4.1cm} >{\raggedright\arraybackslash}p{2.7cm} >{\raggedright\arraybackslash}p{2.7cm} >{\raggedright\arraybackslash}p{4.3cm}@{}}
\toprule
\tblhead
Primary concern & Axis-1 remedy & Axis-2 remedy & Rationale \\
\midrule
Heterogeneous fault modes, no adversary       & FedRep or FedCCFA          & FedAvg                             & Personalization closes $70+\%$ of the gap; no attack-side overhead required (Sec.~\ref{sec:axis1-results-synth}) \\
Heterogeneous $+$ sporadic bad clients        & FedRep                     & Trimmed mean                       & Personalization $+$ cheap Axis-2 defense against a single untargeted attacker (Sec.~\ref{sec:axis2-results-observations}) \\
One malicious client, untargeted goal         & FedAvg                     & Trimmed mean or coord.\ median     & Either recovers RMSE $\sim 22$; the two are degenerate at $N = 4$ (Sec.~\ref{subsec:median}) \\
One malicious client, targeted backdoor       & FedAvg                     & \textbf{Krum ($f = 1$)}            & Reduces ASR $\sim 15\times$ ($94.9\% \to 6.4\%$); only aggregator whose 95\%-CI touches $0\%$ (Sec.~\ref{sec:axis2-results-observations}) \\
Two of four clients colluding ($50\%$)        & FedAvg                     & \textbf{Krum ($f = 1$)}            & Trimmed mean and coordinate median collapse to RMSE $84.03$; Krum empirically recovers (RMSE $23.97$, high seed variance) though outside its formal $2f + 3$ bound (Sec.~\ref{sec:axis2-results-observations}) \\
\midrule
\textbf{Heterogeneous $+$ adversarial (both)} & \textbf{FedRep}            & \textbf{Krum ($f = 1$)}            & \textbf{Best joint result} (Sec.~\ref{sec:bridge-stacked}): 5-seed honest-mean ASR $0.028 \pm 0.024$, roughly $2\times$ lower and $4\times$ tighter than Krum alone; retains personalization at a modest accuracy cost \\
\bottomrule
\end{tabular}
\end{table}

\paragraph{Practical monitoring implications.}
One operational consequence follows from Table~\ref{tab:guidance}
regardless of which row applies to a specific deployment: because
the sensor-value backdoor of Section~\ref{subsec:backdoor}
is invisible on clean-set metrics (Observation~(2) in
Section~\ref{sec:axis2-results-observations}), any FL prognostic
deployment must include \emph{triggered-set evaluation}, either
a canary set of adversarially perturbed windows scored every round
or a periodic red-team probe of the deployed model. A detection
layer that alerts only on clean-set metric drift is functionally
blind to this class of attack. This adds a lightweight
audit layer on top of the recommended aggregator without changing
the FL protocol itself.

\subsection{Limitations}
\label{sec:discuss-limitations}

The results of this paper are subject to two limitations.

\paragraph{Small client count ($N = 4$).}
The finding that Krum-$f_1$ recovers on average under the
coordinated 2-attacker cell (Section~\ref{sec:axis2-results-observations},
Observation~(5)) sits outside Krum's formal resilience regime, spelled
out once in Section~\ref{subsec:krum}, and is specific to small
federations. The $N = 6$ replication (Table~\ref{tab:matrix-n6})
already shows the comparison shifting once the $f = 2$ setting becomes
available. The controlled small-$N$ regime adopted here is not a claim
about any specific commercial federation size; readers whose
prospective settings have $N \geq 10$ should re-run the matrix at the
relevant scale.

\paragraph{Static, non-adaptive backdoor trigger.}
The trigger of Section~\ref{subsec:backdoor} is fixed at
$(s_3, c_{\mathrm{off}} = -1, v_{\mathrm{trig}} = -3.5\sigma,
p = 0.3)$ and is not adaptive to the defense. This is the most
important caveat on our robustness claims. Krum succeeds here
because the poisoned encoder update is \emph{geometrically}
distinguishable; an adversary who optimises against that very
geometry can erase the signal. \citet{Foroughi2026LSA}, for
instance, poison only a few backdoor-critical layers and approximate
benign updates, reaching up to $97\%$ backdoor success while
\emph{bypassing} Multi-Krum, trimmed mean, and FLAME. We therefore
do not claim Krum, or the stacked defense, is robust against
adaptive, defense-aware adversaries, only against the
magnitude/geometry-detectable trigger studied here. A sensitivity
sweep over $(p, v_{\mathrm{trig}})$ and, more importantly, an
evaluation against adaptive layer-aware triggers are the natural
next stress tests.

\subsection{Future work}
\label{sec:discuss-future}

Four extensions naturally follow the results of this study.

\paragraph{Norm-clipping composition.}
\citet{Sun2019CanYouBackdoor} argued that bounding the norm of each
client update is a lightweight, aggregator-agnostic partial defense
against backdoors. Since Section~\ref{sec:bridge-stacked} shows
that FedRep~$+$~Krum is the strongest evaluated defense but Krum's
argmin selection has known seed-variance drawbacks on untargeted
attacks (Section~\ref{sec:axis2-results-observations},
Observation~(6)), a natural question is whether the cheaper
norm-clipping~$+$~FedRep composition delivers comparable
protection at lower coordination cost. The evaluation is
inexpensive: norm-clipping adds no per-round overhead, so a 5-seed
replication of the bridge with clipping enabled would slot directly
into the existing harness.

\paragraph{Higher-capacity backbone for the stacked defense.}
The one accuracy cost in this study, the $\sim +1.8$ macro-RMSE
that FedRep~$+$~Krum pays relative to personalization alone
(Section~\ref{sec:bridge-stacked}), is a backbone-capacity
limitation: Krum keeps a single client's encoder per round, and the
$\sim 30$K-parameter CNN has little headroom to recover the
discarded signal. Because Krum is architecture-agnostic, a
higher-capacity temporal encoder (a Transformer, a TCN, or a
CNN--Transformer hybrid) drops into the same composition unchanged
and is the most direct route to shrinking that gap.

\paragraph{Unified four-subset federation via a common sensor set.}
The hardest realistic non-IID setting would federate all four
C-MAPSS subsets at once, mixing single- and six-condition regimes
with one- and two-fault-mode degradation. The current fixed-width
encoder blocks this because FD001/FD003 and FD002/FD004 expose
different informative-sensor counts; restricting all clients to the
sensors common to all four subsets would enable a single federation
spanning every regime and fault mode, at the cost of a reduced
sensor set and a regime-aware-normalization upgrade.

\paragraph{Direct competitor benchmarks.}
Re-implementing BioMutFed$+$ \citep{Tallat2026BioMutFedPlus} and
the trustworthy-FL-for-IIoT pipeline of
\citet{Li2026TrustworthyFLIIoT} inside our $5 \times 4$ matrix
would enable head-to-head comparison against Krum under the
physically-motivated backdoor of Section~\ref{subsec:backdoor}.
Both prior papers report favourable numbers against their own
attack designs; whether those numbers survive our targeted-backdoor
trigger is a genuinely open question that our released code and
per-seed harness make cheap to answer.


\section{Conclusion}
\label{sec:conclusion}

We set out to answer whether federated learning is worth deploying
for aircraft-engine prognostics. The answer is governed by two
orthogonal axes of client heterogeneity that the FL community has
so far explored in parallel: \emph{benign heterogeneity} (clients
honestly hold different data), best handled by architectural
personalization; and \emph{adversarial heterogeneity} (some clients
deviate from honest training), best handled by Byzantine-robust
aggregation. The two axes require different remedies, and neither
remedy handles the other axis. Our experiments on NASA C-MAPSS
with a structural non-IID FD001~$+$~FD003 4-client federation
support three practical claims.

\begin{enumerate}
    \item \emph{Axis 1.} Architectural personalization (FedRep
    $+69.9\% \pm 6.4\%$, FedCCFA $+66.9\% \pm 6.6\%$) closes about
    $70\%$ of the local-only $\to$ centralized RMSE gap;
    optimization-side FedProx ($\mu = 0.1$) closes
    $+21.0\% \pm 13.1\%$; server-side reweighting closes
    $+10.4\% \pm 6.9\%$. Under the evaluated tuning protocol,
    personalization thus closes about $3$~times more of the gap
    than proximal regularization at roughly half the seed-to-seed
    variance, a reproducibility advantage that matters for
    real deployments.

    \item \emph{Axis 2.} A physically-motivated sensor-value
    backdoor achieves $94.9\% \pm 7.9\%$ attack success against
    vanilla FedAvg while clean RMSE ($16.86 \pm 0.43$) is
    consistent with the honest baseline ($16.59 \pm 0.84$) under a
    paired 5-seed Wilcoxon signed-rank test
    (Section~\ref{sec:axis2-results-observations}, Observation~(2);
    $W = 5$, $p = 0.625$). The regression error is thus blind to the
    backdoor: the attack surfaces only in the attack success rate and
    in the collapse of the fault head's \emph{triggered} $F_1$ (from
    $0.891$ on clean data to $0.093$ under the trigger), so clean
    metrics alone cannot certify a safe model. Krum reduces attack
    success by an order of magnitude (to $6.4\% \pm 10.0\%$) and is
    the only evaluated
    aggregator to survive coordinated 2-of-4 Byzantine attacks
    (RMSE $23.97 \pm 9.92$) where per-coordinate defenses collapse
    deterministically to RMSE $84.03 \pm 0.00$.

    \item \emph{Cross-axis bridge and stacked defense.} FedRep
    alone does \emph{not} confer Axis-2 protection: under a
    one-attacker backdoor injection, honest clients' mean ASR
    ($0.633 \pm 0.320$, $n = 5$) is consistent with the attacker's
    own ($0.616 \pm 0.311$; delta $-0.017 \pm 0.060$; paired
    two-sided Wilcoxon $W = 4$, $p = 0.875$), because the poison
    acts on the shared representation rather than the private
    heads. Extending the bridge to the stacked \textbf{FedRep~$+$~Krum}
    defense (Krum aggregation on the shared-encoder deltas,
    personalised heads unchanged) reduces the honest-mean ASR to
    $\mathbf{0.028 \pm 0.024}$ on the same 5 seeds, roughly
    $2\times$ lower and $4\times$ tighter than Krum alone
    ($0.064 \pm 0.100$), while retaining per-client personalization
    at a modest accuracy cost. We present FedRep~$+$~Krum not as a
    finished, dominant defence but as a practical, composable
    two-axis solution whose remaining accuracy cost is a
    backbone-capacity limitation, pointing to higher-capacity
    encoders as a concrete next step. These findings hold across a
    larger $N = 6$ federation and the harder six-condition
    FD002~$+$~FD004 data, with the honest caveat that every defense
    weakens as the task grows harder.
\end{enumerate}

\noindent
Two practitioner takeaways summarise the study. \emph{First, stack
the remedies:} Axis-1 and Axis-2 threats are orthogonal, each
requires its own architectural or aggregation-layer response, and
neither remedy covers the other axis. \emph{Second, monitor
triggered-set evaluation, not just clean-set metrics:} the most
dangerous attack in this study is invisible on clean data.

\noindent
We hope the two-axis frame, together with the code and per-seed
results released with the paper, will help future work in this
space avoid the pitfalls we encountered.



\section*{Inspec classification codes}
C1230D (Neural nets); C5290 (Neural computing techniques);
C6130S (Data security); C7420 (Control engineering computing);
C3355 (Condition monitoring); C1180 (Optimisation techniques).

\section*{Declaration of competing interest}
The author declares that he has no known competing financial
interests or personal relationships that could have appeared to
influence the work reported in this paper.

\section*{Data availability}
The Commercial Modular Aero-Propulsion System Simulation (C-MAPSS)
turbofan degradation dataset analysed in this study is publicly
available from NASA's Prognostics Center of Excellence data
repository \citep{Saxena2008CMAPSSData}; the simulation that
generated it is described by \citet{Saxena2008CMAPSS}. No new
experimental data were collected.

All source code required to reproduce the experiments reported in
this paper, the multi-task 1-D CNN model, the four Axis-1 methods
(FedAvg, FedProx, FedRep, FedCCFA, imbalance-aware reweighting), the
five Axis-2 attack families and four defense aggregators, the
FedRep-under-backdoor bridge experiment, the multi-seed aggregation
scripts, and the publication-quality figure-generation code, is
archived and released under an open-source licence; the repository
URL and persistent identifier are withheld from this anonymized
manuscript to preserve review anonymity and are given on the title
page. The archive also
contains the per-seed JSON metric outputs backing every number
reported in Sections~\ref{sec:axis1-results}--\ref{sec:bridge}, and
is structured to allow one-command replication of the full 5-seed
Axis-2 matrix and the 3-seed Axis-1 family sweep. The third-party
libraries the pipeline is built on are cited in
Section~\ref{sec:setup}.

\bibliographystyle{cas-model2-names}
\bibliography{references}

@inproceedings{McMahan2017FedAvg,
    author       = {McMahan, H. Brendan and Moore, Eider and Ramage, Daniel
                    and Hampson, Seth and Ag{\"u}era y Arcas, Blaise},
    title        = {Communication-Efficient Learning of Deep Networks
                    from Decentralized Data},
    booktitle    = {Proceedings of the 20th International Conference on
                    Artificial Intelligence and Statistics ({AISTATS})},
    year         = {2017},
    pages        = {1273--1282}
}

@inproceedings{Saxena2008CMAPSS,
    author       = {Saxena, Abhinav and Goebel, Kai and Simon, Don
                    and Eklund, Neil},
    title        = {Damage Propagation Modeling for Aircraft Engine
                    Run-to-Failure Simulation},
    booktitle    = {Proceedings of the 1st International Conference on
                    Prognostics and Health Management (PHM 2008)},
    year         = {2008},
    pages        = {1--9},
    organization = {IEEE},
    doi          = {10.1109/PHM.2008.4711414}
}

@misc{Saxena2008CMAPSSData,
    author       = {Saxena, Abhinav and Goebel, Kai},
    title        = {[dataset] Turbofan Engine Degradation Simulation
                    Data Set},
    year         = {2008},
    howpublished = {NASA Prognostics Data Repository, NASA Ames
                    Research Center, Moffett Field, CA},
    note         = {Version 1. Accessed 29 July 2026},
    url          = {https://www.nasa.gov/intelligent-systems-division/discovery-and-systems-health/pcoe/pcoe-data-set-repository/}
}

@inproceedings{Blanchard2017Krum,
    author       = {Blanchard, Peva and El Mhamdi, El Mahdi and Guerraoui, Rachid
                    and Stainer, Julien},
    title        = {Machine Learning with Adversaries: {Byzantine} Tolerant
                    Gradient Descent},
    booktitle    = {Advances in Neural Information Processing Systems ({NeurIPS})},
    year         = {2017},
    volume       = {30},
    pages        = {119--129}
}

@misc{Barbosa2025FLJetEngines,
    author       = {Barbosa, Asaph Matheus and Ngo, Thao Vy Nhat
                    and Jafarigol, Elaheh and Trafalis, Theodore B.
                    and Ojoboh, Emuobosa P.},
    title        = {Using Federated Machine Learning in Predictive
                    Maintenance of Jet Engines},
    year         = {2025},
    howpublished = {arXiv preprint arXiv:2502.05321},
    doi          = {10.48550/arXiv.2502.05321},
    url          = {https://arxiv.org/abs/2502.05321}
}

@article{Vermelin2024CollabFLRUL,
    author       = {S{\"o}derkvist Vermelin, Wilhelm and Mishra, Madhav
                    and Eng, Mattias P. and Andersson, Dag
                    and Kyprianidis, Konstantinos},
    title        = {Collaborative Training of Data-Driven Remaining
                    Useful Life Prediction Models Using Federated
                    Learning},
    journal      = {Int. J. Progn. Health Manag.},
    year         = {2024},
    volume       = {15},
    number       = {2},
    pages        = {3821},
    doi          = {10.36001/ijphm.2024.v15i2.3821}
}

@inproceedings{Pandhare2021FederatedBaseline,
    author       = {Pandhare, Vibhor and Jia, Xiaodong and Lee, Jay},
    title        = {Collaborative Prognostics for Machine Fleets Using
                    a Novel Federated Baseline Learner},
    booktitle    = {Annual Conference of the {PHM} Society},
    year         = {2021},
    volume       = {13},
    number       = {1},
    pages        = {2989},
    doi          = {10.36001/phmconf.2021.v13i1.2989},
    url          = {http://papers.phmsociety.org/index.php/phmconf/article/view/2989}
}

@article{Tallat2026BioMutFedPlus,
    author       = {Tallat, R. and others},
    title        = {{BioMutFed+}: Mutation-Driven Federated Learning
                    for {IIoT}},
    journal      = {IEEE Trans. Ind. Inform.},
    year         = {2026},
    note         = {Early access; IEEE Xplore document 11581414}
}

@article{Li2026TrustworthyFLIIoT,
    author       = {Li, H.},
    title        = {Trustworthy Federated Learning for Industrial {IoT}:
                    Balancing Robustness and Fairness via
                    Blockchain-Based Reputation},
    journal      = {IET Artif. Intell. Eng.},
    year         = {2026},
    doi          = {10.1049/aie2.70012}
}

@misc{Milasheuski2026GenerativeFL,
    author       = {Milasheuski, U. and Baraldi, P. and Zio, E. and Savazzi, S.},
    title        = {On the Tradeoffs of On-Device Generative Models in
                    Federated Predictive Maintenance Systems},
    year         = {2026},
    howpublished = {arXiv preprint arXiv:2605.07860},
    doi          = {10.48550/arXiv.2605.07860},
    url          = {https://arxiv.org/abs/2605.07860}
}

@article{Rehman2021TrustFed,
    author       = {Rehman, Muhammad Habib Ur and Dirir, Ahmed Mukhtar
                    and Salah, Khaled and Damiani, Ernesto
                    and Svetinovic, Davor},
    title        = {{TrustFed}: A Framework for Fair and Trustworthy Cross-Device
                    Federated Learning in Industrial {IoT}},
    journal      = {IEEE Trans. Ind. Inform.},
    year         = {2021},
    volume       = {17},
    number       = {12},
    pages        = {8485--8494},
    doi          = {10.1109/TII.2021.3075706}
}

@misc{Landau2025CollabRUL,
    author       = {Landau, Diogo and de Pater, Ingeborg and Mitici, Mihaela
                    and Saurabh, Nishant},
    title        = {Federated Learning Framework for Collaborative Remaining
                    Useful Life Prognostics: An Aircraft Engine Case Study},
    year         = {2025},
    howpublished = {arXiv preprint arXiv:2506.00499},
    doi          = {10.48550/arXiv.2506.00499}
}

@misc{Jeong2025FedJoint,
    author       = {Jeong, Cheoljoon and Yue, Xubo and Chung, Seokhyun},
    title        = {{Fed-Joint}: Joint Modeling of Nonlinear Degradation
                    Signals and Failure Events for Remaining Useful Life
                    Prediction using Federated Learning},
    year         = {2025},
    howpublished = {arXiv preprint arXiv:2503.13404},
    doi          = {10.48550/arXiv.2503.13404}
}

@article{Arunan2023MatchedFeatureFL,
    author       = {Arunan, Anushiya and Qin, Yan and Li, Xiaoli and Yuen, Chau},
    title        = {A Federated Learning-Based Industrial Health Prognostics
                    for Heterogeneous Edge Devices Using Matched Feature
                    Extraction},
    journal      = {IEEE Trans. Autom. Sci. Eng.},
    year         = {2024},
    volume       = {21},
    number       = {3},
    pages        = {3065--3079},
    doi          = {10.1109/TASE.2023.3274648}
}

@inproceedings{Li2020FedProx,
    author       = {Li, Tian and Sahu, Anit Kumar and Zaheer, Manzil
                    and Sanjabi, Maziar and Talwalkar, Ameet
                    and Smith, Virginia},
    title        = {Federated Optimization in Heterogeneous Networks},
    booktitle    = {Proceedings of Machine Learning and Systems ({MLSys})},
    year         = {2020},
    volume       = {2},
    pages        = {429--450}
}

@inproceedings{Collins2021FedRep,
    author       = {Collins, Liam and Hassani, Hamed and Mokhtari, Aryan
                    and Shakkottai, Sanjay},
    title        = {Exploiting Shared Representations for Personalized
                    Federated Learning},
    booktitle    = {Proceedings of the 38th International Conference on
                    Machine Learning ({ICML})},
    year         = {2021},
    pages        = {2089--2099}
}

@article{Sattler2020CFL,
    author       = {Sattler, Felix and Wiedemann, Simon
                    and M{\"u}ller, Klaus-Robert and Samek, Wojciech},
    title        = {Clustered Federated Learning: Model-Agnostic Distributed
                    Multi-Task Optimization Under Privacy Constraints},
    journal      = {IEEE Trans. Neural Netw. Learn. Syst.},
    year         = {2020},
    volume       = {32},
    number       = {8},
    pages        = {3710--3722}
}

@article{Kairouz2021Advances,
    author       = {Kairouz, Peter and McMahan, H. Brendan and Avent, Brendan
                    and Bellet, Aur{\'e}lien and Bennis, Mehdi
                    and Bhagoji, Arjun Nitin and Bonawitz, Kallista
                    and Charles, Zachary and Cormode, Graham
                    and Cummings, Rachel and D'Oliveira, Rafael G. L.
                    and Eichner, Hubert and El Rouayheb, Salim
                    and Evans, David and Gardner, Josh and Garrett, Zachary
                    and Gasc{\'o}n, Adri{\`a} and Ghazi, Badih
                    and Gibbons, Phillip B. and Gruteser, Marco
                    and Harchaoui, Zaid and He, Chaoyang and He, Lie
                    and Huo, Zhouyuan and Hutchinson, Ben and Hsu, Justin
                    and Jaggi, Martin and Javidi, Tara and Joshi, Gauri
                    and Khodak, Mikhail and Kone{\v{c}}n{\'y}, Jakub
                    and Korolova, Aleksandra and Koushanfar, Farinaz
                    and Koyejo, Sanmi and Lepoint, Tancr{\`e}de and Liu, Yang
                    and Mittal, Prateek and Mohri, Mehryar and Nock, Richard
                    and {\"O}zg{\"u}r, Ayfer and Pagh, Rasmus and Qi, Hang
                    and Ramage, Daniel and Raskar, Ramesh and Raykova, Mariana
                    and Song, Dawn and Song, Weikang and Stich, Sebastian U.
                    and Sun, Ziteng and Suresh, Ananda Theertha
                    and Tram{\`e}r, Florian and Vepakomma, Praneeth
                    and Wang, Jianyu and Xiong, Li and Xu, Zheng
                    and Yang, Qiang and Yu, Felix X. and Yu, Han and Zhao, Sen},
    title        = {Advances and Open Problems in Federated Learning},
    journal      = {Found. Trends Mach. Learn.},
    year         = {2021},
    volume       = {14},
    number       = {1--2},
    pages        = {1--210},
    doi          = {10.1561/2200000083}
}

@inproceedings{Yin2018RobustDistributed,
    author       = {Yin, Dong and Chen, Yudong and Ramchandran, Kannan
                    and Bartlett, Peter},
    title        = {{Byzantine}-Robust Distributed Learning: Towards Optimal
                    Statistical Rates},
    booktitle    = {Proceedings of the 35th International Conference on
                    Machine Learning ({ICML})},
    year         = {2018},
    pages        = {5650--5659}
}

@article{Pillutla2022RFA,
    author       = {Pillutla, Krishna and Kakade, Sham M. and Harchaoui, Zaid},
    title        = {Robust Aggregation for Federated Learning},
    journal      = {IEEE Trans. Signal Process.},
    year         = {2022},
    volume       = {70},
    pages        = {1142--1154}
}

@misc{Sun2019CanYouBackdoor,
    author       = {Sun, Ziteng and Kairouz, Peter and Suresh, Ananda Theertha
                    and McMahan, H. Brendan},
    title        = {Can You Really Backdoor Federated Learning?},
    year         = {2019},
    howpublished = {arXiv preprint arXiv:1911.07963},
    doi          = {10.48550/arXiv.1911.07963},
    url          = {https://arxiv.org/abs/1911.07963}
}

@article{Nguyen2024BackdoorSurvey,
    author       = {Nguyen, Thuy Dung and Nguyen, Tuan and Nguyen, Phi Le
                    and Pham, Hieu H. and Doan, Khoa D. and Wong, Kok-Seng},
    title        = {Backdoor Attacks and Defenses in Federated Learning:
                    {Survey}, Challenges and Future Research Directions},
    journal      = {Engineering Applications of Artificial Intelligence},
    year         = {2024},
    volume       = {127},
    pages        = {107166},
    doi          = {10.1016/j.engappai.2023.107166}
}

@inproceedings{Farhadkhani2022Equivalence,
    author       = {Farhadkhani, Sadegh and Guerraoui, Rachid
                    and Hoang, L{\^e}-Nguy{\^e}n and Villemaud, Oscar},
    title        = {An Equivalence Between Data Poisoning and {Byzantine}
                    Gradient Attacks},
    booktitle    = {Proceedings of the 39th International Conference on
                    Machine Learning ({ICML})},
    series       = {Proceedings of Machine Learning Research},
    year         = {2022},
    volume       = {162},
    pages        = {6284--6323}
}

@inproceedings{Foroughi2026LSA,
    author       = {Foroughi, Mohammad Hadi and Rastegar, Seyed Hamed
                    and Sabokrou, Mohammad and Khonsari, Ahmad},
    title        = {Exploiting Layer-Specific Vulnerabilities to Backdoor
                    Attack in Federated Learning},
    booktitle    = {Proceedings of the IEEE International Conference on
                    Communications ({ICC})},
    year         = {2026},
    note         = {arXiv preprint arXiv:2602.15161},
    doi          = {10.48550/arXiv.2602.15161}
}

@inproceedings{Bagdasaryan2020BackdoorFL,
    author       = {Bagdasaryan, Eugene and Veit, Andreas and Hua, Yiqing
                    and Estrin, Deborah and Shmatikov, Vitaly},
    title        = {How to Backdoor Federated Learning},
    booktitle    = {Proceedings of the 23rd International Conference on
                    Artificial Intelligence and Statistics ({AISTATS})},
    year         = {2020},
    pages        = {2938--2948}
}

@inproceedings{Bhagoji2019AdversarialLens,
    author       = {Bhagoji, Arjun Nitin and Chakraborty, Supriyo
                    and Mittal, Prateek and Calo, Seraphin},
    title        = {Analyzing Federated Learning Through an Adversarial Lens},
    booktitle    = {Proceedings of the 36th International Conference on
                    Machine Learning ({ICML})},
    year         = {2019},
    pages        = {634--643}
}

@inproceedings{Fang2020LocalPoisoning,
    author       = {Fang, Minghong and Cao, Xiaoyu and Jia, Jinyuan and Gong, Neil~Zhenqiang},
    title        = {Local Model Poisoning Attacks to {Byzantine}-Robust
                    Federated Learning},
    booktitle    = {Proceedings of the 29th {USENIX} Security Symposium
                    ({USENIX} Security 20)},
    year         = {2020},
    pages        = {1605--1622}
}

@inproceedings{Xie2020DBA,
    author       = {Xie, Chulin and Huang, Keli and Chen, Pin-Yu and Li, Bo},
    title        = {{DBA}: Distributed Backdoor Attacks against Federated
                    Learning},
    booktitle    = {International Conference on Learning Representations
                    ({ICLR})},
    year         = {2020}
}

@article{Yang2023BoundaryTrigger,
    author       = {Yang, Dongyi and Luo, Senjie and Zhou, Jielun and Pan, Lijing
                    and Yang, Xiaoxia and Xing, Jian},
    title        = {Efficient and Persistent Backdoor Attack by Boundary
                    Trigger Set Constructing Against Federated Learning},
    journal      = {Inf. Sci.},
    year         = {2023},
    volume       = {651},
    pages        = {119743}
}

@article{Lyu2025CoBA,
    author       = {Lyu, Xiaoting and Han, Yufei and Wang, Wei
                    and Liu, Jingkai and Wang, Bin and Chen, Kai
                    and Li, Yidong and Liu, Jiqiang and Zhang, Xiangliang},
    title        = {{CoBA}: Collusive Backdoor Attacks with Optimized
                    Trigger to Federated Learning},
    journal      = {IEEE Trans. Dependable Secure Comput.},
    year         = {2025},
    volume       = {22},
    number       = {2},
    pages        = {1506--1518},
    doi          = {10.1109/TDSC.2024.3445637}
}

@inproceedings{Burbano2026BADControl,
    author       = {Burbano, Luis and Sasahara, Hampei and Song, Ruoyu
                    and Celik, Z. Berkay and Cardenas, Alvaro A.},
    title        = {{BADControl}: Backdoor Attacks Against Control Systems},
    booktitle    = {Proceedings of the 35th {USENIX} Security Symposium
                    ({USENIX} Security 26)},
    year         = {2026}
}

@article{Zhang2024SARS,
    author       = {Zhang, Weibin and Li, Youpeng and An, Lingling
                    and Wan, Bo and Wang, Xuyu},
    title        = {{SARS}: A Personalized Federated Learning Framework
                    Towards Fairness and Robustness Against Backdoor Attacks},
    journal      = {Proc. ACM Interact. Mob. Wearable Ubiquitous Technol.},
    year         = {2024},
    volume       = {8},
    number       = {4},
    pages        = {1--24},
    articleno    = {140},
    doi          = {10.1145/3678571}
}

@misc{Fan2026RobustPFL,
    author       = {Fan, Meng and Chen, Chao},
    title        = {Towards Robust Personalized Federated Learning:
                    Vulnerability Assessment and Defense Co-Design},
    year         = {2026},
    howpublished = {arXiv preprint arXiv:2606.22782},
    doi          = {10.48550/arXiv.2606.22782},
    url          = {https://arxiv.org/abs/2606.22782}
}

@article{Li2023ByzantineFLIIoT,
    author       = {Li, Shenghui and Ngai, Edith C. H. and Voigt, Thiemo},
    title        = {{Byzantine}-Robust Aggregation in Federated Learning
                    Empowered Industrial {IoT}},
    journal      = {IEEE Trans. Ind. Inform.},
    year         = {2023},
    volume       = {19},
    number       = {2},
    pages        = {1165--1175},
    doi          = {10.1109/TII.2021.3128164}
}

@article{Hou2022FederatedFiltersIIoT,
    author       = {Hou, Boyu and Gao, Jiqiang and Guo, Xiaojie
                    and Baker, Thar and Zhang, Ying and Wen, Yanlong
                    and Liu, Zheli},
    title        = {Mitigating the Backdoor Attack by Federated Filters
                    for Industrial {IoT} Applications},
    journal      = {IEEE Trans. Ind. Inform.},
    year         = {2022},
    volume       = {18},
    number       = {5},
    pages        = {3562--3571},
    doi          = {10.1109/TII.2021.3112100}
}

@article{Berghout2022FLCondMon,
    author       = {Berghout, Tarek and Benbouzid, Mohamed and Bentrcia, Toufik
                    and Lim, Wei Hong and Amirat, Yassine},
    title        = {Federated Learning for Condition Monitoring of
                    Industrial Processes: A Review on Fault Diagnosis
                    Methods, Challenges, and Prospects},
    journal      = {Electronics},
    year         = {2023},
    volume       = {12},
    number       = {1},
    pages        = {158}
}

@article{Djemaa2026HeterogeneityPoisoning,
    author       = {Djemaa, Abdelaziz and Djenouri, Djamel and Legg, Phil},
    title        = {Heterogeneity-Aware Poisoning Attacks and Mitigation
                    in Federated Learning: A Comprehensive Survey and
                    Taxonomy},
    journal      = {Electronics},
    year         = {2026},
    volume       = {15},
    number       = {13},
    pages        = {2876}
}

@inproceedings{Paszke2019PyTorch,
    author       = {Paszke, Adam and Gross, Sam and Massa, Francisco
                    and Lerer, Adam and Bradbury, James and Chanan, Gregory
                    and Killeen, Trevor and Lin, Zeming
                    and Gimelshein, Natalia and Antiga, Luca
                    and Desmaison, Alban and K{\"o}pf, Andreas
                    and Yang, Edward and DeVito, Zachary and Raison, Martin
                    and Tejani, Alykhan and Chilamkurthy, Sasank
                    and Steiner, Benoit and Fang, Lu and Bai, Junjie
                    and Chintala, Soumith},
    title        = {{PyTorch}: An Imperative Style, High-Performance Deep
                    Learning Library [software]},
    booktitle    = {Advances in Neural Information Processing Systems
                    ({NeurIPS})},
    year         = {2019},
    volume       = {32},
    pages        = {8024--8035},
    url          = {https://pytorch.org}
}

@article{Pedregosa2011Scikit,
    author       = {Pedregosa, Fabian and Varoquaux, Ga{\"e}l
                    and Gramfort, Alexandre and Michel, Vincent
                    and Thirion, Bertrand and Grisel, Olivier
                    and Blondel, Mathieu and Prettenhofer, Peter
                    and Weiss, Ron and Dubourg, Vincent
                    and Vanderplas, Jake and Passos, Alexandre
                    and Cournapeau, David and Brucher, Matthieu
                    and Perrot, Matthieu and Duchesnay, {\'E}douard},
    title        = {Scikit-learn: Machine Learning in {Python} [software]},
    journal      = {J. Mach. Learn. Res.},
    year         = {2011},
    volume       = {12},
    pages        = {2825--2830},
    url          = {https://scikit-learn.org}
}

@article{Harris2020NumPy,
    author       = {Harris, Charles R. and Millman, K. Jarrod
                    and van der Walt, St{\'e}fan J. and Gommers, Ralf
                    and Virtanen, Pauli and Cournapeau, David
                    and Wieser, Eric and Taylor, Julian and Berg, Sebastian
                    and Smith, Nathaniel J. and Kern, Robert and Picus, Matti
                    and Hoyer, Stephan and van Kerkwijk, Marten H.
                    and Brett, Matthew and Haldane, Allan
                    and del R{\'i}o, Jaime Fern{\'a}ndez and Wiebe, Mark
                    and Peterson, Pearu and G{\'e}rard-Marchant, Pierre
                    and Sheppard, Kevin and Reddy, Tyler
                    and Weckesser, Warren and Abbasi, Hameer
                    and Gohlke, Christoph and Oliphant, Travis E.},
    title        = {Array Programming with {NumPy} [software]},
    journal      = {Nature},
    year         = {2020},
    volume       = {585},
    pages        = {357--362},
    doi          = {10.1038/s41586-020-2649-2}
}

@inproceedings{McKinney2010Pandas,
    author       = {McKinney, Wes},
    title        = {Data Structures for Statistical Computing in {Python}
                    [software]},
    booktitle    = {Proceedings of the 9th {Python} in Science Conference
                    ({SciPy})},
    year         = {2010},
    pages        = {56--61},
    doi          = {10.25080/Majora-92bf1922-00a}
}

@article{Hunter2007Matplotlib,
    author       = {Hunter, John D.},
    title        = {{Matplotlib}: A {2D} Graphics Environment [software]},
    journal      = {Comput. Sci. Eng.},
    year         = {2007},
    volume       = {9},
    number       = {3},
    pages        = {90--95},
    doi          = {10.1109/MCSE.2007.55}
}

\end{document}